\documentclass[10pt,twocolumn,letterpaper]{article}

\usepackage[pagenumbers]{wacv}

\usepackage{graphicx}
\usepackage{amsmath}
\usepackage{amssymb}
\usepackage{booktabs}

\usepackage{subcaption}
\usepackage{times}
\usepackage{epsfig}

\usepackage{etoolbox}
\usepackage{xstring}
\usepackage[utf8]{inputenc}
\usepackage[T1]{fontenc}
\usepackage{wrapfig}
\usepackage{amsfonts,mathrsfs}
\usepackage{nicefrac}
\usepackage{microtype}
\usepackage{xcolor}
\usepackage{mfirstuc}
\usepackage{mathtools}
\usepackage{siunitx}
\usepackage{fontawesome5}

\usepackage[pagebackref,breaklinks,colorlinks]{hyperref}%
\hypersetup{urlcolor=black}

\usepackage[capitalize]{cleveref}
\crefname{section}{Sec.}{Secs.}
\Crefname{section}{Section}{Sections}
\Crefname{table}{Table}{Tables}
\crefname{table}{Tab.}{Tabs.}
\crefname{appendix}{App.}{Apps.}

\usepackage{bm}
\newcommand{\mathbold}[1]{\bm{#1}}
\newcommand{\mbf}[1]{\mathbf{#1}}

\usepackage{xspace}

\newcommand{\dd}{\,\mathrm{d}}

\newcommand{\R}{\mathbb{R}}
\newcommand{\N}{\mathrm{N}}

\DeclareMathOperator*{\softmax}{softmax}

\DeclareMathOperator*{\nlpd}{NLPD}
\DeclareMathOperator*{\ece}{ECE}

\newcommand{\vphi}[0]{\mathbold{\phi}}

\newcommand{\vtheta}[0]{\mathbold{\theta}}

\newcommand{\MSigma}[0]{\mathbold{\Sigma}}

\renewcommand{\mid}[0]{\,|\,}

\newcommand{\vb}{\mbf{b}}

\newcommand{\vg}{\mbf{g}}

\newcommand{\vx}{\mbf{x}}
\newcommand{\vy}{\mbf{y}}
\newcommand{\vz}{\mbf{z}}

\newcommand{\MH}{\mbf{H}}
\newcommand{\MI}{\mbf{I}}

\newcommand{\ML}{\mbf{L}}

\newcommand{\MV}{\mbf{V}}
\newcommand{\MU}{\mbf{U}}
\newcommand{\MW}{\mbf{W}}

\def\naive{na\"ive\xspace}

\makeatletter
\@namedef{ver@everyshi.sty}{}
\makeatother
\usepackage{tikz,pgfplots}
\usetikzlibrary{arrows}
\tikzset{>=stealth'}
\usepgfplotslibrary{groupplots}
\usetikzlibrary{plotmarks,shapes,arrows,patterns}
\pgfplotsset{compat=newest} 

\usepackage{tabularx}
\usepackage{array,multirow}
\usepackage{colortbl}
\newcommand{\PreserveBackslash}[1]{\let\temp=\\#1\let\\=\temp}
\newcolumntype{C}[1]{>{\PreserveBackslash\centering}p{#1}}
\newlength{\tblw}

\newlength\figureheight
\newlength\figurewidth

\renewcommand{\paragraph}[1]{\smallskip\noindent\textbf{#1}~~}

\definecolor{aaltoyellow}{RGB}{254,203,00}
\definecolor{aaltored}{RGB}{237,41,57}
\definecolor{aaltoblue}{RGB}{00,101,189}
\definecolor{aaltogray}{RGB}{146,139,129}
\definecolor{aaltolgreen}{RGB}{105,190,40}
\definecolor{aaltodgreen}{RGB}{00,155,58}
\definecolor{aaltocyan}{RGB}{00,168,180}
\definecolor{aaltopurple}{RGB}{102,57,183}
\definecolor{aaltomagenta}{RGB}{177,05,157}
\definecolor{aaltoorange}{RGB}{255,121,00}

\definecolor{mycolor0}{rgb}{0.2667,0.4471,0.7098}
\definecolor{mycolor1}{rgb}{0.1647,0.6706,0.3804}
\definecolor{mycolor2}{rgb}{0.8275,0.2627,0.3059}
\definecolor{mycolor3}{rgb}{0.5216,0.4392,0.7176}
\definecolor{mycolor4}{rgb}{0.8118,0.7255,0.4118}
\definecolor{mycolor5}{rgb}{0.2745,0.7176,0.8157}

\definecolor{mylcolor0}{rgb}{0.6902,0.7686,0.8863}
\definecolor{mylcolor1}{rgb}{0.5451,0.8902,0.6941}
\definecolor{mylcolor2}{rgb}{0.9412,0.7490,0.7647}
\definecolor{mylcolor3}{rgb}{0.8627,0.8392,0.9176}
\definecolor{mylcolor4}{rgb}{0.9569,0.9373,0.8667}
\definecolor{mylcolor5}{rgb}{0.7529,0.9020,0.9373}
\definecolor{mylcolor6}{rgb}{0.8750,0.8750,0.8750}

\definecolor{C0}{HTML}{FFA042}
\definecolor{C1}{HTML}{007ABB}

\pgfplotsset{every axis/.append style={
		legend style={inner xsep=1pt, inner ysep=0.5pt, nodes={inner sep=1pt, text depth=0.1em},draw=none,fill=white,rounded corners=.5pt}
}}

\def\wacvPaperID{2070}
\def\confName{WACV}
\def\confYear{2024}

\begin{document}

\title{Fixing Overconfidence in Dynamic Neural Networks}

\author{
Lassi Meronen$^{1,2}$ \qquad Martin Trapp$^1$ \qquad Andrea Pilzer$^3$ \qquad Le Yang$^4$ \qquad Arno Solin$^1$ \\[2pt]
\begin{minipage}{.23\textwidth}\centering
$^1$Aalto University \\
\end{minipage}
\hfill
\begin{minipage}{.23\textwidth}\centering
$^2$Saab Finland Oy \\
\end{minipage}
\hfill
\begin{minipage}{.23\textwidth}\centering
$^3$NVIDIA \\
\end{minipage}
\hfill
\begin{minipage}{.27\textwidth}\centering
$^4$Xi'an Jiaotong University \\
\end{minipage}%
\\[2pt]
{\small\tt \{lassi.meronen, martin.trapp, arno.solin\}@aalto.fi, apilzer@nvidia.com, yangle15@xjtu.edu.cn}
}
\maketitle

\begin{abstract}
Dynamic neural networks are a recent technique that promises a remedy for the increasing size of modern deep learning models by dynamically adapting their computational cost to the difficulty of the inputs. In this way, the model can adjust to a limited computational budget. However, the poor quality of uncertainty estimates in deep learning models makes it difficult to distinguish between hard and easy samples. To address this challenge, we present a computationally efficient approach for post-hoc uncertainty quantification in dynamic neural networks. We show that adequately quantifying and accounting for both aleatoric and epistemic uncertainty through a probabilistic treatment of the last layers improves the predictive performance and aids decision-making when determining the computational budget. In the experiments, we show improvements on CIFAR-100, ImageNet, and Caltech-256 in terms of accuracy, capturing uncertainty, and calibration error.
\end{abstract}

\section{Introduction}
\label{sec:intro}
The ability to scale deep neural networks up to millions of parameters (\eg,~\cite{xu2018scaling,strubell2019energy,liu2021swin}) on massive data sets (\eg,~\cite{deng2009imagenet,lin2014coco}) has been a crucial part of achieving impressive performance, sometimes exceeding human experts on many natural-language processing and computer vision tasks. However, learning and deploying such models entails high computational costs and becomes increasingly difficult~\cite{xu2018scaling,strubell2019energy}, especially as inference costs arise for every deployed instance and can heavily add up~\cite{schwartz2020right}.\looseness-1

\begin{figure}[t]
\centering
\resizebox{\columnwidth}{!}{\input{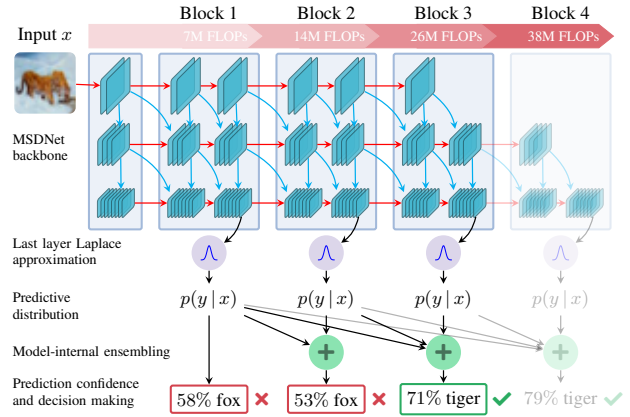}}
\vspace{-1.6em}
  \caption{Increasing depth/scale of a dynamic neural network. Our probabilistic decision-making captures a calibrated confidence estimate, allowing better decisions from the intermediate classifiers on when to stop the evaluation (decision on whether to output the prediction is shown in red and green outline).}
  \label{fig:architecture}
 \vspace*{-1.2em}
\end{figure}

Henceforth, there has been an increased interest in techniques for energy-efficient inference in deep learning~\cite{goel2020lowpower,alyamkin2019lowpower}. A particularly promising direction is to \emph{dynamically} select the most cost-efficient sub-model based on the \emph{difficulty} of the test sample. Dynamic neural networks (DNNs, \cite{huang2018multi,han2021dynamic}) leverage a multi-exit architecture and \emph{dynamically} route test samples based on their difficulty. For example, an image of a sunflower might be easy to classify, requiring less compute, while a tiger in a snowy environment might be more ambiguous and can only be correctly identified after several stages (\cf~\cref{fig:architecture}). The adaptive early exiting is typically based on the confidence scores at the individual exits or learned gating functions~\cite{han2021dynamic}. Thus, it is crucial that the predictive densities or gating functions are robust and allow for trustworthy decision-making. However, current approaches are problematic as typical neural architectures are: {\em (i)}~\textbf{miscalibrated}~\cite{guo2017calibration}, {\em (ii)}~\textbf{overconfident}~\cite{hein2019relu}, and {\em (iii)}~their predictions do not capture \textbf{epistemic} uncertainties~(uncertainty about the true model, see \cite{Kendall2017}).\looseness-1

Uncertainty quantification is the interest of probabilistic (or \emph{Bayesian}) methods in deep learning. Bayesian deep learning \cite{Wilson:ensembles} has recently gained increasing attention in the machine learning community as a means for uncertainty quantification (\eg,~\cite{Kendall2017,wilson2020bayes}) and model selection (\eg,~\cite{immer2021marginal,antoran2022marginal}), compromising, among others, advancements on prior specification (\eg,~\cite{meronen2020stationary,meronen2021periodic,Fortuin2021,nalisnick2018do}) and efficient approximate inference schemes (\eg,~\cite{laplace2021,maddox19_SWAG,Lakshminarayanan17_deep_ensembles}). Even though some of these advancements have recently found application in computer vision (\eg,~\cite{roy2022domain,wang2022bayes,sun2022bayes}), they have not found adoption for decision-making in DNNs.\looseness-1

In this work, we propose a new probabilistic treatment of the multiple exits of DNNs by applying a Bayesian formulation combined with efficient post-hoc approximate inference using multiple last-layer Laplace approximations~\cite{pmlr-v119-kristiadi20a}. Our probabilistic treatment reduces \emph{overconfidence}, improves \emph{calibration}, and captures \emph{epistemic} uncertainties arising through the uncertainty about the true model without increasing the computational burden significantly. Moreover, we propose aggregating predictions across the exits to utilise uncertainties arising at earlier stages and show that accounting for uncertainties in decision-making (bottom in \cref{fig:architecture}) outperforms the standard dynamic neural network (MSDNet~\cite{huang2018multi}) on CIFAR-100, ImageNet, and Caltech-256.\looseness-1

Our contributions can be summarized as follows:
  {\em (i)}~We introduce a probabilistic treatment for early exit dynamic neural networks (DNNs) utilising a Bayesian formulation.
  {\em (ii)}~We present a computationally efficient post-hoc approach for uncertainty-aware decision-making by leveraging our efficient last-layer Laplace approximation implementation combined with model-internal ensembling, which works out of the box without retraining. 
  {\em (iii)}~Finally, we show on CIFAR-100, ImageNet, and Caltech-256 that our probabilistic treatment improves over a conventional approach in accuracy, capturing uncertainties, and calibration error.

\subsection{Related work}
\label{sec:related}
\noindent\textbf{Dynamic neural networks} (DNNs, \cite{han2021dynamic,laskaridis2021adaptive,scardapane2020should}) that utilise intermediate classifiers, allow early exit of easy samples during inference. This tailors the computation depth of each input sample at runtime and offers complementary performance gains to other efficiency optimisations. The early works developing the chain-structured models show limited performance due to the interference among different classifiers~\cite{teerapittayanon2016branchynet,kaya2019shallow}, which is addressed by the proposed Multi-Scale DenseNet (\mbox{MSDNet}, \cite{huang2018multi}) via dense connections and a multi-scale structure. Moreover, the MSDNet is further improved from the aspects of reducing resolution redundancy \cite{yang2020resolution} and training process~\cite{li2019improved,phuong2019distillation}. However, the confidence-based early exiting criterion applied in these aforementioned works can be problematic, and the generated confidence does not necessarily reflect the complexity of the input. Furthermore, the poor estimation of uncertainty in these models makes it difficult to decide which samples are easy and which are hard, and further vulnerable to the slow-down attacks in \cite{hong2020panda}. Despite these shortcomings, multi-exit models have been successfully applied for image segmentation \cite{kouris2022multi}, image caption prediction \cite{fei2022deecap}, and to natural language processing by implementing early exits into a BERT model \cite{schwartz2020right,devlin2018bert}.\looseness-1

\smallskip
\noindent\textbf{Bayesian deep learning} \cite{Neal:1995,Kendall2017} allows including prior knowledge and assumptions into deep learning models, and formulates their predictions through a probabilistic treatment. Calculating the posterior distribution of a Bayesian neural network is usually intractable, and approximate inference techniques need to be used, such as variational inference \cite{David_inference_2017}, deep ensembles \cite{Lakshminarayanan17_deep_ensembles}, MC dropout \cite{Gal+Ghahramani:2016}, or Laplace approximation \cite{ritter2018a_kfac_laplace,pmlr-v119-kristiadi20a,immer2021improving}---each having strengths and weaknesses. A key guiding principle in this work is constraining the computational budget, which leads us to propose an approach that utilizes computationally light Laplace approximations and re-uses predictions of DNN intermediate classifiers in a model-internal ensemble. Prior work does not consider the overconfidence or calibration of confidence estimates that DNNs use to make decisions on which samples require more computational budget.\looseness-1

\section{Background}
\label{sec:background}
We are concerned with image classification under computational budget restrictions subject to a labelled training data set, $\mathcal{D}_\text{train}=\{ \vx_i, \vy_i\}_{i=1}^{n_\text{train}}$, where $\vx_i$ is $d$-dimensional input (\eg, RGB image with $d = 3\times N_\text{pixels}$). Labels $\vy_i$ are $c$-dimensional one-hot encoded vectors indicating the correct class label in the classification task. The computational budget restrictions are applied in the form of a budgeted batch classification setup, in which a fixed computational budget~$B$---measured in average floating point operations per input sample (FLOPs)---must be distributed across a batch of test samples to achieve the highest possible accuracy. Here, a model is trained on the training set $\mathcal{D}_\text{train}$ with an `unlimited' computational budget and tested on a set of test samples $\mathcal{D}_\text{test}=\{ \vx_j, \vy_j\}_{j=1}^{n_\text{test}}$ using a \emph{limited} average budget $B$ per test sample.\looseness-1

For a DNN model having $n_\text{block}$ intermediate classifiers (see \cref{fig:architecture} for a sketch), we refer to the predictive distribution of each of these classifiers as $p_k(\hat{\vy}_i \mid \vx_i), k=1,2,\ldots, n_\text{block}$. The feature representation before the last linear layer of each classifier is referred to as $\vphi_{i,k} = f_k(\vx_i)$, and the parameters of the last linear layer are $\vtheta_k = \{\MW_k, \vb_k \}$. The prediction $p_k(\hat{\vy}_i \mid \vx_i)$ is obtained from the feature representation $\vphi_{i,k}$ as follows: $p_k(\hat{\vy}_i \mid \vx_i) = \softmax(\hat{\vz}_{i,k} )$, where $\hat{\vz}_{i,k} = \MW_k\vphi_{i,k} + \vb_k$.\looseness-1

\begin{figure*}
  \centering\scriptsize
  \pgfplotsset{axis on top}
  \newcommand\upperlefttrue{{"couch","rocket","cattle","bed","tulip","palm tree","seal","butterfly","seal","sweet pepper","bowl","forest","lion","lobster","bottle"}}
  \newcommand\upperleftpred{{"television","cloud","camel","wardrobe","orchid","house","otter","aquarium fish","snail","orange","orange","mountain","tractor","spider","tractor"}}
  \newcommand\lowerlefttrue{{"tank","palm tree","orange","bottle","bottle","orchid","sunflower","palm tree","bicycle","trout","road","bicycle","bottle","palm tree","lawn mower"}}
  \newcommand\lowerleftpred{{"tank","palm tree","orange","bottle","bottle","orchid","sunflower","palm tree","bicycle","trout","road","bicycle","bottle","palm tree","lawn mower"}}
  \newcommand\righttrue{{"porcupine","bear","rose","worm","caterpillar","worm","lobster","flatfish","seal","porcupine","man","tulip","rose","squirrel","beetle"}}
  \newcommand\rightpred{{"crab","cup","road","seal","crab","bottle","snake","flatfish","lamp","squirrel","palm tree","butterfly","lobster","snail","aquarium fish"}}
  \def\blockwidth{.16\textwidth}
  \newcommand{\imageplotnew}[2]{\node[draw=white,fill=black!20,minimum width=\blockwidth,minimum height=\blockwidth,inner sep=0pt] at (0,-#2*\blockwidth) {\includegraphics[width=\blockwidth]{./img/cifar100/#1/im#2.png}};}
  \setlength{\figurewidth}{.24\textwidth}
  \setlength{\figureheight}{\figurewidth}  
  \begin{minipage}[b]{.30\textwidth}
  \pgfplotsset{axis on top,scale only axis,width=\figurewidth,height=\figureheight,
    y tick label style={rotate=90}
  }
  \raggedleft
  \begin{tikzpicture}
    \node (img) {
\begin{tikzpicture}

\begin{axis}[
height=\figureheight,
tick align=outside,
tick pos=left,
title={Block 4: accuracy 73.05\%},
width=\figurewidth,
x grid style={white!69.0196078431373!black},
xlabel={Uncertainty (entropy)},
xmajorgrids,
xmin=0, xmax=4,
xtick style={color=black},
y grid style={white!69.0196078431373!black},
ylabel={Error},
ymajorgrids,
ymin=0, ymax=1,
ytick style={color=black},
xtick={0,2,4},
xticklabels={0,2,4},
ytick={0,0.5,1},
yticklabels={0,0.5,1}
]
\addplot graphics [includegraphics cmd=\pgfimage,xmin=-0.785263656132623, xmax=4.11583719516165, ymin=-0.207140687959535, ymax=1.2214260573898] {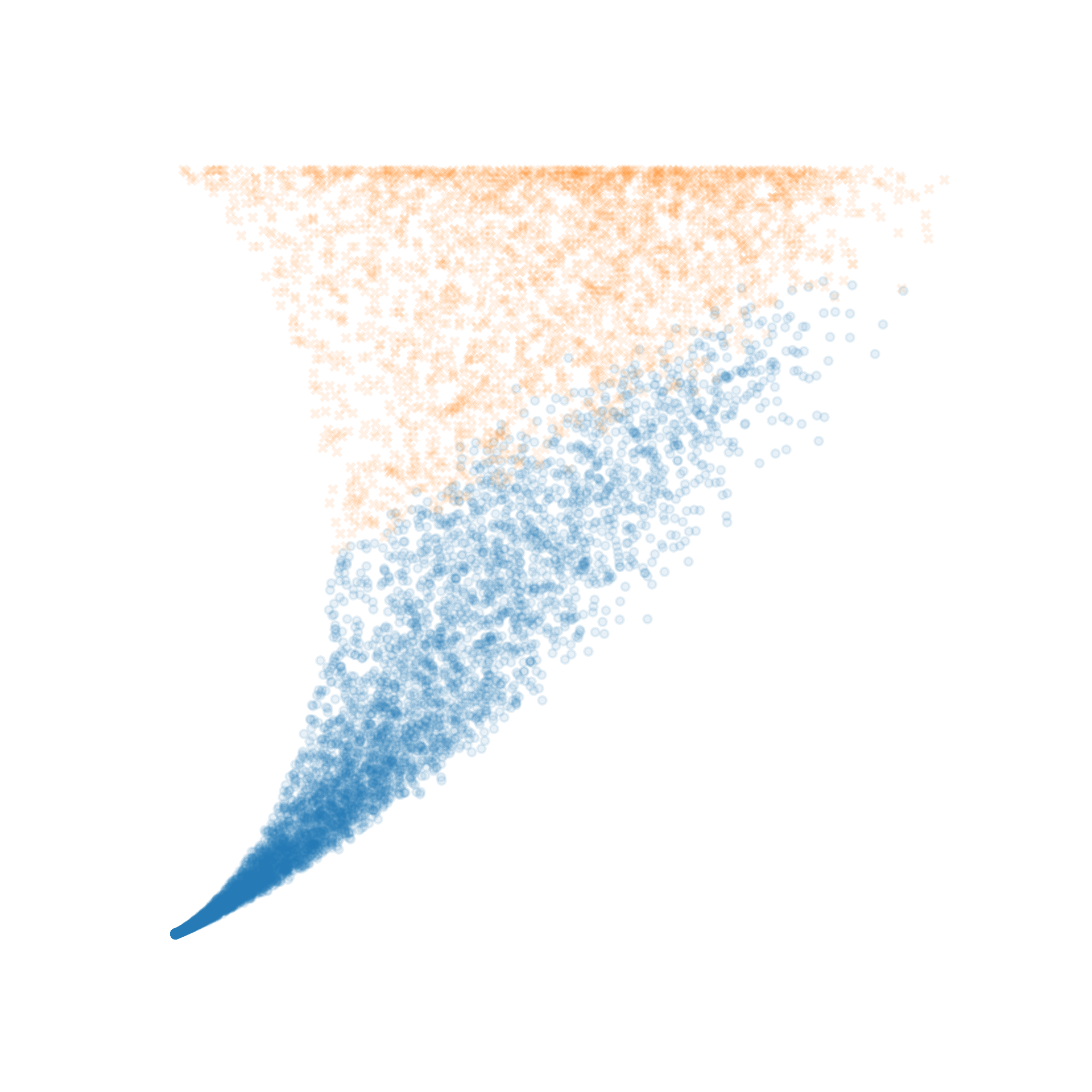};
\end{axis}

\end{tikzpicture}};
    \node [circle, draw=mylcolor0,align=center] at (-1.3,1.8){\normalsize(c)};
    \node [circle, draw=mylcolor0,align=center] at (2.0, 1.8){\normalsize(b)};
    \node [circle, draw=mylcolor0,align=center] at (-1.3, -1.5){\normalsize(a)};
    \end{tikzpicture}\\[0em]
    \begin{tikzpicture}    
    \draw[fill=white,draw=black!80,rounded corners=2pt] (-.45\figurewidth,.2cm) rectangle (.25\figurewidth,1.3cm);
    \node[anchor=west] at (-.4\figurewidth,.5cm) {\includegraphics[width=.7em]{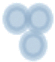}~Correctly classified};    
    \node[anchor=west] at (-.4\figurewidth,1cm) {\includegraphics[width=.7em]{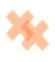}~Misclassified};
    \end{tikzpicture}\hspace*{.75cm}~
  \end{minipage}
  \hfill
  \setlength{\figurewidth}{.22\textwidth} 
  \setlength{\figureheight}{.55\figurewidth}  
  \begin{minipage}[b]{.22\textwidth}
    \centering
    \def\note{\normalsize(a)}
    \begin{tikzpicture}
    \node[font={\scriptsize\it}] at (0,.3) {The model is right and certain of it};
    \node at (0,1.5cm) {\input{fig/gaussian-4.tex}};
    \draw[fill=gray!10,draw=none,rounded corners=2pt] (-.5\figurewidth,0) rectangle (.5\figurewidth,-7.5*\blockwidth);
    \node[font={\it\scriptsize},align=left,text width=2cm] at (-.7cm,1.5cm) {`easy'};
    \node[text width=1.3cm,align=right,anchor=east,font=\bf] at (-.6*\blockwidth,-.3*\blockwidth) {True~\ref{plot:star}~};
    \node[text width=1.3cm,align=left,anchor=west,font=\bf] at (.6*\blockwidth,-.3*\blockwidth) {Predicted};
    \foreach \i [count=\j] in {1,...,7} {
      \imageplotnew{lower_left}{\i} 
      \node[text width=1.35cm,align=right,anchor=east] at (-.6*\blockwidth,-\i*\blockwidth) {\pgfmathparse{\lowerlefttrue[\i-1]}\pgfmathresult};
      \node[text width=1.4cm,align=left,anchor=west] at (.6*\blockwidth,-\i*\blockwidth) {\pgfmathparse{\lowerleftpred[\i-1]}\pgfmathresult};
    }
    \end{tikzpicture}
  \end{minipage}
  \hfill
  \begin{minipage}[b]{.22\textwidth}
    \centering
    \def\note{\normalsize(b)}
    \begin{tikzpicture}
    \node[font={\scriptsize\it}] at (0,.3) {The model is uncertain and knows it};    
    \node at (0,1.5cm) {\input{fig/gaussian-1.tex}};
    \draw[fill=gray!10,draw=none,rounded corners=2pt] (-.5\figurewidth,0) rectangle (.5\figurewidth,-7.5*\blockwidth);
    \node[font={\it\scriptsize},align=left,text width=2cm] at (-.7cm,1.5cm) {`hard'};    
    \node[text width=1.3cm,align=right,anchor=east,font=\bf] at (-.6*\blockwidth,-.3*\blockwidth) {True~\ref{plot:star}~};
    \node[text width=1.3cm,align=left,anchor=west,font=\bf] at (.6*\blockwidth,-.3*\blockwidth) {Predicted};  
    \foreach \i [count=\j] in {1,...,7} {
      \imageplotnew{right}{\i} 
      \node[text width=1.35cm,align=right,anchor=east] at (-.6*\blockwidth,-\i*\blockwidth) {\pgfmathparse{\righttrue[\i-1]}\pgfmathresult};
      \node[text width=1.4cm,align=left,anchor=west] at (.6*\blockwidth,-\i*\blockwidth) {\pgfmathparse{\rightpred[\i-1]}\pgfmathresult};
    }
    \end{tikzpicture}
  \end{minipage}
  \hfill
  \begin{minipage}[b]{.22\textwidth}
    \centering
    \def\note{\normalsize(c)}
    \begin{tikzpicture}[outer sep=0]
    \node[font={\scriptsize\it}] at (0,.3) {The model is wrong but overly certain};        
    \node at (0,1.5cm) {\input{fig/gaussian-2.tex}};
    \draw[fill=gray!10,draw=none,rounded corners=2pt] (-.5\figurewidth,0) rectangle (.5\figurewidth,-7.5*\blockwidth);
    \node[font={\it\scriptsize},align=left,text width=2cm] at (-.7cm,1.4cm) {`ambiguous' \\ `tricky'};
    \node[text width=1.3cm,align=right,anchor=east,font=\bf] at (-.6*\blockwidth,-.3*\blockwidth) {True~\ref{plot:star}~};
    \node[text width=1.3cm,align=left,anchor=west,font=\bf] at (.6*\blockwidth,-.3*\blockwidth) {Predicted};   
    \foreach \i [count=\j] in {1,...,7} {
      \imageplotnew{upper_left}{\i} 
      \node[text width=1.35cm,align=right,anchor=east] at (-.6*\blockwidth,-\i*\blockwidth) {\pgfmathparse{\upperlefttrue[\i-1]}\pgfmathresult};
      \node[text width=1.4cm,align=left,anchor=west] at (.6*\blockwidth,-\i*\blockwidth) {\pgfmathparse{\upperleftpred[\i-1]}\pgfmathresult};
    }
    \end{tikzpicture}
  \end{minipage}
  \vspace{-1em}
  \caption{\textbf{Knowing what the model does not know.} An uncertainty--error scatter plot showing CIFAR-100 test set predictions for the last classifier of an MSDNet with Laplace and model-internal ensembling, and example images of three types corresponding to three areas in the uncertainty--error scatter plot. The error is defined as $1-p(\hat\vy=\vy)$, where $p(\hat\vy=\vy)$ is the predicted probability on the correct label.\looseness-1}
  \label{fig:uncertainty_error_examples}
\end{figure*}

Our focus is on fixing the overconfidence of DNN architectures that dynamically adapt the network depth utilising early exiting by intermediate classifiers, each with increasing computational requirements. Implementing classifiers as early exits into a single network allows reusing computation and reduces the overall inference cost compared to using independent models of varying sizes. Early versions of such DNNs suffer from interference between classifiers during training \cite{teerapittayanon2016branchynet,kaya2019shallow}. This problem can be alleviated by using dense connectivity between intermediate classifiers, and by utilising a multi-scale structure having fine and coarse-scale features throughout the network. This architecture is referred to as a Multi-Scale DenseNet (\mbox{MSDNet}, \cite{huang2018multi}), and we use it as a backbone to demonstrate our methods of improving uncertainty estimation, as at the time of writing it clearly outperforms other image classification DNNs that use intermediate classifiers. However, our methods are applicable to any DNN that utilises early exiting. \cref{fig:architecture} visualizes the structure of the DNN backbone and the applied uncertainty quantification methods. The visualized FLOPs numbers are for the `small' MSDNet model used for CIFAR-100 (see \cref{sec:experiments})\looseness-1

The challenge of efficiently using a DNN backbone with intermediate classifiers is the decision-making problem of when to exit the model. The goal is to use less capacity for clear `easy' samples while unleashing more capacity for more difficult or tricky cases. To achieve this goal, the model must be well-calibrated and know when it is uncertain about its predictions. Many DNN models (\eg \cite{teerapittayanon2016branchynet, huang2018multi}) use predicted confidences from intermediate classifiers to make decisions on early exiting, but the calibration of these predictions is not controlled.

\smallskip
\noindent
\textbf{Uncertainty estimation in deep learning} is often divided into estimating two different types of uncertainty \cite{Kendall2017}: \emph{aleatoric} and \emph{epistemic}. Aleatoric uncertainty is related to randomness intrinsic to the task at hand and cannot be reduced. Epistemic uncertainty is related to our knowledge of the task and can be reduced by learning more about the task, \eg, by obtaining more data. In our problem setting of image classification, epistemic uncertainty is related to the model parameters. In \cref{fig:uncertainty_error_examples}, epistemic uncertainty is present in the predictions for images of type (b): the model has not learned the task well enough to classify these difficult sample images correctly. On the other hand, an example of aleatoric uncertainty is seen in some of the sample images of type (c): an image may contain objects from multiple classes, and the most prominent object is not necessarily labelled as the correct class (`ambiguous'), or some images may be completely mislabelled. We refer to this kind of samples as `tricky'.\looseness-1

A Bayesian treatment to uncertainty estimation means that instead of obtaining a single point estimate of the model parameters $\vtheta$ as the result of neural network training, Bayesian inference is used to obtain a \emph{posterior distribution} over the model parameters given the training data $\mathcal{D}_\text{train}$:
\begin{equation}\label{eq:bayes}
p(\vtheta\mid\mathcal{D}_\text{train}) = \frac{p(\mathcal{D}_\text{train}\mid\vtheta)\,p(\vtheta)}{\int_{\vtheta} p(\mathcal{D}_\text{train}, \vtheta) \dd \vtheta} = \frac{\text{\small[likelihood]}{\small\times}\text{\small[prior]}}{\text{\small[model evidence]}}.
\end{equation}
Usually, calculating the exact posterior for a deep learning model is intractable. This means that the posterior of the model parameters must be approximated. One approach is to consider the $L-1$ first layers of a deep neural network with $L$ layers as a fixed feature extractor and limit the Bayesian treatment to the last ($L$\textsuperscript{th}) layer. This drastically decreases the number of parameters for which the posterior distribution needs to be estimated. However, computations are usually still infeasible as no analytic solution exists, and further approximations are needed.\looseness-1

An efficient approximation to the posterior of the model parameters is the Laplace approximation, which performs a second-order Taylor expansion of \cref{eq:bayes}~around the maximum {\it a~posteriori} (MAP) estimate of the target distribution, resulting in a Gaussian distribution. The model parameters representing the MAP estimate can be found by maximising the unnormalised posterior: $p(\vtheta \mid \mathcal{D}) \propto p(\mathcal{D}_\text{train}\mid\vtheta)\,p(\vtheta) = p(\vtheta, \mathcal{D})$, which is commonly assumed in log-space for numerical stability: $\log p(\vtheta, \mathcal{D}_\text{train}) = \log p(\mathcal{D}_\text{train}\mid\vtheta) + \log p(\vtheta)$. In classification tasks, we typically minimise the cross-entropy loss, which is equivalent to maximising the log-likelihood. Moreover, commonly used regularisation methods such as weight decay can be interpreted as a log-prior. Hence, deep learning models learned with conventional training methods for classification tasks can be seen as maximising the unnormalised log-posterior and we, therefore, \textbf{directly obtain the MAP estimate} ($\vtheta_{\text{MAP}}$) required for Laplace approximations through standard training. The Laplace approximation of the posterior is then formed using a multivariate Gaussian centred at the MAP estimate with covariance given by the inverse of the Hessian $\MH$ of the negative log-posterior, \ie, $\N(\vtheta_{\text{MAP}} \mid \MH^{-1})$ and $\MH \coloneqq - \nabla^{2}_{\vtheta} \log p(\vtheta \mid \mathcal{D}) \mid_{\vtheta_{\text{MAP}}}$. The Hessian can be efficiently approximated using the generalised Gauss-Newton algorithm \cite{pmlr-v70-botev17a} or by Kronecker factorisation \cite{pmlr-v37-martens15,ritter2018a_kfac_laplace}.

\section{Methods}
\label{sec:methods}
Our aim is to couple an early exit DNN backbone with an uncertainty-aware decision-making process. For uncertainty quantification, we leverage the early exit structure of the DNN architecture and propose an approach that uses Laplace approximations and model-internal ensembling. The motivation for improving uncertainty estimation is that if the intermediate classifiers in the DNN can more accurately estimate the uncertainty in their predictions, they can better recognise which samples are hard and require further computation, and for which samples the prediction is already confident enough to be exited early.\looseness-1

\subsection{Laplace approximation of a dynamic NN}
\label{sec:laplace}
\begin{figure}[!t]
\definecolor{color0}{rgb}{1,0.498039215686275,0.0549019607843137}

  \centering\scriptsize
  \pgfplotsset{xlabel={Exit number},axis on top,scale only axis,width=\figurewidth,height=\figureheight,
    tick label style={font=\tiny},
    x tick scale label style={yshift=.75em},
    tick scale binop=\times,
    xtick={}
  }
  \setlength{\figurewidth}{.4\columnwidth}
  \setlength{\figureheight}{0.59\figurewidth} 
  \begin{subfigure}[t]{.49\columnwidth}
    \raggedleft
\begin{tikzpicture}

\definecolor{color0}{rgb}{1,0.498039215686275,0.0549019607843137}

\begin{axis}[
height=\figureheight,
tick align=outside,
tick pos=left,
width=\figurewidth,
x grid style={white!69.0196078431373!black},
xmin=0.47, xmax=5.53,
xtick style={color=black},
y grid style={white!69.0196078431373!black},
ylabel={FLOPs},
ymin=0, ymax=3453468340,
ytick style={color=black},
xtick = {1,2,3,4,5},
xticklabels = {1,2,3,4,5}
]
\draw[draw=none,fill=color0] (axis cs:1,0) rectangle (axis cs:0.7,339900000);
\draw[draw=none,fill=color0] (axis cs:2,0) rectangle (axis cs:1.7,685460000);
\draw[draw=none,fill=color0] (axis cs:3,0) rectangle (axis cs:2.7,1008160000);
\draw[draw=none,fill=color0] (axis cs:4,0) rectangle (axis cs:3.7,1254470000);
\draw[draw=none,fill=color0] (axis cs:5,0) rectangle (axis cs:4.7,1360530000);
\draw[draw=none,fill=black] (axis cs:1,0) rectangle (axis cs:1.3,340148610);
\draw[draw=none,fill=black] (axis cs:2,0) rectangle (axis cs:2.3,685957220);
\draw[draw=none,fill=black] (axis cs:3,0) rectangle (axis cs:3.3,1008834886);
\draw[draw=none,fill=black] (axis cs:4,0) rectangle (axis cs:4.3,1255243320);
\draw[draw=none,fill=black] (axis cs:5,0) rectangle (axis cs:5.3,1362606410);
\draw[draw=none,fill=black,fill opacity=0.5] (axis cs:1,0) rectangle (axis cs:1.3,725431173.333333);
\draw[draw=none,fill=black,fill opacity=0.5] (axis cs:2,0) rectangle (axis cs:2.3,1456522346.66667);
\draw[draw=none,fill=black,fill opacity=0.5] (axis cs:3,0) rectangle (axis cs:3.3,2164682768);
\draw[draw=none,fill=black,fill opacity=0.5] (axis cs:4,0) rectangle (axis cs:4.3,2796374213.33333);
\draw[draw=none,fill=black,fill opacity=0.5] (axis cs:5,0) rectangle (axis cs:5.3,3289017466.66667);
\end{axis}

\end{tikzpicture}\\[-0.3em]
    \caption{\footnotesize ImageNet small model}
  \end{subfigure}
  \hfill
  \begin{subfigure}[t]{.49\columnwidth}
    \raggedleft
    \pgfplotsset{ybar,legend image code/.code={\draw [#1,draw=none] (0cm,-0.08cm) rectangle (0.15cm,0.15cm);}}
\begin{tikzpicture}

\definecolor{color0}{rgb}{1,0.498039215686275,0.0549019607843137}

\begin{axis}[
height=\figureheight,
legend cell align={left},
legend style={
  font=\tiny,
  fill opacity=0.8,
  draw opacity=1,
  text opacity=1,
  at={(0.03,0.97)},
  anchor=north west,
  draw=white!80!black
},
tick align=outside,
tick pos=left,
width=\figurewidth,
x grid style={white!69.0196078431373!black},
xmin=0.47, xmax=5.53,
xtick style={color=black},
y grid style={white!69.0196078431373!black},
ymin=0, ymax=5444748148,
ytick style={color=black},
xtick = {1,2,3,4,5},
xticklabels = {1,2,3,4,5}
]
\draw[draw=none,fill=color0] (axis cs:1,0) rectangle (axis cs:0.7,615600000);
\addlegendimage{ybar,draw=none,fill=color0};
\addlegendentry{Vanilla MSDNet}

\draw[draw=none,fill=color0] (axis cs:2,0) rectangle (axis cs:1.7,1436390000);
\draw[draw=none,fill=color0] (axis cs:3,0) rectangle (axis cs:2.7,2283210000);
\draw[draw=none,fill=color0] (axis cs:4,0) rectangle (axis cs:3.7,2967420000);
\draw[draw=none,fill=color0] (axis cs:5,0) rectangle (axis cs:4.7,3253790000);
\draw[draw=none,fill=black] (axis cs:1,0) rectangle (axis cs:1.3,616033506);
\addlegendimage{ybar,draw=none,fill=black};
\addlegendentry{Efficient Laplace}

\draw[draw=none,fill=black] (axis cs:2,0) rectangle (axis cs:2.3,1437413044);
\draw[draw=none,fill=black] (axis cs:3,0) rectangle (axis cs:3.3,2284624470);
\draw[draw=none,fill=black] (axis cs:4,0) rectangle (axis cs:4.3,2968941480);
\draw[draw=none,fill=black] (axis cs:5,0) rectangle (axis cs:5.3,3259067530);
\draw[draw=none,fill=black,fill opacity=0.5] (axis cs:1,0) rectangle (axis cs:1.3,1001315685.33333);
\addlegendimage{ybar,draw=none,fill=black,fill opacity=0.5};
\addlegendentry{Na\"ive Laplace}

\draw[draw=none,fill=black,fill opacity=0.5] (axis cs:2,0) rectangle (axis cs:2.3,2207977146.66667);
\draw[draw=none,fill=black,fill opacity=0.5] (axis cs:3,0) rectangle (axis cs:3.3,3440470816);
\draw[draw=none,fill=black,fill opacity=0.5] (axis cs:4,0) rectangle (axis cs:4.3,4510070581.33333);
\draw[draw=none,fill=black,fill opacity=0.5] (axis cs:5,0) rectangle (axis cs:5.3,5185474426.66667);
\end{axis}

\end{tikzpicture}\\[-0.3em]
    \caption{\footnotesize ImageNet large model}
  \end{subfigure}\\
  \vspace{-.8em}
  \caption{Test time computational cost of \protect\tikz[baseline]\fill[black,fill opacity=0.5](-0.05,-0.05) rectangle (0.1,0.2);~\naive and \protect\tikz[baseline]\fill[black](-0.05,-0.05) rectangle (0.1,0.2);~efficient methods of sampling the Laplace predictive distribution (50 MC samples) compared against the \protect\tikz[baseline]\fill[color0](-0.05,-0.05) rectangle (0.1,0.2);~vanilla MSDNet, on ImageNet data. Results shown for each intermediate exit. Details in \cref{app:laplace_cost}.\looseness-1}
\label{fig:imagenet_lap_cost_maintext}
\vspace*{-1em}
\end{figure}
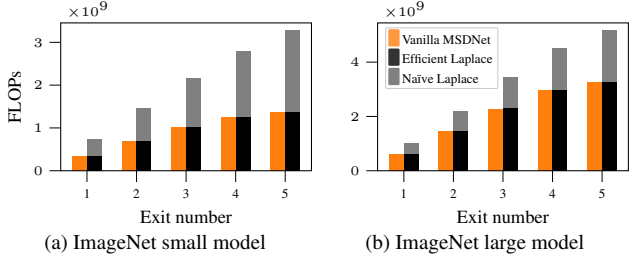
To approximate the predictive posterior we propose an efficient implementation of a last-layer Laplace approximation for each intermediate classifier of the DNN. Using our computationally cheap last-layer approach enables us to stay resource-efficient while at the same time improving the decision-making within DNNs. Note that in this section we have dropped the index $k$ for the intermediate exit for simplicity, as all operations are done independently for each exit.\looseness-1

To save on computational costs of sampling from a larger dimensional Gaussian on the last-layer parameters $\mathrm{N}(\vtheta_{\text{MAP}}, \MH^{-1})$, where $\vtheta_{\text{MAP}} = \{\mbf{W}_{\text{MAP}},\mbf{b}_{\text{MAP}}\}$ are the last-layer MAP parameters of one exit, we linearly project the Gaussian to a predictive distribution $p(\hat{\vz}_{i} \mid \mbf{x}_i)$. This directly allows us to sample pre-softmax outputs $\hat{\vz}_{i}$ and reduces the dimensionality of the Gaussian to the number of classes $c$. The final prediction $\hat{\vy}_{i}$ is then calculated by sampling $n_\text{MC}$ samples from the predictive Gaussian: $\hat{\vy}_{i} = \frac{1}{n_{\text{MC}}}\sum_{l=1}^{n_{\text{MC}}}\text{softmax}(\hat{\vz}_{i}^{(l)})$. If performed na\"ively \cite{pmlr-v119-kristiadi20a}, this forces performing all sampling-related computation at test time, resulting in a per sample cost of $\text{FLOPs}_{\text{\naive}} = 2c^{2}(n_{\text{MC}}+1) + \frac{1}{3}c^{3} + 2p^2 + p -1$ which grows cubically with the number of classes $c$ ($p$ is the feature space dimensionality: $\vphi_i \in \R^p$). Absorbing the last layer biases into the weight matrix allows us to shift most of the computation required for sampling to be performed before test time. The resulting efficient Laplace implementation for DNNs has a cost of $\text{FLOPs}_{\text{efficient}} = 2cn_{\text{MC}} + 2p^2 + 5p + 2$, making Laplace approximation computationally viable in the budget restricted regime. \cref{fig:imagenet_lap_cost_maintext} shows a computational cost comparison for the \naive and efficient Laplace approximations. More detailed analysis is presented in \cref{app:laplace_cost}.\looseness-1

Applying a Laplace approximation only to the last linear layer at each intermediate exit now provides us with a Gaussian distribution $p(\hat{\vz}_{i} \mid \mbf{x}_i) = \mathrm{N}(\hat\MW_{\text{MAP}}^\top \hat\vphi_{i}, (\hat\vphi_{i}^\top \MV \hat\vphi_{i})\MU)$ for each classifier. Here $ \MV^{-1} \otimes \MU^{-1}=\MH^{-1}$ is an approximate inverse Hessian, being a Kronecker factorisation of the generalised Gauss--Newton matrix and $\hat\vphi_{i} = (\vphi_{i}^{\top}, 1)^{\top}$ are the features after augmenting the biases into the weights. Samples from this distributions are calculated as $\hat{\vz}_{i}^{(l)} =  \hat\MW_{\text{MAP}}^\top \hat\vphi_{i} + (\hat\vphi_{i}^\top \MV \hat\vphi_{i})^{\frac{1}{2}} (\ML\vg^{(l)})$, where $\vg^{(l)} \sim \mathrm{N}(\bm{0},\MI)$ and $\ML$ is the Cholesky factor of $\MU$.\looseness-1

To ensure well calibrated predictions from the Laplace approximation, it is useful to utilise temperature scaling \cite{guo2017calibration} on the sampled predictions $\hat{\vy}_i$. In practice, this means dividing each sampled pre-softmax output $\hat{\vz}_{i}^{(l)}$ with a temperature scaling parameter $T$ before taking the softmax. Also the Laplace prior variance $\sigma$ is a hyperparameter affecting the results. We use a different value of $T$ and $\sigma$ for each classifier in the network. To choose appropriate values for the temperature scaling and prior variance parameters, we perform a grid search over possible pairs of values of $T$ and $\sigma$, selecting the pair of values that minimises the negative log-predictive density score on the validation set for the classifier in question. Details on this grid search are in \cref{sec:app_details}.

\begin{figure}
  \centering\scriptsize
  \pgfplotsset{axis on top,scale only axis,width=\figurewidth,height=\figureheight}
  \setlength{\figurewidth}{0.9\columnwidth}
  \setlength{\figureheight}{0.3\figurewidth} 
  \begin{subfigure}[t]{\columnwidth}
    \raggedleft
    \pgfplotsset{axis x line*= bottom, axis y line*= left}
    \pgfplotsset{y axis line style={draw=none,draw opacity=0}}
   Samples in histogram (error > 0.5): Our model: 36.53\%, Vanilla model: 33.33\%
   \vspace*{-0.2em}
    \input{fig/fig3_histogram.tex}
  \end{subfigure}\\
  \setlength{\figurewidth}{0.35\columnwidth}
  \setlength{\figureheight}{\figurewidth} 
  \begin{subfigure}[t]{.48\columnwidth}
    \centering
    \pgfplotsset{ylabel={Error},ylabel style={yshift=6pt},xlabel style={yshift=-6pt}}
    \begin{tikzpicture}
    \node[outer sep=0,inner sep=0]{
\begin{tikzpicture}

\begin{axis}[
height=\figureheight,
tick align=outside,
tick pos=left,
width=\figurewidth,
x grid style={white!69.0196078431373!black},
xlabel={Uncertainty (entropy)},
xmajorgrids,
xmin=0, xmax=4,
xtick style={color=black},
y grid style={white!69.0196078431373!black},
ymajorgrids,
ymin=0, ymax=1,
ytick style={color=black},
xtick={0,2,4},
xticklabels={0,2,4},
ytick={0,0.5,1},
yticklabels={0,0.5,1}
]
\addplot graphics [includegraphics cmd=\pgfimage,xmin=-0.873331765642474, xmax=4.57724946673609, ymin=-0.207142857142857, ymax=1.22142857142857] {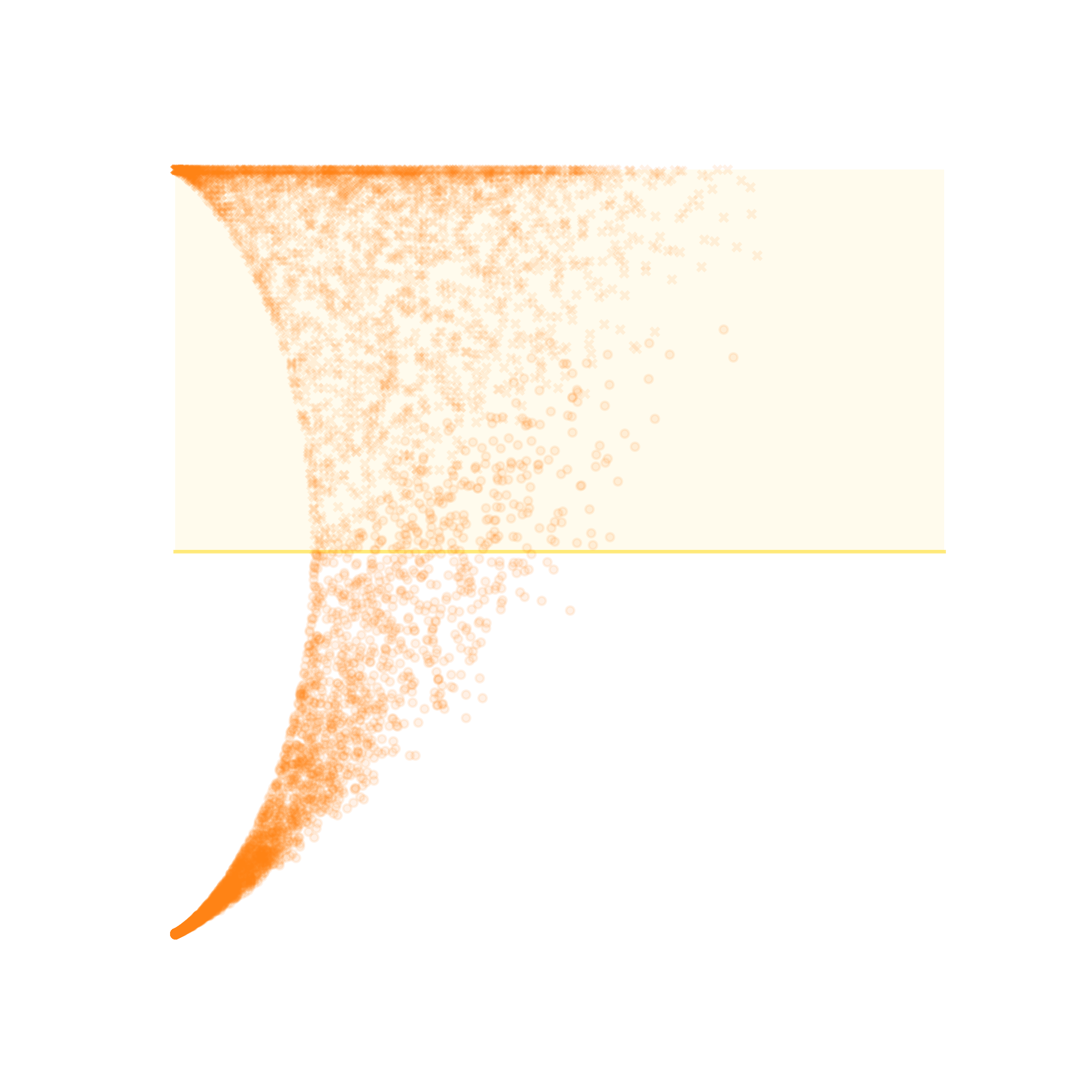};
\end{axis}

\end{tikzpicture}};
    \node[font=\tiny] at (-.75,1.5) {`tricky'};
    \node[font=\tiny] at (1.5,1.5) {`hard'};
    \node[font=\tiny] at (-.75,-.9) {`easy'};
    
    \end{tikzpicture}\\[0.5em]
    \caption{\footnotesize Vanilla MSDNet (ACC: 69.08\%)}
  \end{subfigure}
  \hfill  
  \begin{subfigure}[t]{.48\columnwidth}
    \centering
    \pgfplotsset{ylabel={Error},ylabel style={yshift=6pt},xlabel style={yshift=-6pt}}    
    \begin{tikzpicture}
    \node[outer sep=0,inner sep=0]{
\begin{tikzpicture}

\begin{axis}[
height=\figureheight,
tick align=outside,
tick pos=left,
width=\figurewidth,
x grid style={white!69.0196078431373!black},
xlabel={Uncertainty (entropy)},
xmajorgrids,
xmin=0, xmax=4,
xtick style={color=black},
y grid style={white!69.0196078431373!black},
ylabel={Error},
ymajorgrids,
ymin=0, ymax=1,
ytick style={color=black},
xtick={0,2,4},
xticklabels={0,2,4},
ytick={0,0.5,1},
yticklabels={0,0.5,1}
]
\addplot graphics [includegraphics cmd=\pgfimage,xmin=-0.873331765642474, xmax=4.57724946673609, ymin=-0.207142713240215, ymax=1.22142854503223] {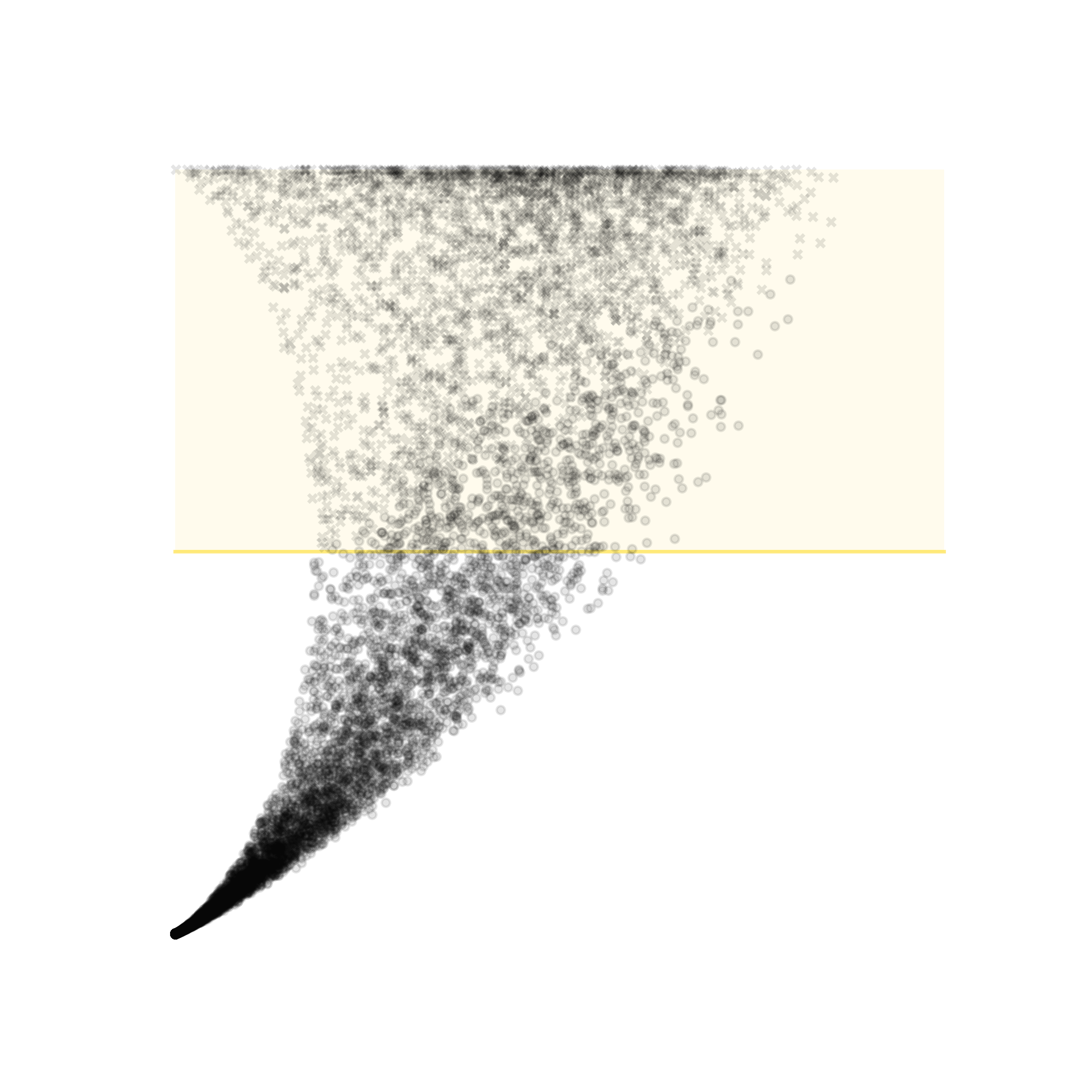};
\end{axis}

\end{tikzpicture}};
    \draw[->,black] (-.75,1.2) -- (1,1.2);
    \end{tikzpicture}\\[0.5em]
    \caption{\footnotesize Our model (ACC: 70.90\%)}    
  \end{subfigure}
  \caption{Comparison of uncertainty--error scatter plots from the \textbf{second to last classifier} of a vanilla MSDNet and our model with Laplace and model-internal ensembling. The uncertainty histogram on the top shows points with error > 0.5: the uncertainty should be high for the model to be able to recognize these samples as `hard', and continue their evaluation to the next block. For our model these samples have a high uncertainty, while the vanilla MSDNet is overconfident. See \cref{sec:explaining_method} for definition of error.}
  \label{fig:uncertainty_error_comparison}
\end{figure}

\subsection{Model-internal ensembling}
\label{sec:ensemble}
To improve the predictive uncertainty and robustness of the DNN predictions, we utilise the idea of ensembling multiple predictions together \cite{Lakshminarayanan17_deep_ensembles}. However, using deep ensembles, where $M$ independent networks are trained, is not feasible in the budget-restricted scenario. Instead, we utilise the predictions of intermediate classifiers in the dynamic neural network to form the ensemble and refer to this as a \emph{model-internal ensemble} (MIE). The predictions from different intermediate classifiers are neither independent nor equal, as they are predictions from different stages of the same computational pipeline, and later classifiers have more capacity compared to earlier classifiers. To account for the difference in capacity in the MIE members, we can scale their influence to the final prediction in proportion to their computational complexity. Predictions from intermediate classifiers for forming an MIE are readily available and require no additional computation, as they need to be calculated to make the decision whether to continue computation further in the network. The model-internal ensemble prediction for the $k$\textsuperscript{th} classifier in an early exit DNN is:\looseness-1
\begin{equation}\label{eq:ensemble}\textstyle
p^\text{ens}_k(\hat{\mbf{y}}_i \mid \mbf{x}_i) = \frac{1}{\sum_{l=1}^{k} w_l} \sum_{m=1}^{k} w_m \, p_m(\hat{\mbf{y}}_i \mid \mbf{x}_i).
\end{equation}
This is a weighted average of intermediate classifiers up to classifier $k$, for which we are calculating the MIE prediction. Later classifiers have more predictions in the average to aggregate, as more already calculated intermediate predictions are available. The weights $w_m$ are the computational costs of the DNN in FLOPs up to classifier $m$. The added computational cost of MIE is marginal (see \cref{app:laplace_cost} for details).\looseness-1

\begin{figure*}[t!]
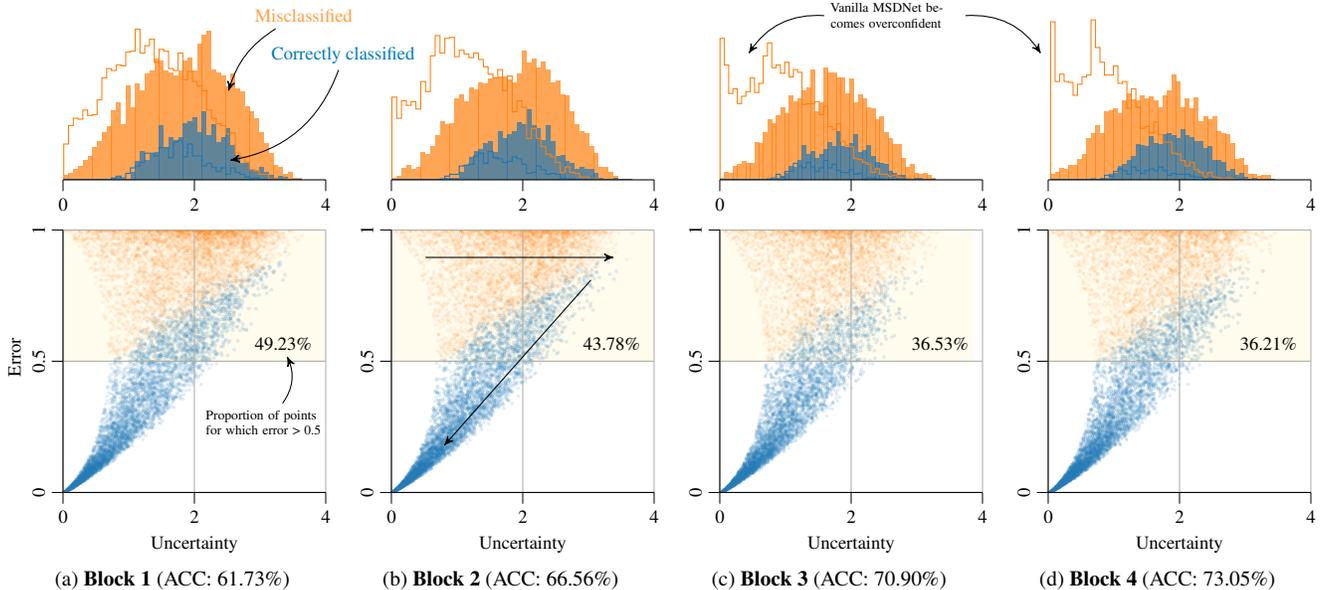

  \centering\scriptsize
  \pgfplotsset{axis on top,scale only axis,width=\figurewidth,height=\figureheight,
    xtick align=outside,
    ytick align=outside,  
    axis x line*= bottom, axis y line*= left,
    y tick label style={rotate=90}}
  \begin{tikzpicture}

    \setlength{\figurewidth}{0.20\textwidth}
    \setlength{\figureheight}{0.7\figurewidth}  

    \begingroup
    
    \pgfplotsset{y axis line style={draw=none,draw opacity=0}}

    \foreach \x [count=\i] in {0,1,2,3} 
      \node[anchor=north] at (\i*0.25*\textwidth,0) {\begin{minipage}{.25\textwidth}\raggedleft{\input{fig/upper_hist_block\x.tex}}\end{minipage}};

    \endgroup

    \setlength{\figurewidth}{0.20\textwidth}
    \setlength{\figureheight}{\figurewidth}  
    \pgfplotsset{xlabel={Uncertainty}}

    \foreach \x [count=\i] in {0,1,2,3}  
      \node[anchor=north] (scatter-\i) at (\i*0.25*\textwidth,-0.17*\textwidth) {\begin{minipage}{.25\textwidth}\raggedleft{\input{fig/errunc_scatter_block\x.tex}}\end{minipage}};

    \foreach \a/\x/\y [count=\i] in {a/49.23/61.73,b/43.78/66.56,c/36.53/70.90,d/36.21/73.05} { 
      \node at (\i*0.25*\textwidth,-0.45*\textwidth) {\footnotesize (\a) \textbf{Block~\i}\ (ACC: \y\%)};     
      \node (perr-\i) at (\i*0.25*\textwidth+6em,-0.27*\textwidth) {\x\%};            
    }
    
    \node[rotate=90] at (0.13*\textwidth,-0.28*\textwidth) {Error};
    
    \node[color=C0] (a) at (0.35*\textwidth,-0.02*\textwidth) {Misclassified};
    \node[color=C1] (b) at (0.38*\textwidth,-0.05*\textwidth) {Correctly classified};
    \draw[->] (a) to[bend right=20] ++(-1cm,-1cm);
    \draw[->] (b) to[bend left=30] ++(-1.5cm,-1.4cm);
    \node[text width=1.55cm,font=\tiny] (c) at (0.32*\textwidth,-0.33*\textwidth) {Proportion of points for which error > 0.5};
    \draw[->] (c) to[bend right=30] (perr-1);
    \node[text width=1.7cm,font=\tiny] (d) at (0.8*\textwidth,-0.02*\textwidth) {Vanilla MSDNet {becomes} overconfident};
    \draw[->] (d.east) to[bend left=30] ++(1cm,-.5cm);
    \draw[->] (d.west) to[bend right=30] ++(-1cm,-.5cm);
        
    \draw[->,draw=black] ($(scatter-2) + (-1cm,1.75cm)$) -- ($(scatter-2) + (1.5cm,1.75cm)$);
    \draw[->,draw=black] ($(scatter-2) + (1.2cm,1.45cm)$) -- ($(scatter-2) + (-.75cm,-.75cm)$);
  
  \end{tikzpicture}
  \vspace*{-2.2em}
  \caption{Model certainty grows with capacity. The bottom scatter plots show \includegraphics[width=.7em]{fig/correct}~correctly classified and \includegraphics[width=.7em]{fig/incorrect}~misclassified test points on an uncertainty vs.\ error axis. The uncertainty of points in the top half is summarized as a histogram above each scatter plot (in scale with each other). Our model consistently adds capacity to calibrate itself to tricky cases (`tricky' $\rightarrow$ `hard') and improves accuracy (`hard' $\rightarrow$ `easy') over the blocks, while the vanilla MSDNet model (solid line in histograms, for reference) does not. \cref{fig:uncertainty_error_app} in \cref{sec:app_results} shows corresponding scatter plots for vanilla MSDNet, and uncertainty histograms including all points in the scatter plots for both models. See \cref{sec:explaining_method} for definition of error.\looseness-1}
\label{fig:uncertainty_error}
\end{figure*}

\subsection{Illustrating the intuition}
\label{sec:explaining_method}
Our model hinges on the realization that properly capturing epistemic uncertainty is key to making informed decisions. This is visible in \cref{fig:uncertainty_error_examples,fig:uncertainty_error_comparison} that demonstrate how better uncertainty quantification can improve the decision-making in a DNN. In \cref{fig:uncertainty_error_examples}, samples of type (c) are ones that the model has predicted with high confidence, but the predicted label is incorrect. These can be mislabelled samples or samples that are very ambiguous. However, for a poorly calibrated model, the type (c) samples may also include a large number of samples that the model predicted overconfidently for no intuitive reason, and that should instead be of type (b): hard samples that the current stage of the model can't accurately classify. This can be seen in \cref{fig:uncertainty_error_comparison} where the standard DNN model has a large number of overconfidently predicted samples in the upper left corner of the scatter plot. In this figure, we can see that the improved calibration of our model allows these samples to move to the upper right corner of the scatter plot. This change prevents these predictions from exiting at the current intermediate classifier and instead allows for potentially improving the prediction in the later steps of the DNN. In figures, the error is defined as $1-p(\hat\vy=\vy)$, where $p(\hat\vy=\vy)$ is the predicted probability on the correct label.\looseness-1

For the decision-making to be efficient in a DNN, each intermediate classifier should have calibrated uncertainties and not show too many samples in the upper left area of the uncertainty--error scatter plots. \cref{fig:uncertainty_error} shows that this is true for our model, and the picture samples in \cref{fig:uncertainty_error_examples} under the label (c) show that the samples remaining in the upper-left corner are mostly ambiguous samples, which even a calibrated model would predict incorrectly but confidently.\looseness-1

\begin{table*}[t!]
  \caption{\small Table of Top-1/Top-5 accuracy, negative log-predictive density (NLPD), and expected calibration error (ECE) for different models on CIFAR-100 and ImageNet data. All numbers are averages over a range of computational budgets in the budgeted batch classification setup. `MIE Laplace $T_\textrm{opt}$ $\sigma_\textrm{opt}$'-model corresponds to `Our model' that is referred to in other figures. MIE stands for model-internal ensembling.}
  \vspace*{-1em}
  \scriptsize
  \renewcommand{\arraystretch}{.9}
  \newcommand{\fooo}[1]{\textcolor{mycolor2}{\tiny #1}}
   \newcommand{\foog}[1]{\textcolor{mycolor1}{\tiny #1}}
   \newcommand{\foe}[0]{\textcolor{mycolor1}{\tiny $\phantom{+0.0}$}}
   \newcommand{\fooe}[0]{\textcolor{mycolor1}{\tiny $\phantom{+0.00}$}}
   \newcommand{\fooee}[0]{\textcolor{mycolor1}{\tiny $\phantom{+0.000}$}}
  \setlength{\tabcolsep}{0pt}
  \setlength{\tblw}{0.1\textwidth}  
  \begin{tabularx}{\textwidth}{l l @{\extracolsep{\fill}} C{\tblw}  C{\tblw} C{\tblw} C{\tblw} | C{\tblw} C{\tblw} C{\tblw}  C{\tblw} }
  \toprule

& & \multicolumn{4}{c}{\sc CIFAR-100} & \multicolumn{4}{c}{\sc ImageNet} \\
& ($n_\textrm{train}$, $d$, $c$, $n_\textrm{batch}$) & \multicolumn{4}{c}{(50000, 3072, 100, 64)} & \multicolumn{4}{c}{(1281167, 150528, 1000, 256)} \\
\midrule
& & Top-1 ACC $\uparrow$ & Top-5 ACC $\uparrow$ & NLPD $\downarrow$ & ECE $\downarrow$ & Top-1 ACC $\uparrow$ & Top-5 ACC $\uparrow$ & NLPD $\downarrow$ & ECE $\downarrow$ \\
\midrule
\parbox[t]{7mm}{\multirow{4}{*}{\rotatebox[origin=c]{90}{\bf Small}}} 
& MSDNet (vanilla) & $ 69.25$~\fooe{} & $ 90.48$~\fooe{} & $ 1.498$~\fooee{} & $ 0.182$~\fooee{} & $ 68.15$~\fooe{} & $\bf 88.22$~\fooe{} & $ 1.338$~\fooee{} & $ 0.019$~\fooee{} \\
& ~~~+ Laplace $T_\textrm{opt}$ $\sigma_\textrm{opt}$ & $ 69.06$~\fooo{$-0.19$} & $ 90.58$~\foog{$+0.10$} & $ 1.208$~\foog{$-0.289$} & $ 0.073$~\foog{$-0.109$} & $ 68.10$~\fooo{$-0.05$} & $ 88.18$~\fooo{$-0.04$} & $\bf 1.337$~\foog{$-0.001$} & $\bf 0.015$~\foog{$-0.005$} \\
& ~~~+ MIE & $\bf 69.97$~\foog{$+0.72$} & $ 90.88$~\foog{$+0.40$} & $ 1.218$~\foog{$-0.280$} & $ 0.080$~\foog{$-0.102$} & $ 68.27$~\foog{$+0.12$} & $ 88.13$~\fooo{$-0.10$} & $ 1.355$~\fooo{$+0.017$} & $ 0.055$~\fooo{$+0.036$} \\
& ~~~+ MIE Laplace $T_\textrm{opt}$ $\sigma_\textrm{opt}$ & $ 69.84$~\foog{$+0.59$} & $\bf 91.09$~\foog{$+0.61$} & $\bf 1.133$~\foog{$-0.364$} & $\bf 0.017$~\foog{$-0.165$} & $\bf 68.31$~\foog{$+0.16$} & $ 88.11$~\fooo{$-0.11$} & $ 1.356$~\fooo{$+0.018$} & $ 0.052$~\fooo{$+0.032$} \\
\midrule
\parbox[t]{7mm}{\multirow{4}{*}{\rotatebox[origin=c]{90}{\bf Medium}}}
& MSDNet (vanilla) & $ 74.12$~\fooe{} & $ 91.94$~\fooe{} & $ 1.549$~\fooee{} & $ 0.190$~\fooee{} & $ 72.78$~\fooe{} & $ 91.01$~\fooe{} & $ 1.123$~\fooee{} & $ 0.033$~\fooee{} \\
& ~~~+ Laplace $T_\textrm{opt}$ $\sigma_\textrm{opt}$ & $ 73.92$~\fooo{$-0.20$} & $ 92.01$~\foog{$+0.06$} & $ 1.070$~\foog{$-0.479$} & $ 0.083$~\foog{$-0.107$} & $ 72.72$~\fooo{$-0.07$} & $ 91.03$~\foog{$+0.03$} & $\bf 1.118$~\foog{$-0.005$} & $\bf 0.018$~\foog{$-0.015$} \\
& ~~~+ MIE & $\bf 75.03$~\foog{$+0.91$} & $ 92.97$~\foog{$+1.03$} & $ 1.011$~\foog{$-0.538$} & $ 0.050$~\foog{$-0.140$} & $ 72.98$~\foog{$+0.20$} & $\bf 91.12$~\foog{$+0.11$} & $ 1.119$~\foog{$-0.004$} & $ 0.042$~\fooo{$+0.009$} \\
& ~~~+ MIE Laplace $T_\textrm{opt}$ $\sigma_\textrm{opt}$ & $ 74.99$~\foog{$+0.86$} & $\bf 93.23$~\foog{$+1.29$} & $\bf 0.944$~\foog{$-0.605$} & $\bf 0.026$~\foog{$-0.164$} & $\bf 73.04$~\foog{$+0.26$} & $ 90.96$~\fooo{$-0.05$} & $ 1.121$~\foog{$-0.002$} & $ 0.031$~\foog{$-0.003$} \\
\midrule
\parbox[t]{7mm}{\multirow{4}{*}{\rotatebox[origin=c]{90}{\bf Large}}}
& MSDNet (vanilla) & $ 75.36$~\fooe{} & $ 92.78$~\fooe{} & $ 1.475$~\fooee{} & $ 0.178$~\fooee{} & $ 74.33$~\fooe{} & $ 91.57$~\fooe{} & $ 1.066$~\fooee{} & $ 0.050$~\fooee{} \\
& ~~~+ Laplace $T_\textrm{opt}$ $\sigma_\textrm{opt}$ & $ 75.32$~\fooo{$-0.05$} & $ 92.83$~\foog{$+0.05$} & $ 0.996$~\foog{$-0.479$} & $ 0.075$~\foog{$-0.103$} & $ 74.29$~\fooo{$-0.04$} & $ 91.53$~\fooo{$-0.04$} & $ 1.053$~\foog{$-0.013$} & $\bf 0.020$~\foog{$-0.030$} \\
& ~~~+ MIE & $ 76.32$~\foog{$+0.95$} & $ 93.50$~\foog{$+0.72$} & $ 0.949$~\foog{$-0.525$} & $ 0.061$~\foog{$-0.117$} & $\bf 74.82$~\foog{$+0.49$} & $\bf 91.88$~\foog{$+0.30$} & $\bf 1.029$~\foog{$-0.037$} & $ 0.028$~\foog{$-0.022$} \\
& ~~~+ MIE Laplace $T_\textrm{opt}$ $\sigma_\textrm{opt}$ & $\bf 76.34$~\foog{$+0.98$} & $\bf 93.84$~\foog{$+1.05$} & $\bf 0.885$~\foog{$-0.590$} & $\bf 0.025$~\foog{$-0.152$} & $ 74.80$~\foog{$+0.47$} & $ 91.81$~\foog{$+0.24$} & $ 1.032$~\foog{$-0.034$} & $ 0.032$~\foog{$-0.019$}\\
  \bottomrule
  \end{tabularx}
  \label{tbl:results}
\end{table*}

\section{Experiments}
\label{sec:experiments}
We performed a series of experiments on benchmark image classification tasks to assess the improvements obtained through our probabilistic treatment applied to an early-exit DNN.
For each model size for all data sets, a common backbone DNN (MSDNet~\cite{huang2018multi}) was trained minimising the L2 regularised sum of cross-entropy losses computed for all exits on the training set. We trained three different-sized DNN backbone models on each data set to cover a larger range of budgets. Depending on the desired budget, only one of these models would be used at a time. We refer to these models as `small', `medium', and `large'.

After training, we evaluated each model on the test set in a budgeted batch classification setup. Early exiting decisions were based on model predicted confidence, for which thresholds $t_{k}, k = 1, 2, \ldots, n_\text{block}$ were calculated on the validation set. We refer to \cite{huang2018multi} for details on the calculation of the thresholds. We report the Top-1 and Top-5 accuracies of each model over a range of computational budgets measured in average floating point operations (FLOPs) per test sample. In addition, following the recommendations for better validation metrics in image analysis \cite{hein2022metrics}, we compare the negative log-predictive density (NLPD) and the expected calibration error (ECE). Note that NLPD captures both accuracy and uncertainty quantification quality while ECE assesses only calibration, \ie, how consistent the confidence scores are with the posterior probabilities. See \cref{sec:app_metrics} for more details on the metrics. In figures, `Our model' refers to an MSDNet backbone using Laplace approximation and MIE, while optimizing temperature scales and Laplace prior variances in a grid search. We additionally trained DenseNet and ResNet models as baselines, see \cref{sec:baselines} for details on them. For more baseline results we refer to \cite[Sec.~5.2]{huang2018multi}.

\smallskip
\noindent
\textbf{Ablation studies}
were performed to investigate the individual contribution of the Laplace approximation and model-internal ensembling (MIE) to the model performance, testing models that use either only Laplace or only MIE. The results of this ablation study are included in \cref{tbl:results} and \cref{tbl:results_caltech}. Results of a more comprehensive ablation study are in \cref{tbl:results_appendix,tbl:results_appendix_caltech} in \cref{sec:app_results}. Note that as Laplace and MIE are applied on the same trained vanilla MSDNet model that is used on its own, the differences in the results are not from randomness between different training runs. Laplace approximation improves the uncertainty quantification properties of the model by lowering NLPD and ECE values, whereas MIE usually improves both accuracy and uncertainty quantification properties. Considering overall performance over all four metrics (Top-1 and Top-5 accuracy, NLPD, and ECE), we can see that Laplace and MIE together give the best performance on CIFAR-100, whereas on ImageNet and Caltech-256 using MIE alone or together with Laplace both give roughly equally good results. Despite some inconsistency in the combination of uncertainty quantification methods giving best performance, it is clear that improving uncertainty estimation improves the DNN performance over the vanilla model. The range of budgets over which results were averaged to obtain the numbers in \cref{tbl:results,tbl:results_caltech} are listed in \cref{tbl:budget_ranges} in \cref{sec:app_details}.\looseness-1

\begin{table}[t!]
  \caption{\small Table of Top-1/Top-5 accuracy, NLPD, and ECE for different models on Caltech-256 data. All numbers are averages over a range of computational budgets in the budgeted batch classification setup. `MIE Laplace $T_\textrm{opt}$ $\sigma_\textrm{opt}$'-model corresponds to `Our model' that is referred to in other figures.}
  \vspace{-1em}
  \scriptsize
  \renewcommand{\arraystretch}{.9}
  \newcommand{\fooo}[1]{\textcolor{mycolor2}{\tiny #1}}
   \newcommand{\foog}[1]{\textcolor{mycolor1}{\tiny #1}}
   \newcommand{\foe}[0]{\textcolor{mycolor1}{\tiny $\phantom{+0.0}$}}
   \newcommand{\fooe}[0]{\textcolor{mycolor1}{\tiny $\phantom{+0.00}$}}
   \newcommand{\fooee}[0]{\textcolor{mycolor1}{\tiny $\phantom{+0.000}$}}
  \setlength{\tabcolsep}{0pt}
  \setlength{\tblw}{0.08\textwidth}  
  \begin{tabularx}{\columnwidth}{l l @{\extracolsep{\fill}} C{\tblw}  C{\tblw} C{\tblw} C{\tblw}  }
  \toprule

& & \multicolumn{4}{c}{\sc Caltech-256} \\
& ($n_\textrm{train}$, $d$, $c$, $n_\textrm{batch}$) & \multicolumn{4}{c}{(25607, 150528, 257, 128)} \\
\midrule
& & Top-1~ACC~$\uparrow$ & Top-5~ACC~$\uparrow$ & NLPD $\downarrow$ & ECE $\downarrow$ \\
\midrule
\parbox[t]{3mm}{\multirow{4}{*}{\rotatebox[origin=c]{90}{\bf Small}}}
& MSDNet (vanilla) & $ 61.0$~\foe{} & $ 78.2$~\foe{} & $ 2.16$~\fooe{} & $ 0.18$~\fooe{} \\
& ~+ Lap $T_\textrm{opt}$ $\sigma_\textrm{opt}$ & $ 60.5$~\fooo{$-0.5$} & $ 78.1$~\fooo{$-0.1$} & $ 1.86$~\foog{$-0.29$} & $\bf 0.05$~\foog{$-0.13$} \\
& ~+ MIE & $\bf 61.9$~\foog{$+0.9$} & $ 78.8$~\foog{$+0.6$} & $ 1.94$~\foog{$-0.21$} & $ 0.08$~\foog{$-0.10$} \\
& ~+ MIE Lap $T_\textrm{opt}$ $\sigma_\textrm{opt}$ & $ 61.7$~\foog{$+0.6$} & $\bf 79.0$~\foog{$+0.8$} & $\bf 1.81$~\foog{$-0.34$} & $ 0.09$~\foog{$-0.09$} \\
\midrule
\parbox[t]{3mm}{\multirow{4}{*}{\rotatebox[origin=c]{90}{\bf Medium}}}
& MSDNet (vanilla) & $ 63.8$~\foe{} & $ 80.2$~\foe{} & $ 1.98$~\fooe{} & $ 0.17$~\fooe{} \\
& ~+ Lap $T_\textrm{opt}$ $\sigma_\textrm{opt}$ & $ 63.4$~\fooo{$-0.4$} & $ 79.9$~\fooo{$-0.3$} & $ 1.74$~\foog{$-0.24$} & $\bf 0.07$~\foog{$-0.10$} \\
& ~+ MIE & $\bf 65.1$~\foog{$+1.3$} & $\bf 81.4$~\foog{$+1.2$} & $ 1.72$~\foog{$-0.26$} & $ 0.08$~\foog{$-0.09$} \\
& ~+ MIE Lap $T_\textrm{opt}$ $\sigma_\textrm{opt}$ & $ 64.3$~\foog{$+0.5$} & $ 81.3$~\foog{$+1.1$} & $\bf 1.65$~\foog{$-0.33$} & $ 0.08$~\foog{$-0.09$} \\
\midrule
\parbox[t]{3mm}{\multirow{4}{*}{\rotatebox[origin=c]{90}{\bf Large}}}
& MSDNet (vanilla) & $ 64.9$~\foe{} & $ 80.7$~\foe{} & $ 1.90$~\fooe{} & $ 0.16$~\fooe{} \\
& ~+ Lap $T_\textrm{opt}$ $\sigma_\textrm{opt}$ & $ 64.7$~\fooo{$-0.2$} & $ 80.7$~\foog{$+0.0$} & $ 1.65$~\foog{$-0.25$} & $\bf 0.04$~\foog{$-0.12$} \\
& ~+ MIE & $\bf 65.9$~\foog{$+0.9$} & $ 82.4$~\foog{$+1.8$} & $ 1.62$~\foog{$-0.28$} & $ 0.06$~\foog{$-0.10$} \\
& ~+ MIE Lap $T_\textrm{opt}$ $\sigma_\textrm{opt}$ & $ 65.6$~\foog{$+0.7$} & $\bf 82.5$~\foog{$+1.8$} & $\bf 1.58$~\foog{$-0.32$} & $ 0.09$~\foog{$-0.07$} \\
  \bottomrule
  \end{tabularx}
  \label{tbl:results_caltech}
\end{table}

\vskip -0.3em

\begin{figure}[t!]
  \centering\scriptsize
  \pgfplotsset{axis on top,scale only axis,width=\figurewidth,height=\figureheight,
    tick label style={font=\tiny},
    grid style={line width=.1pt, draw=gray!05,densely dashed},
    x tick scale label style={yshift=.75em},
    tick scale binop=\times,
    xtick={}
  }
  \setlength{\figurewidth}{.18\textwidth}
  \setlength{\figureheight}{.8\figurewidth}  
  \begin{subfigure}[t]{.49\columnwidth}
    \centering
\begin{tikzpicture}

\definecolor{color0}{rgb}{1,0.498039215686275,0.0549019607843137}
\definecolor{color1}{rgb}{0,1,1}

\begin{axis}[
legend cell align={left},
legend style={
  font=\tiny,
  fill opacity=0.8,
  draw opacity=1,
  text opacity=1,
  at={(0.97,0.03)},
  anchor=south east,
  draw=white!80!black
},
tick align=outside,
tick pos=left,
x grid style={white!69.0196078431373!black},
xlabel={FLOPs},
xmajorgrids,
xmin=-5280348.53027194, xmax=150000000,
xtick style={color=black},
xtick={-50000000,0,50000000,100000000,150000000,200000000,250000000,300000000},
xticklabels={−0.5,0.0,0.5,1.0,1.5,2.0,2.5,3.0},
y grid style={white!69.0196078431373!black},
ylabel={Top-1 accuracy (\%) \(\displaystyle \rightarrow\)},
ymajorgrids,
ymin=68, ymax=77,
ytick style={color=black}
]
\addplot [thick, color0]
table {%
7107824 62.33
7433575.5 62.97
7824412 63.61
8216395.5 64.14
8683373 64.85
9287369 65.25
9930998 65.82
10515920 66.43
11173550 66.92
11935525 67.58
12679738 68.14
13384698 68.71
14116088 68.98
14772887 69.35
15558605 70
16324249 70.34
17068580 70.42
17845372 70.68
18466548 70.94
19134888 71.18
19820094 71.35
20431254 71.58
20972820 71.63
21512296 71.48
22070524 71.67
22691414 71.78
23179666 71.81
23633144 71.74
24105429.4119488 71.7649643578724
};
\addlegendentry{MSDNet (vanilla)}

\addplot [thick, black, opacity=0.4, forget plot]
table {%
26383677.4924007 73.0662871572738
26502622.3776224 73.07
26844975.9272237 73.11
27156992.7428124 73.1
};
\addplot [thick, black, forget plot]
table {%
26383677.4924007 73.0662871572738
27114957.2485038 73.28
29833869.7630945 73.9
32193888.3221878 74.43
34902012.545505 74.77
37505023 75.07
39965614.9145572 75.2
42091339.3743385 75.22
44270834.2128857 75.4
46139542.7845058 75.41
48170764.7514962 75.52
50171590.9730974 75.57
52056350.5428953 75.65
53776141.9914138 75.64
55140028.6506293 75.7
56692586.1654135 75.71
58068878.1313194 75.71
59659736.3985777 75.7274168604498
};
\addplot [thick, black, opacity=0.4, forget plot]
table {%
17670232.3676155 69.89
19641217.800252 70.74
21816748.9328533 71.58
26383677.4924007 73.0662871572738
};
\addplot [thick, black, opacity=0.4, forget plot]
table {%
59659736.3985777 75.7274168604498
60628822.0878741 75.72
61616735.0994646 75.71
62521620.4327495 75.71
};
\addplot [thick, black, forget plot]
table {%
59659736.3985777 75.7274168604498
64233524.5772711 75.98
71038681 76.3
77993048.4721301 76.38
84715634.1080382 76.5
91594311.8345746 76.47
97614458.5695152 76.61
103256333.376398 76.56
108160176.534447 76.62
113064533.113946 76.6
117581385.339844 76.59
121380453.806219 76.58
124818427.473434 76.57
128518258.319017 76.57
131570776.498282 76.58
134615441.88127 76.58
137322823.854245 76.58
139457809.29821 76.58
141946674.679242 76.58
144026238.203987 76.58
146046644.008189 76.58
147637730.35694 76.58
};
\addplot [thick, black, opacity=0.4, forget plot]
table {%
37026894.5865197 73.38
43367200.6236015 74.39
49827344.4937234 74.86
59659736.3985777 75.7274168604498
};
\addplot [thick, color1]
table {%
7125618.77811424 62.3
7400580.23537354 62.89
7783853.88093973 63.44
8203447.8974359 63.96
8728716.41176471 64.4
9366414.42625265 64.86
9946913.26569331 65.55
10601025.3990148 66.46
11273676.0225106 66.96
11991577.3333333 67.52
12722417.4660393 68.39
13492461.1029412 69.06
14335418.7962401 69.35
15221740.0552704 69.72
15892448.2571429 70.04
16592819.4363144 70.43
17382477.0011376 71
18061638.4158186 71.16
18710095.8763435 71.43
19329389 71.61
20053034.5407036 71.94
20642700.1366085 71.9
21233120.0923223 71.91
21839705.195231 71.92
22333076.5636856 71.97
22803949.5238403 72.01
23531488.5266145 71.9104464086334
};
\addlegendentry{Vanilla predictions}

\addplot [thick, color0, opacity=0.4, forget plot]
table {%
24105429.4119488 71.7649643578724
24411224 71.8
24769926 71.8
25141960 71.79
};
\addplot [thick, color0, forget plot]
table {%
24105429.4119488 71.7649643578724
24304066 71.83
26532344 72.35
29241978 72.58
31584578 73.24
34147504 73.71
36494240 74.1
38787372 74.38
41154776 74.56
43361168 74.58
45359856 74.66
47402496 74.8
49196656 74.8
50801440 74.69
53011407.411766 74.7800000000001
};
\addplot [thick, color0, opacity=0.4, forget plot]
table {%
15969418 68.78
17779276 69.65
19822496 70.42
24105429.4119488 71.7649643578724
};
\addplot [thick, color0, opacity=0.4, forget plot]
table {%
53011407.411766 74.7800000000001
53927804 74.78
55503776 74.81
57045520 74.78
};
\addplot [thick, color0, forget plot]
table {%
53011407.411766 74.7800000000001
53689056 74.85
60636272 75.18
67259920 75.46
74724928 75.56
81708528 75.52
88334648 75.57
94084560 75.45
99491560 75.43
104465712 75.44
109811712 75.39
114441208 75.39
118463360 75.41
121874664 75.39
125652376 75.36
129252896 75.39
132409536 75.32
135074560 75.29
137454576 75.3
140058576 75.28
141967392 75.25
143816160 75.25
145699632 75.24
};
\addplot [thick, color0, opacity=0.4, forget plot]
table {%
29415024 71.42
34514704 72.47
40481004 73.2
53011407.411766 74.7800000000001
};
\addplot [thick, black]
table {%
7125618.77811424 62.26
7400580.23537354 62.83
7783853.88093973 63.38
8203447.8974359 64.07
8728716.41176471 64.57
9366414.42625265 65.02
9946913.26569331 65.68
10601025.3990148 66.55
11273676.0225106 67.15
11991577.3333333 67.77
12722417.4660393 68.51
13492461.1029412 69.18
14335418.7962401 69.61
15221740.0552704 70.05
15892448.2571429 70.52
16592819.4363144 70.9
17382477.0011376 71.42
18061638.4158186 71.63
18710095.8763435 71.91
19329389 72.1
20053034.5407036 72.46
20642700.1366085 72.47
21233120.0923223 72.57
21839705.195231 72.71
22333076.5636856 72.83
22803949.5238403 72.98
23289995.511524 73.02
23752341.9954955 73.07
24210268.4186222 73.08
24630222.1538462 73.06
24963878.472173 73.02
25412812.5203112 73.04
25821995.3059544 73.07
26383677.4924007 73.0662871572738
};
\addlegendentry{Our model}

\addplot [thick, color1, opacity=0.4, forget plot]
table {%
23531488.5266145 71.9104464086334
23752341.9954955 71.92
24210268.4186222 71.79
24630222.1538462 71.71
};
\addplot [thick, color1, forget plot]
table {%
23531488.5266145 71.9104464086334
24445963.6471636 72.22
27114957.2485038 73.1
29833869.7630945 73.56
32193888.3221878 74.02
34902012.545505 74.36
37505023 74.7
39965614.9145572 74.73
42091339.3743385 74.8
44270834.2128857 74.94
46139542.7845058 74.98
48170764.7514962 75.03
50171590.9730974 75.09
52056350.5428953 75
54314890.7913802 74.9797504974602
};
\addplot [thick, color1, opacity=0.4, forget plot]
table {%
15866300.2508412 69.06
17670232.3676155 69.76
19641217.800252 70.55
23531488.5266145 71.9104464086334
};
\addplot [thick, color1, opacity=0.4, forget plot]
table {%
54314890.7913802 74.9797504974602
55140028.6506293 75.01
56692586.1654135 74.98
58068878.1313194 75
};
\addplot [thick, color1, forget plot]
table {%
54314890.7913802 74.9797504974602
56990304.4845342 75.23
64233524.5772711 75.77
71038681 76.06
77993048.4721301 76.01
84715634.1080382 76.05
91594311.8345746 75.91
97614458.5695152 76
103256333.376398 75.84
108160176.534447 75.91
113064533.113946 75.81
117581385.339844 75.79
121380453.806219 75.7
124818427.473434 75.55
128518258.319017 75.45
131570776.498282 75.43
134615441.88127 75.46
137322823.854245 75.43
139457809.29821 75.37
141946674.679242 75.34
144026238.203987 75.31
146046644.008189 75.29
147637730.35694 75.29
};
\addplot [thick, color1, opacity=0.4, forget plot]
table {%
31276498.1800991 72.1
37026894.5865197 73.26
43367200.6236015 74.14
54314890.7913802 74.9797504974602
};
\addlegendentry{ResNets}
\end{axis}

\end{tikzpicture}
  \end{subfigure}
  \hfill
  \begin{subfigure}[t]{.49\columnwidth}
    \centering
    \input{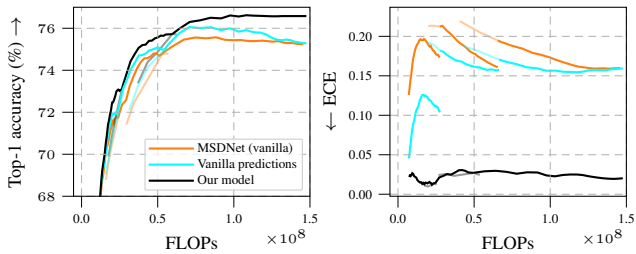}
  \end{subfigure}
  \vspace{-2.2em}
  \caption{A decision-making experiment, where vanilla MSDNet and `Our model' are compared to results obtained using our model for decision-making, and taking predictions from vanilla MSDNet (labelled `Vanilla predictions'). Full results in \cref{fig:cifar100_result_decision} in \cref{sec:app_results}.\looseness-1}
\label{fig:cifar100_result_decision_maintext}
\vspace*{-1em}
\end{figure}

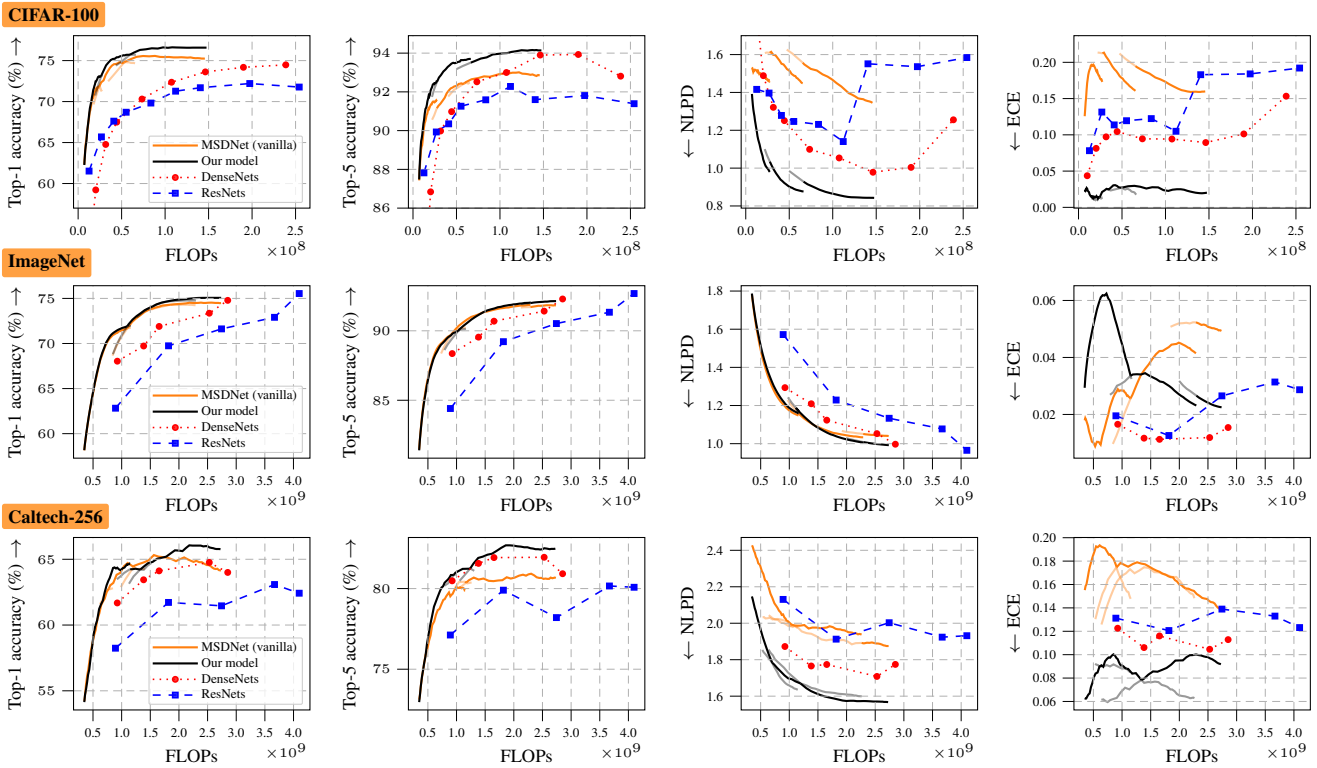
\begin{figure*}[t!]
  \centering\scriptsize
  \pgfplotsset{axis on top,scale only axis,width=\figurewidth,height=\figureheight,
    tick label style={font=\tiny},
    grid style={line width=.1pt, draw=gray!05,densely dashed},
    x tick scale label style={yshift=.75em},
    tick scale binop=\times,
    xtick={}
  }
  \setlength{\figurewidth}{.18\textwidth}
  \setlength{\figureheight}{.73\figurewidth}  
  \begin{subfigure}[b]{.24\textwidth}
    \raggedright
    \tikz\node[font=\bf,fill=C0,rounded corners=1pt]{CIFAR-100};\\
\begin{tikzpicture}

\definecolor{color0}{rgb}{1,0.498039215686275,0.0549019607843137}

\begin{axis}[
legend cell align={left},
legend style={
  font=\tiny,
  fill opacity=0.8,
  draw opacity=1,
  text opacity=1,
  at={(0.97,0.03)},
  anchor=south east,
  draw=white!80!black
},
tick align=outside,
tick pos=left,
x grid style={white!69.0196078431373!black},
xlabel={FLOPs},
xmajorgrids,
xmin=-5280348.53027194, xmax=266564960.977632,
xtick style={color=black},
xtick={-50000000,0,50000000,100000000,150000000,200000000,250000000,300000000},
xticklabels={−0.5,0.0,0.5,1.0,1.5,2.0,2.5,3.0},
y grid style={white!69.0196078431373!black},
ylabel={Top-1 accuracy (\%) \(\displaystyle \rightarrow\)},
ymajorgrids,
ymin=57, ymax=78.0425,
ytick style={color=black}
]
\addplot [thick, color0]
table {%
7107824 62.33
7433575.5 62.97
7824412 63.61
8216395.5 64.14
8683373 64.85
9287369 65.25
9930998 65.82
10515920 66.43
11173550 66.92
11935525 67.58
12679738 68.14
13384698 68.71
14116088 68.98
14772887 69.35
15558605 70
16324249 70.34
17068580 70.42
17845372 70.68
18466548 70.94
19134888 71.18
19820094 71.35
20431254 71.58
20972820 71.63
21512296 71.48
22070524 71.67
22691414 71.78
23179666 71.81
23633144 71.74
24105429.4119488 71.7649643578724
};
\addlegendentry{MSDNet (vanilla)}
\addplot [thick, color0, opacity=0.4, forget plot]
table {%
24105429.4119488 71.7649643578724
24411224 71.8
24769926 71.8
25141960 71.79
25548760 71.66
25957200 71.57
26300288 71.4
26504228 71.34
26837524 71.34
27163056 71.3
27421310 71.25
};
\addplot [thick, color0, forget plot]
table {%
24105429.4119488 71.7649643578724
24304066 71.83
26532344 72.35
29241978 72.58
31584578 73.24
34147504 73.71
36494240 74.1
38787372 74.38
41154776 74.56
43361168 74.58
45359856 74.66
47402496 74.8
49196656 74.8
50801440 74.69
53011407.411766 74.7800000000001
};
\addplot [thick, color0, opacity=0.4, forget plot]
table {%
17779276 69.65
19822496 70.42
24105429.4119488 71.7649643578724
};
\addplot [thick, color0, opacity=0.4, forget plot]
table {%
53011407.411766 74.7800000000001
53927804 74.78
55503776 74.81
57045520 74.78
58413856 74.8
59573664 74.75
60552740 74.72
61621636 74.71
62616744 74.71
63702584 74.71
64671696 74.69
65501088 74.67
};
\addplot [thick, color0, forget plot]
table {%
53011407.411766 74.7800000000001
53689056 74.85
60636272 75.18
67259920 75.46
74724928 75.56
81708528 75.52
88334648 75.57
94084560 75.45
99491560 75.43
104465712 75.44
109811712 75.39
114441208 75.39
118463360 75.41
121874664 75.39
125652376 75.36
129252896 75.39
132409536 75.32
135074560 75.29
137454576 75.3
140058576 75.28
141967392 75.25
143816160 75.25
145699632 75.24
};
\addplot [thick, color0, opacity=0.4, forget plot]
table {%
34514704 72.47
40481004 73.2
53011407.411766 74.7800000000001
};
\addplot [thick, black]
table {%
7125618.77811424 62.26
7400580.23537354 62.83
7783853.88093973 63.38
8203447.8974359 64.07
8728716.41176471 64.57
9366414.42625265 65.02
9946913.26569331 65.68
10601025.3990148 66.55
11273676.0225106 67.15
11991577.3333333 67.77
12722417.4660393 68.51
13492461.1029412 69.18
14335418.7962401 69.61
15221740.0552704 70.05
15892448.2571429 70.52
16592819.4363144 70.9
17382477.0011376 71.42
18061638.4158186 71.63
18710095.8763435 71.91
19329389 72.1
20053034.5407036 72.46
20642700.1366085 72.47
21233120.0923223 72.57
21839705.195231 72.71
22333076.5636856 72.83
22803949.5238403 72.98
23289995.511524 73.02
23752341.9954955 73.07
24210268.4186222 73.08
24630222.1538462 73.06
24963878.472173 73.02
25412812.5203112 73.04
25821995.3059544 73.07
26383677.4924007 73.0662871572738
};
\addlegendentry{Our model}
\addplot [thick, black, opacity=0.4, forget plot]
table {%
26383677.4924007 73.0662871572738
26502622.3776224 73.07
26844975.9272237 73.11
27156992.7428124 73.1
27386683.0826539 73.08
27701139.0634998 73.05
};
\addplot [thick, black, forget plot]
table {%
26383677.4924007 73.0662871572738
27114957.2485038 73.28
29833869.7630945 73.9
32193888.3221878 74.43
34902012.545505 74.77
37505023 75.07
39965614.9145572 75.2
42091339.3743385 75.22
44270834.2128857 75.4
46139542.7845058 75.41
48170764.7514962 75.52
50171590.9730974 75.57
52056350.5428953 75.65
53776141.9914138 75.64
55140028.6506293 75.7
56692586.1654135 75.71
58068878.1313194 75.71
59659736.3985777 75.7274168604498
};
\addplot [thick, black, opacity=0.4, forget plot]
table {%
19641217.800252 70.74
21816748.9328533 71.58
26383677.4924007 73.0662871572738
};
\addplot [thick, black, opacity=0.4, forget plot]
table {%
59659736.3985777 75.7274168604498
60628822.0878741 75.72
61616735.0994646 75.71
62521620.4327495 75.71
63512250.9704391 75.72
64602786.9168033 75.71
65517615.8154501 75.71
66434312.5786651 75.71
};
\addplot [thick, black, forget plot]
table {%
59659736.3985777 75.7274168604498
64233524.5772711 75.98
71038681 76.3
77993048.4721301 76.38
84715634.1080382 76.5
91594311.8345746 76.47
97614458.5695152 76.61
103256333.376398 76.56
108160176.534447 76.62
113064533.113946 76.6
117581385.339844 76.59
121380453.806219 76.58
124818427.473434 76.57
128518258.319017 76.57
131570776.498282 76.58
134615441.88127 76.58
137322823.854245 76.58
139457809.29821 76.58
141946674.679242 76.58
144026238.203987 76.58
146046644.008189 76.58
147637730.35694 76.58
};
\addplot [thick, black, opacity=0.4, forget plot]
table {%
43367200.6236015 74.39
49827344.4937234 74.86
59659736.3985777 75.7274168604498
};
\addplot [semithick, red, dotted, mark=*, mark size=1, mark options={solid}]
table {%
10020784 48.17
20250148 59.22
31707160 64.77
44391820 67.49
73444084 70.31
107406940 72.36
146280388 73.62
190064428 74.18
238759060 74.48
};
\addlegendentry{DenseNets}
\addplot [semithick, blue, dashed, mark=square*, mark size=1, mark options={solid}]
table {%
12585316 61.52
26798436 65.69
41011556 67.61
55224676 68.7
83650916 69.82
112077156 71.27
140503396 71.69
197355876 72.2
254208356 71.77
};
\addlegendentry{ResNets}
\end{axis}

\end{tikzpicture}
  \end{subfigure}
  \hfill
  \begin{subfigure}[b]{.24\textwidth}
    \raggedleft
\begin{tikzpicture}

\definecolor{color0}{rgb}{1,0.498039215686275,0.0549019607843137}

\begin{axis}[
tick align=outside,
tick pos=left,
x grid style={white!69.0196078431373!black},
xlabel={FLOPs},
xmajorgrids,
xmin=-5280348.53027194, xmax=266564960.977632,
xtick style={color=black},
xtick={-50000000,0,50000000,100000000,150000000,200000000,250000000,300000000},
xticklabels={−0.5,0.0,0.5,1.0,1.5,2.0,2.5,3.0},
y grid style={white!69.0196078431373!black},
ylabel={Top-5 accuracy (\%) \(\displaystyle \rightarrow\)},
ymajorgrids,
ymin=86, ymax=94.9014958343506,
ytick style={color=black}
]
\addplot [thick, color0]
table {%
7107824 87.4300003051758
7433575.5 87.7399978637695
7824412 88.0599975585938
8216395.5 88.379997253418
8683373 88.6699981689453
9287369 89.0199966430664
9930998 89.1800003051758
10515920 89.2399978637695
11173550 89.4899978637695
11935525 89.9000015258789
12679738 90.1800003051758
13384698 90.379997253418
14116088 90.4700012207031
14772887 90.5999984741211
15558605 90.7900009155273
16324249 90.8600006103516
17068580 91.0199966430664
17845372 91.1500015258789
18466548 91.1199951171875
19134888 91.1999969482422
19820094 91.2799987792969
20431254 91.3299942016602
20972820 91.3600006103516
21512296 91.3600006103516
22070524 91.4499969482422
22691414 91.4599990844727
23179666 91.5
23633144 91.5
24062100 91.5
24411224 91.4799957275391
24769926 91.4599990844727
25141960 91.4700012207031
25548760 91.5
25957200 91.5099945068359
26300288 91.5299987792969
26504228 91.5299987792969
26837524 91.5599975585938
27163056 91.5400009155273
27421310 91.5199966430664
};
\addplot [thick, color0, opacity=0.4]
table {%
27421310 91.1523315869786
};
\addplot [thick, color0]
table {%
27421310 91.1523315869786
29241978 91.2799987792969
31584578 91.3899993896484
34147504 91.5499954223633
36494240 91.6999969482422
38787372 91.9300003051758
41154776 92.0699996948242
43361168 92.1800003051758
45359856 92.2200012207031
47402496 92.2799987792969
49196656 92.2999954223633
50801440 92.3299942016602
52584412 92.4099960327148
53927804 92.4199981689453
55503776 92.4399948120117
57045520 92.4099960327148
59092337.1641704 92.4399948120117
};
\addplot [thick, color0, opacity=0.4]
table {%
21952276 90.5599975585938
24304066 90.7799987792969
27421310 91.1523315869786
};
\addplot [thick, color0, opacity=0.4]
table {%
59092337.1641704 92.4399948120117
59573664 92.4399948120117
60552740 92.4799957275391
61621636 92.4599990844727
62616744 92.4399948120117
63702584 92.4300003051758
64671696 92.4700012207031
65501088 92.4799957275391
};
\addplot [thick, color0]
table {%
59092337.1641704 92.4399948120117
60636272 92.5
67259920 92.6500015258789
74724928 92.7099990844727
81708528 92.7999954223633
88334648 92.7799987792969
94084560 92.8499984741211
99491560 92.9199981689453
104465712 92.9499969482422
109811712 92.9499969482422
114441208 92.9899978637695
118463360 92.9899978637695
121874664 93.0099945068359
125652376 92.9399948120117
129252896 92.9300003051758
132409536 92.9099960327148
135074560 92.879997253418
137454576 92.9000015258789
140058576 92.8499984741211
141967392 92.8199996948242
143816160 92.8699951171875
145699632 92.8499984741211
};
\addplot [thick, color0, opacity=0.4]
table {%
40481004 91.7599945068359
47106184 92.0999984741211
59092337.1641704 92.4399948120117
};
\addplot [thick, black]
table {%
7125618.77811424 87.4799957275391
7400580.23537354 87.9499969482422
7783853.88093973 88.2599945068359
8203447.8974359 88.629997253418
8728716.41176471 89.0799942016602
9366414.42625265 89.4599990844727
9946913.26569331 89.5799942016602
10601025.3990148 89.7799987792969
11273676.0225106 90.0499954223633
11991577.3333333 90.3600006103516
12722417.4660393 90.5599975585938
13492461.1029412 90.8699951171875
14335418.7962401 91.1199951171875
15221740.0552704 91.1899948120117
15892448.2571429 91.3199996948242
16592819.4363144 91.5199966430664
17382477.0011376 91.6500015258789
18061638.4158186 91.8099975585938
18710095.8763435 91.7599945068359
19329389 91.8399963378906
20053034.5407036 92.0299987792969
20642700.1366085 92.1199951171875
21233120.0923223 92.1800003051758
21839705.195231 92.2399978637695
22333076.5636856 92.3099975585938
22803949.5238403 92.379997253418
23289995.511524 92.4000015258789
23752341.9954955 92.4099960327148
24210268.4186222 92.4300003051758
24630222.1538462 92.4099960327148
24963878.472173 92.4399948120117
25412812.5203112 92.4499969482422
25821995.3059544 92.4499969482422
26182261.6890412 92.4599990844727
26502622.3776224 92.4799957275391
26844975.9272237 92.5199966430664
27156992.7428124 92.5499954223633
27386683.0826539 92.5499954223633
27701139.0634998 92.5699996948242
};
\addplot [thick, black, opacity=0.4]
table {%
27701139.0634998 92.5282109355896
};
\addplot [thick, black]
table {%
27701139.0634998 92.5282109355896
29833869.7630945 92.7399978637695
32193888.3221878 92.8699951171875
34902012.545505 93.0499954223633
37505023 93.1599960327148
39965614.9145572 93.2900009155273
42091339.3743385 93.3699951171875
44270834.2128857 93.4599990844727
46139542.7845058 93.5
48170764.7514962 93.4799957275391
50171590.9730974 93.5099945068359
52056350.5428953 93.5499954223633
53776141.9914138 93.5599975585938
55140028.6506293 93.6399993896484
56692586.1654135 93.6399993896484
58068878.1313194 93.6399993896484
59322223.8366617 93.6500015258789
60628822.0878741 93.6500015258789
61616735.0994646 93.6599960327148
62521620.4327495 93.6800003051758
63512250.9704391 93.6899948120117
64602786.9168033 93.6999969482422
65517615.8154501 93.6999969482422
66434312.5786651 93.6999969482422
};
\addplot [thick, black, opacity=0.4]
table {%
21816748.9328533 91.7399978637695
24445963.6471636 92.1500015258789
27701139.0634998 92.5282109355896
};
\addplot [thick, black, opacity=0.4]
table {%
66434312.5786651 93.5088076489066
};
\addplot [thick, black]
table {%
66434312.5786651 93.5088076489066
71038681 93.5899963378906
77993048.4721301 93.7200012207031
84715634.1080382 93.7799987792969
91594311.8345746 93.9300003051758
97614458.5695152 93.9599990844727
103256333.376398 93.9899978637695
108160176.534447 94
113064533.113946 94.0599975585938
117581385.339844 94.0899963378906
121380453.806219 94.1199951171875
124818427.473434 94.1500015258789
128518258.319017 94.1399993896484
131570776.498282 94.129997253418
134615441.88127 94.1500015258789
137322823.854245 94.1599960327148
139457809.29821 94.1500015258789
141946674.679242 94.1599960327148
144026238.203987 94.1399993896484
146046644.008189 94.1399993896484
147637730.35694 94.1399993896484
};
\addplot [thick, black, opacity=0.4]
table {%
49827344.4937234 93.1699981689453
56990304.4845342 93.3299942016602
66434312.5786651 93.5088076489066
};
\addplot [semithick, red, dotted, mark=*, mark size=1, mark options={solid}]
table {%
10020784 79.33
20250148 86.84
31707160 89.98
44391820 90.98
73444084 92.52
107406940 93
146280388 93.9
190064428 93.93
238759060 92.81
};
\addplot [semithick, blue, dashed, mark=square*, mark size=1, mark options={solid}]
table {%
12585316 87.82
26798436 89.93
41011556 90.35
55224676 91.26
83650916 91.59
112077156 92.28
140503396 91.6
197355876 91.81
254208356 91.39
};
\end{axis}

\end{tikzpicture}
  \end{subfigure}
  \hfill
  \begin{subfigure}[b]{.24\textwidth}
    \raggedleft
\begin{tikzpicture}

\definecolor{color0}{rgb}{1,0.498039215686275,0.0549019607843137}

\begin{axis}[
tick align=outside,
tick pos=left,
x grid style={white!69.0196078431373!black},
xlabel={FLOPs},
xmajorgrids,
xmin=-5280348.53027194, xmax=266564960.977632,
xtick style={color=black},
xtick={-50000000,0,50000000,100000000,150000000,200000000,250000000,300000000},
xticklabels={−0.5,0.0,0.5,1.0,1.5,2.0,2.5,3.0},
y grid style={white!69.0196078431373!black},
ylabel={\(\displaystyle \leftarrow\) NLPD},
ymajorgrids,
ymin=0.78861289411068, ymax=1.7,
ytick style={color=black},
ytick={0.6,0.8,1,1.2,1.4,1.6,1.8,2},
yticklabels={0.6,0.8,1.0,1.2,1.4,1.6,1.8,2.0}
]
\addplot [thick, color0]
table {%
7107824 1.53221654891968
7433575.5 1.53026330471039
7824412 1.52114403247833
8216395.5 1.51638972759247
8683373 1.50746464729309
9287369 1.51275110244751
9930998 1.52015745639801
10515920 1.5202271938324
11173550 1.51708471775055
11935525 1.51682269573212
12679738 1.51703798770905
13384698 1.50862097740173
14116088 1.51342737674713
14772887 1.50671911239624
15558605 1.49602687358856
16324249 1.49739158153534
17068580 1.50067985057831
17845372 1.49756264686584
18466548 1.49463427066803
19134888 1.49020540714264
19820094 1.48761200904846
20431254 1.48252701759338
20972820 1.48506891727448
21512296 1.48740041255951
22070524 1.48143911361694
22691414 1.47531700134277
23179666 1.47300362586975
23633144 1.47468423843384
24062100 1.4703768491745
24411224 1.46878707408905
24769926 1.46887612342834
25141960 1.46533048152924
25548760 1.46266162395477
25957200 1.46381425857544
26300288 1.46618616580963
26504228 1.4659880399704
26837524 1.45825302600861
27163056 1.45764005184174
27421310 1.45943999290466
};
\addplot [thick, color0, opacity=0.4]
table {%
27421310 1.60932041523352
};
\addplot [thick, color0]
table {%
27421310 1.60932041523352
29241978 1.6144837141037
31584578 1.5986579656601
34147504 1.59670615196228
36494240 1.58254945278168
38787372 1.5687552690506
41154776 1.55127739906311
43361168 1.54797494411469
45359856 1.53658127784729
47402496 1.52331256866455
49196656 1.51787507534027
50801440 1.5122162103653
52584412 1.49828326702118
53927804 1.49374842643738
55503776 1.48822367191315
57045520 1.48637354373932
58413856 1.47673881053925
59573664 1.47412884235382
60552740 1.46920943260193
61621636 1.4645984172821
62616744 1.46028101444244
63702584 1.45335233211517
64671696 1.44980072975159
65501088 1.44865679740906
};
\addplot [thick, color0, opacity=0.4]
table {%
21952276 1.60813295841217
24304066 1.61088764667511
27421310 1.60932041523352
};
\addplot [thick, color0, opacity=0.4]
table {%
65501088 1.55719135398931
};
\addplot [thick, color0]
table {%
65501088 1.55719135398931
67259920 1.5482177734375
74724928 1.5243821144104
81708528 1.51604390144348
88334648 1.4953898191452
94084560 1.4835911989212
99491560 1.46759378910065
104465712 1.44900834560394
109811712 1.43167054653168
114441208 1.41258752346039
118463360 1.39602065086365
121874664 1.38881480693817
125652376 1.37928998470306
129252896 1.36846578121185
132409536 1.36591815948486
135074560 1.36413991451263
137454576 1.35842049121857
140058576 1.3547431230545
141967392 1.3527592420578
143816160 1.35003280639648
145699632 1.34514927864075
};
\addplot [thick, color0, opacity=0.4]
table {%
47106184 1.6265504360199
53689056 1.60502779483795
65501088 1.55719135398931
};
\addplot [thick, black]
table {%
7125618.77811424 1.39191710948944
7400580.23537354 1.37367022037506
7783853.88093973 1.35329794883728
8203447.8974359 1.33163392543793
8728716.41176471 1.3100129365921
9366414.42625265 1.29025208950043
9946913.26569331 1.27536225318909
10601025.3990148 1.25139665603638
11273676.0225106 1.23306381702423
11991577.3333333 1.21226370334625
12722417.4660393 1.19287014007568
13492461.1029412 1.1688619852066
14335418.7962401 1.15073227882385
15221740.0552704 1.13609528541565
15892448.2571429 1.12529730796814
16592819.4363144 1.11065495014191
17382477.0011376 1.09422099590302
18061638.4158186 1.08301723003387
18710095.8763435 1.0735809803009
19329389 1.0636123418808
20053034.5407036 1.04609036445618
20642700.1366085 1.0384407043457
21233120.0923223 1.03008592128754
21839705.195231 1.02251851558685
22333076.5636856 1.01672446727753
22803949.5238403 1.01155185699463
23289995.511524 1.0091198682785
23752341.9954955 1.005255818367
24210268.4186222 1.00255751609802
24630222.1538462 1.00090026855469
24963878.472173 0.999146699905396
25412812.5203112 0.99626749753952
25821995.3059544 0.993661105632782
26182261.6890412 0.990850210189819
26502622.3776224 0.988641202449799
26844975.9272237 0.984931826591492
27156992.7428124 0.982524693012238
27386683.0826539 0.981906354427338
27701139.0634998 0.980204999446869
};
\addplot [thick, black, opacity=0.4]
table {%
27701139.0634998 1.03438766633219
};
\addplot [thick, black]
table {%
27701139.0634998 1.03438766633219
29833869.7630945 1.01321005821228
32193888.3221878 0.992564558982849
34902012.545505 0.972136914730072
37505023 0.955434083938599
39965614.9145572 0.939627230167389
42091339.3743385 0.931402921676636
44270834.2128857 0.921686232089996
46139542.7845058 0.915391802787781
48170764.7514962 0.907984852790833
50171590.9730974 0.901490211486816
52056350.5428953 0.897752165794373
53776141.9914138 0.894763946533203
55140028.6506293 0.890767574310303
56692586.1654135 0.887837409973145
58068878.1313194 0.88525652885437
59322223.8366617 0.884019613265991
60628822.0878741 0.882425487041473
61616735.0994646 0.881085574626923
62521620.4327495 0.880298018455505
63512250.9704391 0.879160046577454
64602786.9168033 0.877530872821808
65517615.8154501 0.877112090587616
66434312.5786651 0.876872181892395
};
\addplot [thick, black, opacity=0.4]
table {%
21816748.9328533 1.100630402565
24445963.6471636 1.06735980510712
27701139.0634998 1.03438766633219
};
\addplot [thick, black, opacity=0.4]
table {%
66434312.5786651 0.92834067110563
};
\addplot [thick, black]
table {%
66434312.5786651 0.92834067110563
71038681 0.914643943309784
77993048.4721301 0.900363743305206
84715634.1080382 0.885916113853455
91594311.8345746 0.874632298946381
97614458.5695152 0.867303788661957
103256333.376398 0.861395716667175
108160176.534447 0.855931043624878
113064533.113946 0.851162135601044
117581385.339844 0.847744166851044
121380453.806219 0.846085429191589
124818427.473434 0.845715999603271
128518258.319017 0.845103740692139
131570776.498282 0.844457149505615
134615441.88127 0.843655169010162
137322823.854245 0.843315720558167
139457809.29821 0.843172252178192
141946674.679242 0.843230962753296
144026238.203987 0.843006730079651
146046644.008189 0.843057692050934
147637730.35694 0.842688858509064
};
\addplot [thick, black, opacity=0.4]
table {%
49827344.4937234 0.985268771648407
56990304.4845342 0.960482716560364
66434312.5786651 0.92834067110563
};
\addplot [semithick, red, dotted, mark=*, mark size=1, mark options={solid}]
table {%
10020784 1.92420814647675
20250148 1.48857076835632
31707160 1.32116101036072
44391820 1.25081512260437
73444084 1.09929659519196
107406940 1.05370669059753
146280388 0.978194415664673
190064428 1.00394229564667
238759060 1.25530332393646
};
\addplot [semithick, blue, dashed, mark=square*, mark size=1, mark options={solid}]
table {%
12585316 1.41618544445038
26798436 1.396649954319
41011556 1.2784730796814
55224676 1.24661586036682
83650916 1.2310604347229
112077156 1.14040693912506
140503396 1.55117616882324
197355876 1.53617264156342
254208356 1.58421141624451
};
\end{axis}

\end{tikzpicture}
  \end{subfigure}
  \hfill  
  \begin{subfigure}[b]{.24\textwidth}
    \raggedleft
\begin{tikzpicture}

\definecolor{color0}{rgb}{1,0.498039215686275,0.0549019607843137}

\begin{axis}[
tick align=outside,
tick pos=left,
x grid style={white!69.0196078431373!black},
xlabel={FLOPs},
xmajorgrids,
xmin=-5280348.53027194, xmax=266564960.977632,
xtick style={color=black},
xtick={-50000000,0,50000000,100000000,150000000,200000000,250000000,300000000},
xticklabels={−0.5,0.0,0.5,1.0,1.5,2.0,2.5,3.0},
y grid style={white!69.0196078431373!black},
ylabel={\(\displaystyle \leftarrow\) ECE},
ymajorgrids,
ymin=-0.00105453347351614, ymax=0.236764725942711,
ytick style={color=black},
ytick={-0.05,0,0.05,0.1,0.15,0.2,0.25},
yticklabels={−0.05,0.00,0.05,0.10,0.15,0.20,0.25}
]
\addplot [thick, color0]
table {%
7107824 0.125691256588034
7433575.5 0.131132694413129
7824412 0.138093302758624
8216395.5 0.145217961976149
8683373 0.150959387936161
9287369 0.161809195950515
9930998 0.171424663099079
10515920 0.176538433449701
11173550 0.182517133075224
11935525 0.186842160508411
12679738 0.189807648439687
13384698 0.191681967043158
14116088 0.195293171665187
14772887 0.196290191104493
15558605 0.195109016830147
16324249 0.195044528076616
17068580 0.196621438247307
17845372 0.195495778096115
18466548 0.193705397985705
19134888 0.192073783028795
19820094 0.190000215259281
20431254 0.187736083699896
20972820 0.186902926228307
21512296 0.187899220240367
22070524 0.185143866219201
22691414 0.183051617486818
23179666 0.181893138794808
23633144 0.181685936964071
24062100 0.180284867391691
24411224 0.178740080394166
24769926 0.177896889128779
25141960 0.177020836339553
25548760 0.176906982484708
25957200 0.17638052370475
26300288 0.176480059845281
26504228 0.176338956997164
26837524 0.174585657950694
27163056 0.173666706166389
27421310 0.173031636435328
};
\addplot [thick, color0, opacity=0.4]
table {%
27421310 0.212378415443505
};
\addplot [thick, color0]
table {%
27421310 0.212378415443505
29241978 0.213310940608572
31584578 0.208109662358843
34147504 0.203821525906863
36494240 0.199343009539195
38787372 0.194578351511492
41154776 0.190681307651651
43361168 0.18797970030584
45359856 0.184620796530003
47402496 0.180800246070132
49196656 0.178822652975734
50801440 0.177494783029841
52584412 0.174404388009719
53927804 0.173055208979018
55503776 0.171072647322056
57045520 0.16947423597679
58413856 0.167558643676558
59573664 0.16634855664559
60552740 0.165486104599421
61621636 0.164311396297681
62616744 0.163223343356685
63702584 0.162339198376109
64671696 0.1614372193394
65501088 0.160728011348467
};
\addplot [thick, color0, opacity=0.4]
table {%
21952276 0.213481832586462
24304066 0.21317407700738
27421310 0.212378415443505
};
\addplot [thick, color0, opacity=0.4]
table {%
65501088 0.194786536811069
};
\addplot [thick, color0]
table {%
65501088 0.194786536811069
67259920 0.193251952124525
74724928 0.188171053236046
81708528 0.184436166132334
88334648 0.179080073355003
94084560 0.176522957752557
99491560 0.173425103667605
104465712 0.169769499530629
109811712 0.166422312412897
114441208 0.163668311444659
118463360 0.161655673081801
121874664 0.160797876427674
125652376 0.159748578547803
129252896 0.159146144067526
132409536 0.159382183729945
135074560 0.159467474215586
137454576 0.15917434980573
140058576 0.159168275197347
141967392 0.159575841478491
143816160 0.159590249647591
145699632 0.159860229631541
};
\addplot [thick, color0, opacity=0.4]
table {%
47106184 0.212208152951398
53689056 0.205106769048831
65501088 0.194786536811069
};
\addplot [thick, black]
table {%
7125618.77811424 0.0238555273374017
7400580.23537354 0.0233700092535804
7783853.88093973 0.0239214900493956
8203447.8974359 0.0228506272906649
8728716.41176471 0.0237732884204081
9366414.42625265 0.0266257504997688
9946913.26569331 0.0258452236765634
10601025.3990148 0.0228200620180941
11273676.0225106 0.0219775503224666
11991577.3333333 0.0210762568468666
12722417.4660393 0.0179553761018707
13492461.1029412 0.0163937924456413
14335418.7962401 0.0149385573530402
15221740.0552704 0.0151730704360013
15892448.2571429 0.0131098090714127
16592819.4363144 0.014314853014833
17382477.0011376 0.0127630620935533
18061638.4158186 0.0146844753957722
18710095.8763435 0.0143113362752195
19329389 0.0137434137297912
20053034.5407036 0.0147941627255892
20642700.1366085 0.0156039929304115
21233120.0923223 0.0142822293224711
21839705.195231 0.0139473676502687
22333076.5636856 0.0120413191963022
22803949.5238403 0.0132274561815828
23289995.511524 0.0134312027525289
23752341.9954955 0.014081820182436
24378432.9274324 0.0139068781971686
};
\addplot [thick, black, opacity=0.4]
table {%
24378432.9274324 0.0139068781971686
24630222.1538462 0.0136664557700182
24963878.472173 0.0127545710634811
25412812.5203112 0.0123592739654119
25821995.3059544 0.0122664210521909
26182261.6890412 0.0121911464959267
26502622.3776224 0.0124845489440816
26844975.9272237 0.0128978410057097
27156992.7428124 0.0133544998307719
27386683.0826539 0.013227259323518
27701139.0634998 0.0132054859059419
};
\addplot [thick, black]
table {%
24378432.9274324 0.0139068781971686
24445963.6471636 0.0139773871215326
27114957.2485038 0.0212200531725956
29833869.7630945 0.0232908355236459
32193888.3221878 0.0254787849390905
34902012.545505 0.0248021448083605
37505023 0.0275524263099733
39965614.9145572 0.0303387551364075
42091339.3743385 0.0306162329647368
44270834.2128857 0.0295236018713796
46139542.7845058 0.02765607138286
49463342.7636604 0.026992242601355
};
\addplot [thick, black, opacity=0.4]
table {%
15866300.2508412 0.0156761989545932
17670232.3676155 0.0107801765200495
19641217.800252 0.00975543286358508
24378432.9274324 0.0139068781971686
};
\addplot [thick, black, opacity=0.4]
table {%
49463342.7636604 0.026992242601355
50171590.9730974 0.0269163886867993
52056350.5428953 0.0252962679097141
53776141.9914138 0.0246213221295425
55140028.6506293 0.0252780040640739
56692586.1654135 0.0256418204186534
58068878.1313194 0.0245494510658766
59322223.8366617 0.0238139661393015
60628822.0878741 0.0221708823955997
61616735.0994646 0.021463349531605
62521620.4327495 0.0212965040287153
63512250.9704391 0.0209904369559257
64602786.9168033 0.0202181206274257
65517615.8154501 0.0186642087708356
66434312.5786651 0.0175356078065021
};
\addplot [thick, black]
table {%
49463342.7636604 0.026992242601355
49827344.4937234 0.0271820944920182
56990304.4845342 0.0290630959644914
64233524.5772711 0.0296611581936479
71038681 0.0283886822834611
77993048.4721301 0.0259745495930314
84715634.1080382 0.0277459469929337
91594311.8345746 0.0275373440876603
97614458.5695152 0.0237703320637345
103256333.376398 0.0226670575276017
108160176.534447 0.0243544781818986
113064533.113946 0.0249755856648087
117581385.339844 0.0249558285847306
121380453.806219 0.024272326387465
124818427.473434 0.0229147824421525
128518258.319017 0.0217599911823869
131570776.498282 0.0209537503376603
134615441.88127 0.020195943941176
137322823.854245 0.0198008572712541
139457809.29821 0.0195520016804337
141946674.679242 0.0194208218708634
144026238.203987 0.0196415310993791
146046644.008189 0.0199018704548478
147637730.35694 0.0202818020954728
};
\addplot [thick, black, opacity=0.4]
table {%
31276498.1800991 0.0256029928341508
37026894.5865197 0.0232933746472001
49463342.7636604 0.026992242601355
};
\addplot [semithick, red, dotted, mark=*, mark size=1, mark options={solid}]
table {%
10020784 0.0436130916237831
20250148 0.0813491616358231
31707160 0.0972212723827025
44391820 0.104533269165434
73444084 0.0944896605248973
107406940 0.0941999876169305
146280388 0.0893585283668399
190064428 0.101241570053398
238759060 0.153297982804741
};
\addplot [semithick, blue, dashed, mark=square*, mark size=1, mark options={solid}]
table {%
12585316 0.0781401724364308
26798436 0.131407951038215
41011556 0.11369190828443
55224676 0.119475737234394
83650916 0.122380159324937
112077156 0.104896311524089
140503396 0.182920041335952
197355876 0.18403668255756
254208356 0.192096498747122
};
\end{axis}

\end{tikzpicture}
  \end{subfigure}\\[-8pt]
  \begin{subfigure}[b]{.24\textwidth}
    \raggedright
    \tikz\node[font=\bf,fill=C0,rounded corners=1pt]{ImageNet};\\
\begin{tikzpicture}

\definecolor{color0}{rgb}{1,0.498039215686275,0.0549019607843137}

\begin{axis}[
legend cell align={left},
legend style={
  font=\tiny,
  fill opacity=0.8,
  draw opacity=1,
  text opacity=1,
  at={(0.97,0.03)},
  anchor=south east,
  draw=white!80!black
},
tick align=outside,
tick pos=left,
x grid style={white!69.0196078431373!black},
xlabel={FLOPs},
xmajorgrids,
xmin=159159500.973725, xmax=4288393083.57268,
xtick style={color=black},
xtick={0,500000000,1000000000,1500000000,2000000000,2500000000,3000000000,3500000000,4000000000,4500000000},
xticklabels={0.0,0.5,1.0,1.5,2.0,2.5,3.0,3.5,4.0,4.5},
y grid style={white!69.0196078431373!black},
ylabel={Top-1 accuracy (\%) \(\displaystyle \rightarrow\)},
ymajorgrids,
ymin=57.2826, ymax=76.4094,
ytick style={color=black}
]
\addplot [thick, color0]
table {%
346995744 58.152
357505248 58.802
371037504 59.456
386024896 60.166
404027264 60.966
424161568 61.802
447164320 62.704
473096832 63.65
500726976 64.524
529794336 65.53
559776896 66.322
592046336 67.304
625636480 68.148
658033472 68.72
691794240 69.368
725067776 69.864
757585856 70.27
787494912 70.53
817910784 70.746
847693504 70.942
875857408 71.034
902254464 71.14
925686656 71.188
949450176 71.346
971245888 71.37
990896896 71.434
1010342464 71.498
1028224896 71.502
1044511360 71.568
1059933632 71.594
1073724416 71.646
1097831516.99615 71.6848311802666
};
\addlegendentry{MSDNet (vanilla)}
\addplot [thick, color0, opacity=0.4, forget plot]
table {%
1097831516.99615 71.6848311802666
1101383040 71.68
1113404800 71.708
1124635392 71.714
1135067136 71.728
1144408576 71.73
1152757376 71.74
1160763264 71.738
};
\addplot [thick, color0, forget plot]
table {%
1097831516.99615 71.6848311802666
1107946752 71.772
1182614784 72.392
1256653184 72.814
1327144064 73.138
1397747712 73.488
1464224000 73.706
1528752128 73.86
1591923072 73.986
1652542976 74.084
1709289216 74.206
1764005248 74.244
1845598371.95577 74.2954750602472
};
\addplot [thick, color0, opacity=0.4, forget plot]
table {%
839445568 68.65
900612736 69.582
969176704 70.474
1097831516.99615 71.6848311802666
};
\addplot [thick, color0, opacity=0.4, forget plot]
table {%
1845598371.95577 74.2954750602472
1863520512 74.3
1909294848 74.318
1950706688 74.314
1990510848 74.282
2029310720 74.292
2063007232 74.308
2095977472 74.302
2126244352 74.298
2155067392 74.304
2180452864 74.28
2206156544 74.292
2229196800 74.284
2251187456 74.262
2271287296 74.252
2290162688 74.24
};
\addplot [thick, color0, forget plot]
table {%
1845598371.95577 74.2954750602472
1845984256 74.296
1917285248 74.342
1988681856 74.382
2058906368 74.412
2122890752 74.45
2182199552 74.518
2237765632 74.5
2289782272 74.492
2338529536 74.504
2385252096 74.534
2428157184 74.496
2468098560 74.498
2504747008 74.518
2539344896 74.528
2572972544 74.532
2603235840 74.516
2633176576 74.5
2660816896 74.498
2687334400 74.5
2712068864 74.476
2735580160 74.484
};
\addplot [thick, color0, opacity=0.4, forget plot]
table {%
1517935744 73.77
1603648000 73.954
1686444672 74.07
1845598371.95577 74.2954750602472
};
\addplot [thick, black]
table {%
346851936.546405 58.164
358240994.393484 58.862
371561891.290908 59.526
387283553.083227 60.28
405721826.234604 61.09
426178884.127289 61.94
449012787.667885 62.792
474567282.376334 63.71
503753193.132286 64.746
532124072.322581 65.682
563224531.628758 66.626
595474294.48855 67.47
628851165.624025 68.328
660765042.032743 68.962
694380498.389245 69.502
726993896.690148 70.084
759002490.957051 70.418
790152822.690972 70.756
820056311.433738 70.986
848674057.2 71.146
876724519.72255 71.274
903988027.459305 71.388
927802389.20633 71.482
950815721.003225 71.558
971727560.846264 71.614
991760415.566542 71.668
1011470188.09799 71.736
1029988413.25742 71.754
1046120110.08454 71.802
1062020386.73934 71.81
1076851269.9231 71.864
1097482897.86871 71.905445755678
};
\addlegendentry{Our model}
\addplot [thick, black, opacity=0.4, forget plot]
table {%
1097482897.86871 71.905445755678
1103566167.16458 71.914
1115966762.38387 71.922
1127059956.07717 71.928
1137087847.46256 71.936
1145969206.73664 71.948
1155260562.64469 71.952
1163464362.18028 71.95
};
\addplot [thick, black, forget plot]
table {%
1097482897.86871 71.905445755678
1117425765.40606 72.104
1189177793.94418 72.642
1261930395.92574 73.046
1332769334.48929 73.36
1402404844.12197 73.694
1472095937.04383 73.912
1544398035.79929 74.1257031274826
};
\addplot [thick, black, opacity=0.4, forget plot]
table {%
845770492.650127 68.816
910685328.322581 69.858
975758820.214837 70.664
1097482897.86871 71.905445755678
};
\addplot [thick, black, opacity=0.4, forget plot]
table {%
1544398035.79929 74.1257031274826
1600815543.6 74.304
1660960163.86695 74.422
1719172900.16219 74.55
1773697314.87405 74.612
1825049064.52161 74.668
1869740040.85769 74.69
1914463953.90353 74.696
1956221245.24056 74.714
1995166563.62023 74.736
2032532466.78295 74.744
2067156901.27962 74.76
2102164156.04338 74.762
2133557369.44864 74.766
2162584477.60515 74.766
2188856035.65818 74.766
2212754687.89394 74.766
2236651934.47951 74.766
2257912162.69482 74.766
2278258918.53995 74.766
2296954298.62362 74.766
};
\addplot [thick, black, forget plot]
table {%
1544398035.79929 74.1257031274826
1610850518.87006 74.344
1693814095.65563 74.526
1773504663.78653 74.64
1851103257.39458 74.736
1925961398 74.818
1996971719.26114 74.872
2064973721.42035 74.898
2129527265.63336 74.936
2188041138.00258 74.974
2245596908.7149 75.004
2297799479.28266 75.038
2345732206.54187 75.052
2391204763.78863 75.056
2433258219.23197 75.06
2473269923.80095 75.06
2511630368.25243 75.06
2546793417.47293 75.062
2578100497.84147 75.062
2608692337.51209 75.062
2637502819.47805 75.062
2664672099.29004 75.062
2690971115.31399 75.062
2715956083.42889 75.062
2738007920.0796 75.062
};
\addplot [thick, black, opacity=0.4, forget plot]
table {%
1264257988.27897 72.896
1352017421.18766 73.376
1439588540.10321 73.786
1544398035.79929 74.1257031274826
};
\addplot [semithick, red, dotted, mark=*, mark size=1, mark options={solid}]
table {%
924089448 68.038
1383862344 69.72
1654405592 71.892
2530169384 73.358
2852288744 74.784
};
\addlegendentry{DenseNets}
\addplot [semithick, blue, dashed, mark=square*, mark size=1, mark options={solid}]
table {%
892617192 62.824
1818213864 69.742
2743810536 71.62
3669156328 72.902
4100700648 75.54
};
\addlegendentry{ResNets}
\end{axis}

\end{tikzpicture}
  \end{subfigure}
  \hfill
  \begin{subfigure}[b]{.24\textwidth}
    \raggedleft
\begin{tikzpicture}

\definecolor{color0}{rgb}{1,0.498039215686275,0.0549019607843137}

\begin{axis}[
tick align=outside,
tick pos=left,
x grid style={white!69.0196078431373!black},
xlabel={FLOPs},
xmajorgrids,
xmin=159159500.973725, xmax=4288393083.57268,
xtick style={color=black},
xtick={0,500000000,1000000000,1500000000,2000000000,2500000000,3000000000,3500000000,4000000000,4500000000},
xticklabels={0.0,0.5,1.0,1.5,2.0,2.5,3.0,3.5,4.0,4.5},
y grid style={white!69.0196078431373!black},
ylabel={Top-5 accuracy (\%) \(\displaystyle \rightarrow\)},
ymajorgrids,
ymin=80.8317039093018, ymax=93.2462998138428,
ytick style={color=black}
]
\addplot [thick, color0]
table {%
346995744 81.4540023803711
357505248 82.0800018310547
371037504 82.7280044555664
386024896 83.4140014648438
404027264 84.1220016479492
424161568 84.7800064086914
447164320 85.390007019043
473096832 86.0720062255859
500726976 86.5900039672852
529794336 87.1440048217773
559776896 87.5820007324219
592046336 88.0560073852539
625636480 88.4240036010742
658033472 88.7100067138672
691794240 88.9180068969727
725067776 89.0760040283203
757585856 89.1900024414062
787494912 89.3320007324219
817910784 89.4580078125
847693504 89.5700073242188
875857408 89.6340026855469
902254464 89.7160034179688
941631761.111556 89.8876826426638
};
\addplot [thick, color0, opacity=0.4]
table {%
941631761.111556 89.8876826426638
949450176 89.9140014648438
971245888 89.9460067749023
990896896 90.0400009155273
1010342464 90.068000793457
1028224896 90.10400390625
1044511360 90.1500015258789
1059933632 90.156005859375
1073724416 90.1900024414062
1088150784 90.1880035400391
1101383040 90.2180023193359
1113404800 90.234001159668
1124635392 90.2300033569336
1135067136 90.2280044555664
1144408576 90.2380065917969
1152757376 90.2420043945312
1160763264 90.2520065307617
};
\addplot [thick, color0]
table {%
941631761.111556 89.8876826426638
969176704 90.0540008544922
1037393280 90.3560028076172
1107946752 90.5840072631836
1182614784 90.8100051879883
1256653184 90.9980010986328
1327144064 91.0920028686523
1397747712 91.2020034790039
1464224000 91.3340072631836
1528752128 91.3820037841797
1591923072 91.4520034790039
1652542976 91.5160064697266
1709289216 91.568000793457
1764005248 91.6160049438477
1815991552 91.6380081176758
1863520512 91.6700057983398
1909294848 91.6780014038086
1950706688 91.7200012207031
1990510848 91.7460021972656
2029310720 91.7680053710938
2063007232 91.7840042114258
2095977472 91.8060073852539
2126244352 91.8200073242188
2155067392 91.8140029907227
2180452864 91.8200073242188
2206156544 91.8320007324219
2229196800 91.8440017700195
2251187456 91.8600006103516
2271287296 91.8560028076172
2290162688 91.85400390625
};
\addplot [thick, color0, opacity=0.4]
table {%
729799424 88.3640060424805
782145664 88.8360061645508
839445568 89.2800064086914
941631761.111556 89.8876826426638
};
\addplot [thick, color0, opacity=0.4]
table {%
2290162688 91.7062666452691
};
\addplot [thick, color0]
table {%
2290162688 91.7062666452691
2338529536 91.7400054931641
2385252096 91.7700042724609
2428157184 91.8000030517578
2468098560 91.7980041503906
2504747008 91.8060073852539
2539344896 91.8240051269531
2572972544 91.8400039672852
2603235840 91.838005065918
2633176576 91.838005065918
2660816896 91.8360061645508
2687334400 91.8220062255859
2712068864 91.8360061645508
2735580160 91.8300018310547
};
\addplot [thick, color0, opacity=0.4]
table {%
2122890752 91.672004699707
2182199552 91.6980056762695
2237765632 91.7000045776367
2290162688 91.7062666452691
};
\addplot [thick, black]
table {%
346851936.546405 81.3960037231445
358240994.393484 82.068000793457
371561891.290908 82.7020034790039
387283553.083227 83.3620071411133
405721826.234604 84.0460052490234
426178884.127289 84.6800003051758
449012787.667885 85.2620010375977
474567282.376334 85.8860015869141
503753193.132286 86.5280075073242
532124072.322581 87.0480041503906
563224531.628758 87.5100021362305
595474294.48855 87.9480056762695
628851165.624025 88.2800064086914
660765042.032743 88.5180053710938
694380498.389245 88.786003112793
726993896.690148 88.9900054931641
759002490.957051 89.1180038452148
790152822.690972 89.2480010986328
820056311.433738 89.3940048217773
848674057.2 89.5160064697266
876724519.72255 89.6020050048828
903988027.459305 89.6800079345703
927802389.20633 89.7920074462891
950815721.003225 89.8460006713867
971727560.846264 89.8840026855469
991760415.566542 89.9280014038086
1027310596.04842 89.9865602386881
};
\addplot [thick, black, opacity=0.4]
table {%
1027310596.04842 89.9865602386881
1029988413.25742 89.9880065917969
1046120110.08454 90.0260009765625
1062020386.73934 90.0540008544922
1076851269.9231 90.0700073242188
1090765638.77199 90.0960006713867
1103566167.16458 90.1080017089844
1115966762.38387 90.120002746582
1127059956.07717 90.1180038452148
1137087847.46256 90.120002746582
1145969206.73664 90.1220016479492
1155260562.64469 90.1360015869141
1163464362.18028 90.1440048217773
};
\addplot [thick, black]
table {%
1027310596.04842 89.9865602386881
1045309558.87162 90.088005065918
1117425765.40606 90.3460006713867
1248661952.23575 90.7735014535924
1261930395.92574 90.818000793457
1332769334.48929 91.0120010375977
1402404844.12197 91.1880035400391
1472095937.04383 91.3460006713867
1536264935.2308 91.4660034179688
1600815543.6 91.5460052490234
1660960163.86695 91.6080017089844
1719172900.16219 91.6940078735352
1773697314.87405 91.7260055541992
1825049064.52161 91.786003112793
1869740040.85769 91.8160018920898
1914463953.90353 91.8500061035156
1956221245.24056 91.8680038452148
1995166563.62023 91.9020080566406
2032532466.78295 91.9120025634766
2067156901.27962 91.9340057373047
2102164156.04338 91.9580078125
2133557369.44864 91.9640045166016
2162584477.60515 91.9820022583008
2188856035.65818 92.0040054321289
2212754687.89394 92.0060043334961
2236651934.47951 92.0100021362305
2257912162.69482 92.0140075683594
2278258918.53995 92.0240020751953
2296954298.62362 92.036003112793
};
\addplot [thick, black, opacity=0.4]
table {%
788719851.548012 88.6140060424805
845770492.650127 89.0200042724609
910685328.322581 89.3740005493164
1027310596.04842 89.9865602386881
};
\addplot [thick, black]
table {%
2297799479.28266 92.0000076293945
2345732206.54187 92.0320053100586
2391204763.78863 92.052001953125
2433258219.23197 92.0660018920898
2473269923.80095 92.0760040283203
2511630368.25243 92.0860061645508
2546793417.47293 92.1000061035156
2578100497.84147 92.0980072021484
2608692337.51209 92.1120071411133
2637502819.47805 92.1280059814453
2664672099.29004 92.1320037841797
2690971115.31399 92.1340026855469
2715956083.42889 92.1360015869141
2738007920.0796 92.1360015869141
};
\addplot [thick, black, opacity=0.4]
table {%
2129527265.63336 91.922004699707
2188041138.00258 91.9360046386719
2245596908.7149 91.974006652832
2297799479.28266 92.0000076293945
};
\addplot [semithick, red, dotted, mark=*, mark size=1, mark options={solid}]
table {%
924089448 88.36
1383862344 89.544
1654405592 90.692
2530169384 91.416
2852288744 92.29
};
\addplot [semithick, blue, dashed, mark=square*, mark size=1, mark options={solid}]
table {%
892617192 84.406
1818213864 89.218
2743810536 90.52
3669156328 91.336
4100700648 92.682
};
\end{axis}

\end{tikzpicture}
  \end{subfigure}
  \hfill
  \begin{subfigure}[b]{.24\textwidth}
    \raggedleft
\begin{tikzpicture}

\definecolor{color0}{rgb}{1,0.498039215686275,0.0549019607843137}

\begin{axis}[
tick align=outside,
tick pos=left,
x grid style={white!69.0196078431373!black},
xlabel={FLOPs},
xmajorgrids,
xmin=159159500.973725, xmax=4288393083.57268,
xtick style={color=black},
xtick={0,500000000,1000000000,1500000000,2000000000,2500000000,3000000000,3500000000,4000000000,4500000000},
xticklabels={0.0,0.5,1.0,1.5,2.0,2.5,3.0,3.5,4.0,4.5},
y grid style={white!69.0196078431373!black},
ylabel={\(\displaystyle \leftarrow\) NLPD},
ymajorgrids,
ymin=0.924073558322906, ymax=1.82806794504929,
ytick style={color=black},
ytick={0.8,1,1.2,1.4,1.6,1.8,2},
yticklabels={0.8,1.0,1.2,1.4,1.6,1.8,2.0}
]
\addplot [thick, color0]
table {%
346995744 1.78210949897766
357505248 1.75376152992249
371037504 1.72032904624939
386024896 1.68600738048553
404027264 1.64881908893585
424161568 1.60951840877533
447164320 1.57223677635193
473096832 1.53019785881042
500726976 1.49017155170441
529794336 1.45168936252594
559776896 1.41946911811829
592046336 1.38632082939148
625636480 1.3550568819046
658033472 1.32907557487488
691794240 1.30359280109406
725067776 1.28176701068878
757585856 1.26254236698151
787494912 1.24738132953644
817910784 1.23340499401093
847693504 1.22104120254517
875857408 1.21183693408966
902254464 1.20330560207367
925686656 1.19542002677917
949450176 1.18688917160034
971245888 1.18175208568573
990896896 1.17559409141541
1010342464 1.17203426361084
1028224896 1.16870820522308
1044511360 1.16449284553528
1059933632 1.16207659244537
1073724416 1.15940606594086
1088150784 1.15726542472839
1101383040 1.15579450130463
1113404800 1.1537811756134
1124635392 1.15200197696686
1135067136 1.15124118328094
1144408576 1.14986205101013
1152757376 1.14876353740692
1160763264 1.14809155464172
};
\addplot [thick, color0, opacity=0.4]
table {%
1160763264 1.15921014899309
};
\addplot [thick, color0]
table {%
1160763264 1.15921014899309
1182614784 1.1520688533783
1256653184 1.13248074054718
1327144064 1.11734855175018
1397747712 1.10300576686859
1464224000 1.09104251861572
1528752128 1.08419585227966
1591923072 1.07567858695984
1652542976 1.06900930404663
1709289216 1.06251573562622
1764005248 1.05671322345734
1815991552 1.05422413349152
1863520512 1.05127739906311
1909294848 1.04848551750183
1950706688 1.04648792743683
1990510848 1.04417824745178
2029310720 1.04216742515564
2063007232 1.04035353660583
2095977472 1.03934979438782
2126244352 1.03796863555908
2155067392 1.03790509700775
2180452864 1.03696763515472
2206156544 1.03572022914886
2229196800 1.03457224369049
2251187456 1.03372192382812
2271287296 1.03352701663971
2290162688 1.03359651565552
};
\addplot [thick, color0, opacity=0.4]
table {%
969176704 1.22868597507477
1037393280 1.20037877559662
1160763264 1.15921014899309
};
\addplot [thick, color0, opacity=0.4]
table {%
2290162688 1.0508841483725
};
\addplot [thick, color0]
table {%
2290162688 1.0508841483725
2338529536 1.04882550239563
2385252096 1.04741275310516
2428157184 1.04617464542389
2468098560 1.04582560062408
2504747008 1.04511070251465
2539344896 1.04399299621582
2572972544 1.0427577495575
2603235840 1.04243493080139
2633176576 1.04191088676453
2660816896 1.04126238822937
2687334400 1.04122412204742
2712068864 1.04093253612518
2735580160 1.04002892971039
};
\addplot [thick, color0, opacity=0.4]
table {%
1917285248 1.06691539287567
1988681856 1.06198227405548
2058906368 1.05952346324921
2122890752 1.05744516849518
2182199552 1.05459976196289
2237765632 1.05304491519928
2290162688 1.0508841483725
};
\addplot [thick, black]
table {%
346851936.546405 1.78697729110718
358240994.393484 1.75766909122467
371561891.290908 1.72727859020233
387283553.083227 1.69409120082855
405721826.234604 1.65897750854492
426178884.127289 1.6229887008667
449012787.667885 1.58679986000061
474567282.376334 1.54951119422913
503753193.132286 1.50797736644745
532124072.322581 1.47255671024323
563224531.628758 1.43841814994812
595474294.48855 1.40696549415588
628851165.624025 1.37604582309723
660765042.032743 1.3510913848877
694380498.389245 1.3258148431778
726993896.690148 1.30303311347961
759002490.957051 1.2842972278595
790152822.690972 1.26729893684387
820056311.433738 1.25304269790649
848674057.2 1.24004793167114
876724519.72255 1.22832453250885
903988027.459305 1.21837043762207
927802389.20633 1.20872116088867
950815721.003225 1.20123505592346
971727560.846264 1.19471073150635
991760415.566542 1.18944442272186
1011470188.09799 1.18482565879822
1029988413.25742 1.18104839324951
1046120110.08454 1.17750096321106
1062020386.73934 1.17441129684448
1076851269.9231 1.17195105552673
1090765638.77199 1.16988456249237
1103566167.16458 1.16753995418549
1115966762.38387 1.16581857204437
1127059956.07717 1.1643328666687
1137087847.46256 1.16265487670898
1145969206.73664 1.16109549999237
1155260562.64469 1.15948152542114
1163464362.18028 1.15831863880157
};
\addplot [thick, black, opacity=0.4]
table {%
1163464362.18028 1.17094789129424
};
\addplot [thick, black]
table {%
1163464362.18028 1.17094789129424
1117204437.20966 1.18651593408933
1117425765.40606 1.18643450737
1189177793.94418 1.16229832172394
1261930395.92574 1.14080393314362
1332769334.48929 1.12148118019104
1402404844.12197 1.10352861881256
1472095937.04383 1.08873057365417
1536264935.2308 1.07755672931671
1600815543.6 1.06688177585602
1660960163.86695 1.05737996101379
1719172900.16219 1.04928827285767
1773697314.87405 1.04409384727478
1825049064.52161 1.03795158863068
1869740040.85769 1.03359663486481
1914463953.90353 1.02939331531525
1956221245.24056 1.02598106861115
1995166563.62023 1.0229743719101
2032532466.78295 1.02009010314941
2067156901.27962 1.01791751384735
2102164156.04338 1.01606285572052
2133557369.44864 1.01423609256744
2162584477.60515 1.01239669322968
2188856035.65818 1.01085543632507
2212754687.89394 1.00946950912476
2236651934.47951 1.00829589366913
2257912162.69482 1.00740301609039
2278258918.53995 1.0062974691391
2296954298.62362 1.0054680109024
};
\addplot [thick, black, opacity=0.4]
table {%
975758820.214837 1.24291253089905
1045309558.87162 1.21296608448029
1163464362.18028 1.17094789129424
};
\addplot [thick, black]
table {%
2297799479.28266 1.00815057754517
2345732206.54187 1.00575792789459
2391204763.78863 1.0034590959549
2433258219.23197 1.00158524513245
2473269923.80095 1.00002110004425
2511630368.25243 0.998584747314453
2546793417.47293 0.997495234012604
2578100497.84147 0.996750831604004
2608692337.51209 0.995694220066071
2637502819.47805 0.995008766651154
2664672099.29004 0.994107007980347
2690971115.31399 0.993364989757538
2715956083.42889 0.992761552333832
2738007920.0796 0.992346823215485
};
\addplot [thick, black, opacity=0.4]
table {%
1996971719.26114 1.02878034114838
2064973721.42035 1.02337801456451
2129527265.63336 1.01895236968994
2188041138.00258 1.01493763923645
2245596908.7149 1.01136565208435
2297799479.28266 1.00815057754517
};
\addplot [semithick, red, dotted, mark=*, mark size=1, mark options={solid}]
table {%
924089448 1.29369465522766
1383862344 1.20918216156006
1654405592 1.1232120275116
2530169384 1.05259418296814
2852288744 0.997138261070251
};
\addplot [semithick, blue, dashed, mark=square*, mark size=1, mark options={solid}]
table {%
892617192 1.57189665672302
1818213864 1.22871706542969
2743810536 1.13310672416687
3669156328 1.07734281196594
4100700648 0.965164212265015
};
\end{axis}

\end{tikzpicture}
  \end{subfigure}
  \hfill
  \begin{subfigure}[b]{.24\textwidth}
    \raggedleft
\begin{tikzpicture}

\definecolor{color0}{rgb}{1,0.498039215686275,0.0549019607843137}

\begin{axis}[
tick align=outside,
tick pos=left,
x grid style={white!69.0196078431373!black},
xlabel={FLOPs},
xmajorgrids,
xmin=159159500.973725, xmax=4288393083.57268,
xtick style={color=black},
xtick={0,500000000,1000000000,1500000000,2000000000,2500000000,3000000000,3500000000,4000000000,4500000000},
xticklabels={0.0,0.5,1.0,1.5,2.0,2.5,3.0,3.5,4.0,4.5},
y grid style={white!69.0196078431373!black},
ylabel={\(\displaystyle \leftarrow\) ECE},
ymajorgrids,
ymin=0.00466249343116208, ymax=0.0651734114294064,
ytick style={color=black},
ytick={0.02,0.04,0.06},
yticklabels={0.02,0.04,0.06},
scaled y ticks=false
]
\addplot [thick, color0]
table {%
346995744 0.0179202110043177
357505248 0.0187155629480387
371037504 0.0181050945008768
386024896 0.0171862575740284
404027264 0.0160647777393414
424161568 0.013994854138721
447164320 0.0125188376608413
473096832 0.0104472668982188
500726976 0.0100885464974925
529794336 0.00881249547140602
559776896 0.0104059022022115
592046336 0.0103572842101984
625636480 0.00940299750819235
658033472 0.0126144027549741
691794240 0.0148999874833922
725067776 0.0174947638051882
757585856 0.01986702026521
787494912 0.0224946897772184
817910784 0.024359909233402
847693504 0.0260442520404363
875857408 0.0272300172565952
902254464 0.0279418872679211
925686656 0.0288538510028013
949450176 0.0281799440920459
971245888 0.0284392000706225
990896896 0.0284607342384885
1010342464 0.0283531191711481
1028224896 0.0283891080501053
1044511360 0.0278115861236902
1059933632 0.0275961829508807
1073724416 0.0270607288213879
1088150784 0.0265118091515883
1101383040 0.0265826766578056
1113404800 0.0262512704935495
1124635392 0.0260661834788209
1135067136 0.0258639818639879
1148482323.21711 0.0256209533407938
};
\addplot [thick, color0, opacity=0.4]
table {%
1148482323.21711 0.0256209533407938
1152757376 0.0255088954165557
1160763264 0.0254459705740746
};
\addplot [thick, color0]
table {%
1148482323.21711 0.0256209533407938
1182614784 0.0269426247098855
1256653184 0.0305628063855222
1327144064 0.0337836450332522
1397747712 0.0358432204883722
1464224000 0.037524733191197
1528752128 0.0394691661151949
1591923072 0.0411267899430254
1652542976 0.0422530836595431
1709289216 0.0422937040169331
1764005248 0.0429904775387378
1815991552 0.0436260337679288
1863520512 0.0444922829261803
1909294848 0.0446407843496357
1950706688 0.0447493386872316
1990510848 0.0451942685938028
2029310720 0.0448645627175795
2063007232 0.0445073377352209
2095977472 0.0441903984829437
2126244352 0.0438220194899715
2155067392 0.0434253575053281
2180452864 0.0432913122265247
2206156544 0.0426918052140176
2229196800 0.0423404792102939
2251187456 0.0420759056390698
2271287296 0.0417541898895804
2290162688 0.041460137945496
};
\addplot [thick, color0, opacity=0.4]
table {%
839445568 0.00965567375579384
900612736 0.0124628919370532
969176704 0.0160854908534938
1037393280 0.0200914711272131
1148482323.21711 0.0256209533407938
};
\addplot [thick, color0, opacity=0.4]
table {%
2290162688 0.0523300028946609
};
\addplot [thick, color0]
table {%
2290162688 0.0523300028946609
2338529536 0.0519367817872256
2385252096 0.0514140949301539
2428157184 0.0515433833226606
2468098560 0.0512501960230913
2504747008 0.0506900827797395
2539344896 0.0505212630050424
2572972544 0.0502722391821809
2603235840 0.0502374699602701
2633176576 0.0500094916253453
2660816896 0.0498738510939188
2687334400 0.049498303572305
2712068864 0.0495502695319032
2735580160 0.0493161843237483
};
\addplot [thick, color0, opacity=0.4]
table {%
1845984256 0.0507569109975501
1917285248 0.0513921894750259
1988681856 0.051943520232354
2058906368 0.0520242840717555
2122890752 0.0521672487244603
2182199552 0.0519287701293679
2237765632 0.0523252970210236
2290162688 0.0523300028946609
};
\addplot [thick, black]
table {%
346851936.546405 0.0293541674056137
358240994.393484 0.0328408368704484
371561891.290908 0.0356438131570249
387283553.083227 0.0384065280945511
405721826.234604 0.0413678164348504
426178884.127289 0.0439517484998443
449012787.667885 0.0464645412750784
474567282.376334 0.0490107528209524
503753193.132286 0.0518144208213449
532124072.322581 0.054290176017296
563224531.628758 0.0569883596539729
595474294.48855 0.0591272059320276
628851165.624025 0.0611027465414726
660765042.032743 0.0618600981142853
694380498.389245 0.0617061094388087
726993896.690148 0.062422915156759
759002490.957051 0.0607164753883151
790152822.690972 0.0598142686234767
820056311.433738 0.0578938231100743
848674057.2 0.0556833028254498
876724519.72255 0.0534543184264259
903988027.459305 0.0516417260222341
927802389.20633 0.0499065741316963
950815721.003225 0.0481117652863006
971727560.846264 0.0466279929350723
991760415.566542 0.0452127449035492
1011470188.09799 0.043895274706489
1029988413.25742 0.0422005728437017
1046120110.08454 0.041133551991566
1062020386.73934 0.0397477405839315
1076851269.9231 0.0389140593595258
1090765638.77199 0.0380309679672292
1103566167.16458 0.0371791027853943
1115966762.38387 0.0360988321565442
1127059956.07717 0.0352103451485228
1137087847.46256 0.0344831162484761
1149341932.3835 0.0335827099371929
};
\addplot [thick, black, opacity=0.4]
table {%
1149341932.3835 0.0335827099371929
1155260562.64469 0.0331133891469028
1163464362.18028 0.0324214246240186
};
\addplot [thick, black]
table {%
1149341932.3835 0.0335827099371929
1189177793.94418 0.0340341597252477
1261930395.92574 0.0341143413825918
1332769334.48929 0.0340320164328702
1402404844.12197 0.0344925979444577
1472095937.04383 0.0336316212522676
1536264935.2308 0.0330134887724305
1600815543.6 0.032366501955741
1660960163.86695 0.031565128506397
1719172900.16219 0.0313326629195853
1773697314.87405 0.0306148491858692
1825049064.52161 0.0299488446548411
1869740040.85769 0.0292571511334707
1914463953.90353 0.0283705574738631
1956221245.24056 0.0276717974659481
1995166563.62023 0.0270522855639913
2032532466.78295 0.0265061393871615
2067156901.27962 0.0260885065227855
2102164156.04338 0.0255008922314371
2133557369.44864 0.0250700169147121
2162584477.60515 0.0246563831537556
2188856035.65818 0.0243169281224675
2212754687.89394 0.0240350743408606
2236651934.47951 0.023776780600767
2257912162.69482 0.0235179424161909
2278258918.53995 0.0233199990041976
2296954298.62362 0.0231387551449454
};
\addplot [thick, black, opacity=0.4]
table {%
788719851.548012 0.0269095576422942
845770492.650127 0.0277495686632482
910685328.322581 0.0300738172089186
975758820.214837 0.0306480457459491
1045309558.87162 0.0315674947913631
1149341932.3835 0.0335827099371929
};
\addplot [thick, black, opacity=0.4]
table {%
2296954298.62362 0.0266041852922198
};
\addplot [thick, black]
table {%
2296954298.62362 0.0266041852922198
2297799479.28266 0.026594133696798
2345732206.54187 0.0258686123399434
2391204763.78863 0.025328923939018
2433258219.23197 0.0248394046979707
2473269923.80095 0.0243891153665214
2511630368.25243 0.024052303664149
2546793417.47293 0.023677341326425
2578100497.84147 0.0233912677226171
2608692337.51209 0.0231542888880348
2637502819.47805 0.0229877485697089
2664672099.29004 0.0228171157997157
2690971115.31399 0.0226855273627533
2715956083.42889 0.0225840951179691
2738007920.0796 0.0225269375352559
};
\addplot [thick, black, opacity=0.4]
table {%
1996971719.26114 0.0317654384004725
2064973721.42035 0.0301794131520577
2129527265.63336 0.0290013401388912
2188041138.00258 0.028228445479652
2296954298.62362 0.0266041852922198
};
\addplot [semithick, red, dotted, mark=*, mark size=1, mark options={solid}]
table {%
924089448 0.0165507481447515
1383862344 0.0116737130537158
1654405592 0.0112626456661934
2530169384 0.0118924257137106
2852288744 0.0154113541252368
};
\addplot [semithick, blue, dashed, mark=square*, mark size=1, mark options={solid}]
table {%
892617192 0.0195234786491959
1818213864 0.0126059513697918
2743810536 0.026521215850036
3669156328 0.0313839649666444
4100700648 0.0286724907267321
};
\end{axis}

\end{tikzpicture}
  \end{subfigure}\\[-6pt]
  \begin{subfigure}[b]{.24\textwidth}
    \raggedright
    \tikz\node[font=\bf,fill=C0,rounded corners=1pt]{Caltech-256};\\
\begin{tikzpicture}

\definecolor{color0}{rgb}{1,0.498039215686275,0.0549019607843137}

\begin{axis}[
legend cell align={left},
legend style={
  font=\tiny,
  fill opacity=0.8,
  draw opacity=1,
  text opacity=1,
  at={(0.97,0.03)},
  anchor=south east,
  draw=white!80!black
},
tick align=outside,
tick pos=left,
x grid style={white!69.0196078431373!black},
xlabel={FLOPs},
xmajorgrids,
xmin=160314380.392992, xmax=4286743186.74319,
xtick style={color=black},
xtick={0,500000000,1000000000,1500000000,2000000000,2500000000,3000000000,3500000000,4000000000,4500000000},
xticklabels={0.0,0.5,1.0,1.5,2.0,2.5,3.0,3.5,4.0,4.5},
y grid style={white!69.0196078431373!black},
ylabel={Top-1 accuracy (\%) \(\displaystyle \rightarrow\)},
ymajorgrids,
ymin=53.544, ymax=66.656,
ytick style={color=black}
]
\addplot [thick, color0]
table {%
350043872 54.22
363169184 54.78
378871872 55.5
399106464 56.16
419256128 56.66
438078144 57.26
459118400 57.98
487951136 58.54
519883456 59.24
547457344 59.64
576095552 60.44
609939392 60.88
641712512 61.5
671926976 61.78
703289856 62
740006784 62.52
776731933.093015 62.7770610183927
};
\addlegendentry{MSDNet (vanilla)}
\addplot [thick, color0, opacity=0.4, forget plot]
table {%
776731933.093015 62.7770610183927
793556032 62.72
820766464 62.76
847834368 62.8
};
\addplot [thick, color0, forget plot]
table {%
776731933.093015 62.7770610183927
800943232 63.02
854569152 63.2
912004352 63.82
973277824 63.96
1041062464 64.48
1106610048 64.72
1208943100.68019 64.4159299457126
};
\addplot [thick, color0, opacity=0.4, forget plot]
table {%
621326656 61.58
660280640 61.88
706429696 62.38
776731933.093015 62.7770610183927
};
\addplot [thick, color0, opacity=0.4, forget plot]
table {%
1208943100.68019 64.4159299457126
1262928384 64.26
1337498496 64.28
1402526848 64.26
};
\addplot [thick, color0, forget plot]
table {%
1208943100.68019 64.4159299457126
1217613056 64.5
1301457920 64.64
1385014400 64.76
1470155264 64.92
1562782336 65.32
1645307648 65.2
1724982656 65.16
1801000192 65.12
1876115968 65
1947548928 64.86
2015549184 65
2091250304 65.14
2160307712 64.98
2212439296 64.88
2254626048 64.8
2302617856 64.88
2354772736 64.78
2398968320 64.78
2441886208 64.66
2486350848 64.56
2517301760 64.56
2552888832 64.48
2584985088 64.36
2615736832 64.28
2643860224 64.26
2671111168 64.28
2695002624 64.18
2714739712 64.16
2733680640 64.14
};
\addplot [thick, color0, opacity=0.4, forget plot]
table {%
985516224 62.88
1067997248 63.5
1208943100.68019 64.4159299457126
};
\addplot [thick, black]
table {%
347879326.136183 54.14
361727626.552966 54.72
375162379.739126 55.02
397002573.00128 55.64
414159506.715542 56.22
434178158.21416 56.6
457168729.465346 57.46
485193404.560621 58.36
508919294.319254 59.24
538296192.193548 59.9
572700420.392139 60.36
606902204.048577 61.22
640841690.595128 61.58
672645355.359634 61.86
703755366.737516 62.48
737182980.945264 63.02
765590797.142231 63.34
791882848.444775 63.58
820263168.940833 63.94
850212309.2 64.34
876102824.972457 64.38
901540139.818152 64.3
923664046.354734 64.36
946516830.704365 64.26
964395588.591147 64.2
981285181.41582 64.18
998996536.525652 64.24
1018002227.81026 64.32
1035067064.6887 64.34
1051586649.50711 64.38
1064085167.33591 64.46
1077417048.13744 64.52
1089882741.26516 64.6
1100909610.25237 64.58
1110445595.84832 64.62
1120330735.26257 64.62
1132233545.58206 64.64
1141316559.32981 64.68
1149065380.13846 64.68
};
\addlegendentry{Our model}
\addplot [thick, black, opacity=0.4, forget plot]
table {%
1149065380.13846 64.2561626003472
};
\addplot [thick, black, forget plot]
table {%
1149065380.13846 64.2561626003472
1177827707.27107 64.24
1241401776.27401 64.28
1323899090.298 64.2162368754691
};
\addplot [thick, black, opacity=0.4, forget plot]
table {%
917087528.193548 63.48
979892564.978218 63.7
1042815868.43164 63.96
1149065380.13846 64.2561626003472
};
\addplot [thick, black, opacity=0.4, forget plot]
table {%
1323899090.298 64.2162368754691
1386299646.30715 64.2
1454149866.79764 64.52
1514510576.7379 64.7
};
\addplot [thick, black, forget plot]
table {%
1323899090.298 64.2162368754691
1379151434.15876 64.42
1459289205.4301 64.7
1540591557.59027 64.8
1619778035.12518 65.04
1701705825.84081 65.12
1778102849.54033 65.2
1859458594.90168 65.46
1930174082 65.68
1996965960.51104 65.66
2062168905.7792 65.62
2122966586.78177 65.86
2180806055.70372 66.06
2237340776.45978 66.04
2291210901.13193 66.04
2333941754.96954 66.04
2380579218.34147 66.04
2425551477.83613 65.98
2462665690.56872 66
2497077257.66524 66
2532060506.83838 65.9
2573285423.94205 65.86
2600748017.38058 65.84
2626554379.24919 65.8
2652768747.09005 65.78
2683710462.15941 65.78
2705761200.11402 65.78
2725283946.03778 65.76
};
\addplot [thick, black, opacity=0.4, forget plot]
table {%
1128385031.41935 63.14
1208144577.6515 63.7
1323899090.298 64.2162368754691
};
\addplot [semithick, red, dotted, mark=*, mark size=1, mark options={solid}]
table {%
923732065 61.68
1383475241 63.44
1653911497 64.12
2529598017 64.76
2851527169 64
};
\addlegendentry{DenseNets}
\addplot [semithick, blue, dashed, mark=square*, mark size=1, mark options={solid}]
table {%
892236033 58.24
1817832705 61.72
2743429377 61.46
3668775169 63.08
4099178241 62.42
};
\addlegendentry{ResNets}
\end{axis}

\end{tikzpicture}
  \end{subfigure}
  \hfill
  \begin{subfigure}[b]{.24\textwidth}
    \raggedleft
\begin{tikzpicture}

\definecolor{color0}{rgb}{1,0.498039215686275,0.0549019607843137}

\begin{axis}[
tick align=outside,
tick pos=left,
x grid style={white!69.0196078431373!black},
xlabel={FLOPs},
xmajorgrids,
xmin=160314380.392992, xmax=4286743186.74319,
xtick style={color=black},
xtick={0,500000000,1000000000,1500000000,2000000000,2500000000,3000000000,3500000000,4000000000,4500000000},
xticklabels={0.0,0.5,1.0,1.5,2.0,2.5,3.0,3.5,4.0,4.5},
y grid style={white!69.0196078431373!black},
ylabel={Top-5 accuracy (\%) \(\displaystyle \rightarrow\)},
ymajorgrids,
ymin=72.4739990234375, ymax=83.1660003662109,
ytick style={color=black}
]
\addplot [thick, color0]
table {%
350043872 73.1199951171875
363169184 73.4599990844727
378871872 74.0199966430664
399106464 74.5400009155273
419256128 75
438078144 75.4199981689453
459118400 75.879997253418
487951136 76.1999969482422
519883456 76.7599945068359
547457344 77.0999984741211
576095552 77.7399978637695
609939392 77.9599990844727
641712512 78.379997253418
671926976 78.6599960327148
703289856 78.7399978637695
740006784 79.0599975585938
769968512 79.2999954223633
793556032 79.3399963378906
820766464 79.5199966430664
847834368 79.6599960327148
868118016 79.7599945068359
891813760 79.5199966430664
915454592 79.6199951171875
933698560 79.6800003051758
956083264 79.8600006103516
976659200 79.9799957275391
998404096 80.0999984741211
1015859200 80.0599975585938
1030473472 80.0400009155273
1047750784 80.0400009155273
1061431360 80.0800018310547
1073492096 80.0599975585938
1088581632 80.0199966430664
1101310464 80.0400009155273
1111330176 80.0400009155273
1123051520 80.0199966430664
1133661696 79.9799957275391
1143388416 79.879997253418
1150651776 79.879997253418
};
\addplot [thick, color0, opacity=0.4]
table {%
1150651776 80.4281225306875
};
\addplot [thick, color0]
table {%
1150651776 80.4281225306875
1169337343.12358 80.372171962947
};
\addplot [thick, color0, opacity=0.4]
table {%
973277824 80.0400009155273
1041062464 80.4000015258789
1150651776 80.4281225306875
};
\addplot [thick, color0, opacity=0.4]
table {%
1169337343.12358 80.372171962947
1186761088 80.3199996948242
1262928384 80.379997253418
};
\addplot [thick, color0]
table {%
1169337343.12358 80.372171962947
1217613056 80.5199966430664
1301457920 80.6800003051758
1385014400 80.5599975585938
1470155264 80.6199951171875
1562782336 80.4799957275391
1645307648 80.4599990844727
1724982656 80.5
1801000192 80.5199966430664
1876115968 80.6999969482422
1947548928 80.6199951171875
2015549184 80.6800003051758
2091250304 80.7999954223633
2160307712 80.8600006103516
2212439296 80.7999954223633
2254626048 80.879997253418
2302617856 80.9199981689453
2354772736 80.8199996948242
2398968320 80.6999969482422
2441886208 80.6599960327148
2486350848 80.6599960327148
2517301760 80.5999984741211
2552888832 80.6199951171875
2584985088 80.5999984741211
2615736832 80.5999984741211
2643860224 80.6800003051758
2671111168 80.6399993896484
2695002624 80.6599960327148
2714739712 80.6800003051758
2733680640 80.6599960327148
};
\addplot [thick, color0, opacity=0.4]
table {%
985516224 79.9799957275391
1067997248 80.0800018310547
1169337343.12358 80.372171962947
};
\addplot [thick, black]
table {%
347879326.136183 72.9599990844727
361727626.552966 73.4000015258789
375162379.739126 73.9000015258789
397002573.00128 74.5400009155273
414159506.715542 74.9000015258789
434178158.21416 75.3600006103516
457168729.465346 75.7799987792969
485193404.560621 76.5999984741211
508919294.319254 77.0800018310547
538296192.193548 77.6999969482422
572700420.392139 78.2799987792969
606902204.048577 78.879997253418
640841690.595128 79.1999969482422
672645355.359634 79.5
703755366.737516 79.7799987792969
737182980.945264 80.0599975585938
765590797.142231 80.1399993896484
791882848.444775 80.3600006103516
820263168.940833 80.4599990844727
850212309.2 80.5599975585938
876102824.972457 80.6800003051758
901540139.818152 80.8399963378906
923664046.354734 80.8399963378906
953204625.045865 80.8525177381627
};
\addplot [thick, black, opacity=0.4]
table {%
953204625.045865 80.8525177381627
964395588.591147 80.8399963378906
981285181.41582 80.8600006103516
};
\addplot [thick, black]
table {%
953204625.045865 80.8525177381627
979892564.978218 80.9799957275391
1042815868.43164 81.0999984741211
1106645282.37716 81.2200012207031
1236627754.83001 81.2324885936016
};
\addplot [thick, black, opacity=0.4]
table {%
789286261.732299 80.1800003051758
851181137.837095 80.3199996948242
953204625.045865 80.8525177381627
};
\addplot [thick, black, opacity=0.4]
table {%
1236627754.83001 81.2324885936016
1241401776.27401 81.2399978637695
1309436882.74441 81.0999984741211
};
\addplot [thick, black]
table {%
1236627754.83001 81.2324885936016
1297803529.839 81.5599975585938
1379151434.15876 81.9000015258789
1459289205.4301 81.9799957275391
1540591557.59027 82.0999984741211
1619778035.12518 82.1599960327148
1701705825.84081 82.3399963378906
1778102849.54033 82.5199966430664
1859458594.90168 82.6800003051758
1930174082 82.6800003051758
1996965960.51104 82.6599960327148
2062168905.7792 82.6199951171875
2122966586.78177 82.5999984741211
2180806055.70372 82.5599975585938
2237340776.45978 82.5599975585938
2291210901.13193 82.5
2333941754.96954 82.4799957275391
2380579218.34147 82.4399948120117
2425551477.83613 82.4399948120117
2462665690.56872 82.4199981689453
2497077257.66524 82.4399948120117
2532060506.83838 82.5
2573285423.94205 82.4799957275391
2600748017.38058 82.4599990844727
2626554379.24919 82.4399948120117
2652768747.09005 82.4599990844727
2683710462.15941 82.4599990844727
2705761200.11402 82.4599990844727
2725283946.03778 82.4799957275391
};
\addplot [thick, black, opacity=0.4]
table {%
1053295162.00452 80.4799957275391
1128385031.41935 80.7799987792969
1236627754.83001 81.2324885936016
};
\addplot [semithick, red, dotted, mark=*, mark size=1, mark options={solid}]
table {%
923732065 80.48
1383475241 81.56
1653911497 81.92
2529598017 81.94
2851527169 80.92
};
\addplot [semithick, blue, dashed, mark=square*, mark size=1, mark options={solid}]
table {%
892236033 77.12
1817832705 79.9
2743429377 78.2
3668775169 80.16
4099178241 80.08
};
\end{axis}

\end{tikzpicture}
  \end{subfigure}
  \hfill
  \begin{subfigure}[b]{.24\textwidth}
    \raggedleft
\begin{tikzpicture}

\definecolor{color0}{rgb}{1,0.498039215686275,0.0549019607843137}

\begin{axis}[
tick align=outside,
tick pos=left,
x grid style={white!69.0196078431373!black},
xlabel={FLOPs},
xmajorgrids,
xmin=160314380.392992, xmax=4286743186.74319,
xtick style={color=black},
xtick={0,500000000,1000000000,1500000000,2000000000,2500000000,3000000000,3500000000,4000000000,4500000000},
xticklabels={0.0,0.5,1.0,1.5,2.0,2.5,3.0,3.5,4.0,4.5},
y grid style={white!69.0196078431373!black},
ylabel={\(\displaystyle \leftarrow\) NLPD},
ymajorgrids,
ymin=1.52434445023537, ymax=2.47036229968071,
ytick style={color=black},
ytick={1.4,1.6,1.8,2,2.2,2.4,2.6},
yticklabels={1.4,1.6,1.8,2.0,2.2,2.4,2.6}
]
\addplot [thick, color0]
table {%
350043872 2.42736148834229
363169184 2.41786766052246
378871872 2.40111088752747
399106464 2.38611841201782
419256128 2.36306262016296
438078144 2.35245800018311
459118400 2.328533411026
487951136 2.31990742683411
519883456 2.28668594360352
547457344 2.27013659477234
576095552 2.23586416244507
609939392 2.22816467285156
641712512 2.19503974914551
671926976 2.16183519363403
703289856 2.13962697982788
740006784 2.10958433151245
769968512 2.09588813781738
793556032 2.08492469787598
820766464 2.06865048408508
847834368 2.05995225906372
868118016 2.04581689834595
891813760 2.04828882217407
915454592 2.03117489814758
933698560 2.02574872970581
959907376.309353 2.01081123506452
};
\addplot [thick, color0, opacity=0.4]
table {%
959907376.309353 2.01081123506452
976659200 2.00423312187195
998404096 1.99531996250153
1015859200 1.98846518993378
1030473472 1.98257887363434
1047750784 1.9824047088623
1061431360 1.97962093353271
1073492096 1.98135411739349
1088581632 1.98457479476929
1101310464 1.98483061790466
1111330176 1.98608887195587
1123051520 1.97875487804413
1133661696 1.97648453712463
1143388416 1.97726702690125
1150651776 1.97557640075684
};
\addplot [thick, color0]
table {%
959907376.309353 2.01081123506452
973277824 2.01169061660767
1041062464 1.99215996265411
1106610048 1.98683536052704
1186761088 1.98662269115448
1262928384 1.99009156227112
1337498496 1.97096228599548
1402526848 1.97625708580017
1476535808 1.97606956958771
1538601088 1.9726699590683
1597438592 1.97433924674988
1653711616 1.97652924060822
1702875648 1.98354196548462
1760286464 1.97862029075623
1809481216 1.9694664478302
1848656640 1.96679449081421
1889161472 1.96645474433899
1922826624 1.96257758140564
1960281600 1.96122419834137
1997938944 1.95644807815552
2029762944 1.95076894760132
2069010048 1.94977748394012
2097710336 1.94823145866394
2121824000 1.94472444057465
2153474560 1.94513869285583
2174902272 1.9434140920639
2199744000 1.93912756443024
2223665408 1.94134473800659
2245219584 1.93988025188446
2261681152 1.93745100498199
};
\addplot [thick, color0, opacity=0.4]
table {%
534425824 2.03301310539246
558068736 2.03363466262817
583957824 2.02631807327271
621326656 2.027428150177
660280640 2.03004050254822
706429696 2.02080368995667
753106432 2.02241277694702
800943232 2.02345848083496
854569152 2.01991081237793
959907376.309353 2.01081123506452
};
\addplot [thick, color0, opacity=0.4]
table {%
2261681152 1.89200709145123
};
\addplot [thick, color0]
table {%
2261681152 1.89200709145123
2302617856 1.88949608802795
2354772736 1.89307284355164
2398968320 1.89230859279633
2441886208 1.89454019069672
2486350848 1.89093518257141
2517301760 1.8890255689621
2552888832 1.88863217830658
2584985088 1.88651216030121
2615736832 1.88308727741241
2643860224 1.87881088256836
2671111168 1.87736010551453
2695002624 1.87511229515076
2714739712 1.87492501735687
2733680640 1.87511420249939
};
\addplot [thick, color0, opacity=0.4]
table {%
640757632 2.04394507408142
672270272 2.03672885894775
709599424 2.02737140655518
750136640 2.02063965797424
801086272 2.01830387115479
848351232 2.0046980381012
912814976 1.9996851682663
985516224 1.98485350608826
1067997248 1.98678052425385
1139235968 1.98000776767731
1217613056 1.95502984523773
1301457920 1.94566988945007
1385014400 1.94203650951385
1470155264 1.92908847332001
1562782336 1.91100800037384
1645307648 1.90614295005798
1724982656 1.91038990020752
1801000192 1.90843963623047
1876115968 1.90688169002533
1947548928 1.90804839134216
2015549184 1.90043354034424
2091250304 1.88420855998993
2160307712 1.89086246490479
2212439296 1.89129066467285
2261681152 1.89200709145123
};
\addplot [thick, black]
table {%
347879326.136183 2.14687061309814
361727626.552966 2.13130164146423
375162379.739126 2.11325216293335
397002573.00128 2.08358311653137
414159506.715542 2.06706571578979
434178158.21416 2.039794921875
457168729.465346 2.01291728019714
485193404.560621 1.9836345911026
508919294.319254 1.95543789863586
538296192.193548 1.92363631725311
572700420.392139 1.89095640182495
606902204.048577 1.85980689525604
640841690.595128 1.83021342754364
701484481.115504 1.78349309453675
};
\addplot [thick, black, opacity=0.4]
table {%
701484481.115504 1.78349309453675
703755366.737516 1.78179430961609
737182980.945264 1.75980854034424
765590797.142231 1.74487030506134
791882848.444775 1.73082113265991
820263168.940833 1.71275746822357
850212309.2 1.69924116134644
876102824.972457 1.69035649299622
901540139.818152 1.68150699138641
923664046.354734 1.67388939857483
946516830.704365 1.66758060455322
964395588.591147 1.66395211219788
981285181.41582 1.66075778007507
998996536.525652 1.65893459320068
1018002227.81026 1.65431249141693
1035067064.6887 1.65237927436829
1051586649.50711 1.65012300014496
1064085167.33591 1.64703500270844
1077417048.13744 1.64393377304077
1089882741.26516 1.64177680015564
1100909610.25237 1.6405873298645
1110445595.84832 1.64020991325378
1120330735.26257 1.63906478881836
1132233545.58206 1.63702011108398
1141316559.32981 1.63653802871704
1149065380.13846 1.63574457168579
};
\addplot [thick, black]
table {%
701484481.115504 1.78349309453675
734521130.69566 1.7741060256958
789286261.732299 1.75461602210999
851181137.837095 1.73496913909912
917087528.193548 1.71381890773773
979892564.978218 1.69795799255371
1042815868.43164 1.69309139251709
1106645282.37716 1.68134236335754
1198865897.12434 1.6691701318964
};
\addplot [thick, black, opacity=0.4]
table {%
530076098.996907 1.85380780696869
552953709.02907 1.84084570407867
577974252.155445 1.83237016201019
613206015.695262 1.81882226467133
650716356.991202 1.80438733100891
701484481.115504 1.78349309453675
};
\addplot [thick, black, opacity=0.4]
table {%
1198865897.12434 1.6691701318964
1241401776.27401 1.66521108150482
1309436882.74441 1.65760707855225
1386299646.30715 1.64530646800995
1454149866.79764 1.64091444015503
1514510576.7379 1.63380312919617
1572376707.6 1.62719202041626
1635273381.11685 1.6221114397049
1691619412.52104 1.62095201015472
1746383356.02245 1.61883723735809
1793302366.22275 1.61740267276764
1844683396.60257 1.61388754844666
1874277935.75281 1.61267721652985
1918197833.66823 1.61174964904785
1957966170.17307 1.61068558692932
1992154109.38711 1.60939836502075
2030950492.04739 1.60841619968414
2064577317.4562 1.60790765285492
2098766602.81409 1.60680294036865
2136810427.70573 1.60371840000153
2161613667.52667 1.60358512401581
2183584679.66508 1.60193920135498
2203023398.27952 1.60038352012634
2222314613.54025 1.5999995470047
2244763427.22507 1.59956324100494
2263886772.5818 1.59869480133057
};
\addplot [thick, black]
table {%
1198865897.12434 1.6691701318964
1208144577.6515 1.66650140285492
1297803529.839 1.64361894130707
1379151434.15876 1.62778413295746
1459289205.4301 1.61661648750305
1540591557.59027 1.60653257369995
1619778035.12518 1.59820115566254
1701705825.84081 1.59231781959534
1778102849.54033 1.58724224567413
1859458594.90168 1.58134162425995
1930174082 1.57608449459076
1996965960.51104 1.57349681854248
2062168905.7792 1.57467126846313
2122966586.78177 1.5746762752533
2180806055.70372 1.57264685630798
2237340776.45978 1.57274305820465
2291210901.13193 1.57327806949615
2333941754.96954 1.57303667068481
2380579218.34147 1.57238733768463
2425551477.83613 1.5715457201004
2462665690.56872 1.57063472270966
2497077257.66524 1.57044208049774
2532060506.83838 1.5705589056015
2573285423.94205 1.57018136978149
2600748017.38058 1.56973123550415
2626554379.24919 1.56975591182709
2652768747.09005 1.56838667392731
2683710462.15941 1.56829941272736
2705761200.11402 1.56799030303955
2725283946.03778 1.56734526157379
};
\addplot [thick, black, opacity=0.4]
table {%
638825850.402111 1.85557150840759
674049073.02727 1.83755743503571
713364238.577661 1.82397770881653
748529850.6863 1.80255663394928
792661100.052786 1.78921294212341
841684837.232335 1.77538549900055
907612872.451938 1.75603044033051
979744392.417071 1.72904765605927
1053295162.00452 1.70903944969177
1198865897.12434 1.6691701318964
};
\addplot [semithick, red, dotted, mark=*, mark size=1, mark options={solid}]
table {%
923732065 1.87231917686462
1383475241 1.76546770420074
1653911497 1.77366101169586
2529598017 1.70817250990868
2851527169 1.77414384460449
};
\addplot [semithick, blue, dashed, mark=square*, mark size=1, mark options={solid}]
table {%
892236033 2.13071036300659
1817832705 1.91269983944893
2743429377 2.00284180908203
3668775169 1.92367886199951
4099178241 1.93158780784607
};
\end{axis}

\end{tikzpicture}
  \end{subfigure}
  \hfill  
  \begin{subfigure}[b]{.24\textwidth}
    \raggedleft
\begin{tikzpicture}

\definecolor{color0}{rgb}{1,0.498039215686275,0.0549019607843137}

\begin{axis}[
tick align=outside,
tick pos=left,
x grid style={white!69.0196078431373!black},
xlabel={FLOPs},
xmajorgrids,
xmin=160314380.392992, xmax=4286743186.74319,
xtick style={color=black},
xtick={0,500000000,1000000000,1500000000,2000000000,2500000000,3000000000,3500000000,4000000000,4500000000},
xticklabels={0.0,0.5,1.0,1.5,2.0,2.5,3.0,3.5,4.0,4.5},
y grid style={white!69.0196078431373!black},
ylabel={\(\displaystyle \leftarrow\) ECE},
ymajorgrids,
ymin=0.0527349494136518, ymax=0.20018813233971,
ytick style={color=black},
ytick={0.04,0.06,0.08,0.1,0.12,0.14,0.16,0.18,0.2,0.22},
yticklabels={0.04,0.06,0.08,0.10,0.12,0.14,0.16,0.18,0.20,0.22}
]
\addplot [thick, color0]
table {%
350043872 0.155193475311126
363169184 0.158600241982158
378871872 0.16009372759105
399106464 0.16530854172443
419256128 0.170970181029047
438078144 0.17532481286993
459118400 0.178908718051623
487951136 0.183879235550087
519883456 0.188459517660752
547457344 0.192589842546825
576095552 0.191897547788656
609939392 0.19348571493398
641712512 0.192256690209834
671926976 0.191590130808365
703289856 0.191005867509986
740006784 0.187999251943138
769968512 0.185467454045262
793556032 0.185635137977313
820766464 0.1836440429376
847834368 0.182229392744189
868118016 0.179946915600168
905592992.789988 0.174993729547522
};
\addplot [thick, color0, opacity=0.4]
table {%
905592992.789988 0.174993729547522
915454592 0.172610925346164
933698560 0.17005572610165
956083264 0.167303583813672
976659200 0.165758986437141
998404096 0.16289488527643
1015859200 0.161358838572574
1030473472 0.158058562626192
1047750784 0.157139934451137
1061431360 0.156198975920078
1073492096 0.154868258459484
1088581632 0.153908041703159
1101310464 0.15378654982371
1111330176 0.153414984272443
1123051520 0.151491393835377
1133661696 0.149734177664686
1143388416 0.148560539994099
1150651776 0.147763653384368
};
\addplot [thick, color0]
table {%
905592992.789988 0.174993729547522
912004352 0.175055215748102
973277824 0.178432733228351
1041062464 0.176756509744466
1106610048 0.175191094827834
1186761088 0.177506322615054
1262928384 0.178981536061837
1337498496 0.177854528895421
1402526848 0.177255749438557
1476535808 0.175206368970881
1538601088 0.173380319778049
1597438592 0.171946306059744
1653711616 0.170890079798845
1702875648 0.169389053618411
1760286464 0.168416325775856
1809481216 0.167012357884524
1848656640 0.166283652441294
1889161472 0.166163394666577
1922826624 0.163821208162743
1960281600 0.162581611802613
1997938944 0.160057076941366
2029762944 0.159796693522765
2069010048 0.157894249555569
2105630587.24861 0.156666446615239
};
\addplot [thick, color0, opacity=0.4]
table {%
534425824 0.130377598657145
558068736 0.134790560445404
583957824 0.140059019783225
621326656 0.146138722231077
660280640 0.152229133950654
706429696 0.156856082867127
753106432 0.164224958103124
800943232 0.167660655168538
905592992.789988 0.174993729547522
};
\addplot [thick, color0, opacity=0.4]
table {%
2105630587.24861 0.156666446615239
2121824000 0.155553678211514
2153474560 0.154534076525936
2174902272 0.153290918374028
2199744000 0.152259467856275
2223665408 0.152090408943182
2245219584 0.150963284582368
2261681152 0.148260672583102
};
\addplot [thick, color0]
table {%
2105630587.24861 0.156666446615239
2160307712 0.156400823228996
2212439296 0.155154406928556
2254626048 0.154507465013737
2302617856 0.151905817330075
2354772736 0.150211273761424
2398968320 0.148512610381056
2441886208 0.147978841779316
2486350848 0.14757660203808
2517301760 0.146014867584038
2552888832 0.144947508619529
2584985088 0.144728211440457
2615736832 0.143765745534774
2643860224 0.141792036808572
2671111168 0.140187262051358
2695002624 0.139724326766189
2714739712 0.138204740270541
2733680640 0.136995987724419
};
\addplot [thick, color0, opacity=0.4]
table {%
640757632 0.125784087267336
672270272 0.130811194323267
709599424 0.137376862611035
750136640 0.143164780218502
801086272 0.1499576306276
848351232 0.155639997512773
912814976 0.162537813545998
985516224 0.167170508282553
1067997248 0.169535916777765
1139235968 0.173029486209729
1217613056 0.1696287916548
1301457920 0.171942635789705
1385014400 0.174734316496747
1470155264 0.174098515571649
1562782336 0.17056190038034
1645307648 0.170346583488286
1724982656 0.168860418907699
1801000192 0.167800888467065
1876115968 0.166420896049481
1947548928 0.166422900567101
2015549184 0.162679915163678
2105630587.24861 0.156666446615239
};
\addplot [thick, black]
table {%
347879326.136183 0.0627697816848755
361727626.552966 0.0624302961349487
375162379.739126 0.0625412159919739
397002573.00128 0.0646107396125793
414159506.715542 0.0660044274330139
434178158.21416 0.0668316246032715
457168729.465346 0.0722052109718323
485193404.560621 0.076294847869873
508919294.319254 0.0807818494796753
538296192.193548 0.0839327396392822
572700420.392139 0.0847259467124939
606902204.048577 0.0899785264015198
640841690.595128 0.0909231706619263
672645355.359634 0.0896499978065491
703755366.737516 0.0944049683570861
737182980.945264 0.0963841767311096
765590797.142231 0.0976305886268616
791882848.444775 0.0973784768104553
820263168.940833 0.0991159188270569
850212309.2 0.10027128534317
876102824.972457 0.0992673256874085
901540139.818152 0.0971867867469788
923664046.354734 0.0962186608314514
946516830.704365 0.0934933800697327
964395588.591147 0.0917438576698303
981285181.41582 0.0907638512611389
998996536.525652 0.0908104348182679
1018002227.81026 0.0902389168739319
1035067064.6887 0.0895361886024475
1051586649.50711 0.0886742226600647
1064085167.33591 0.0883954018592834
1077598060.62801 0.0879469275657289
};
\addplot [thick, black, opacity=0.4]
table {%
1077598060.62801 0.0879469275657289
1089882741.26516 0.0875484322547913
1100909610.25237 0.0866962563514709
1110445595.84832 0.0865510659217835
1120330735.26257 0.0860543160438538
1132233545.58206 0.0853587197303772
1141316559.32981 0.0853135628700256
1149065380.13846 0.0848393715858459
};
\addplot [thick, black]
table {%
1077598060.62801 0.0879469275657289
1106645282.37716 0.087820198726654
1177827707.27107 0.0837903520584106
1241401776.27401 0.0842215388298034
1338535430.23125 0.0783651012621516
};
\addplot [thick, black, opacity=0.4]
table {%
530076098.996907 0.0922437232971191
552953709.02907 0.0914090307235718
577974252.155445 0.0911619410514831
613206015.695262 0.0894221395492554
650716356.991202 0.090229087305069
688591282.012911 0.0903244026184082
734521130.69566 0.0910555548667907
789286261.732299 0.0899306080341339
851181137.837095 0.091703108215332
917087528.193548 0.090032872390747
979892564.978218 0.0884833257675171
1077598060.62801 0.0879469275657289
};
\addplot [thick, black, opacity=0.4]
table {%
1338535430.23125 0.0783651012621516
1386299646.30715 0.0758878161430359
1454149866.79764 0.0765885081291199
1514510576.7379 0.0764079951286316
1572376707.6 0.0769802058219909
1635273381.11685 0.0765419474601746
1691619412.52104 0.0741820414543152
1746383356.02245 0.0734226847648621
1793302366.22275 0.0724638926506043
1844683396.60257 0.0702677149772644
1874277935.75281 0.0702602412223816
1918197833.66823 0.0687374369621277
1957966170.17307 0.0663825053215027
1992154109.38711 0.0653662852287293
2030950492.04739 0.0648372294425965
2064577317.4562 0.0645650370597839
2098766602.81409 0.0643542262077332
2136810427.70573 0.063772538280487
2161613667.52667 0.0634048289299011
2183584679.66508 0.0628319804191589
2203023398.27952 0.0628475962638855
2222314613.54025 0.0631069773674011
2244763427.22507 0.0632972224235535
2263886772.5818 0.0628553416252136
};
\addplot [thick, black]
table {%
1338535430.23125 0.0783651012621516
1379151434.15876 0.0793070693969727
1459289205.4301 0.0829323866844177
1540591557.59027 0.0856651432991028
1619778035.12518 0.0854379765510559
1701705825.84081 0.0870637761116028
1778102849.54033 0.0886550236701965
1859458594.90168 0.0915581013679504
1930174082 0.0949601933479309
1996965960.51104 0.0950552830696106
2062168905.7792 0.0951385533332825
2122966586.78177 0.0978146519660949
2180806055.70372 0.0995126347541809
2237340776.45978 0.0996651967048645
2291210901.13193 0.10016862077713
2333941754.96954 0.0996743977546692
2380579218.34147 0.0993029500007629
2425551477.83613 0.0985471913337707
2462665690.56872 0.0981131398200989
2497077257.66524 0.0979377095222473
2532060506.83838 0.0969354146003723
2573285423.94205 0.0960854756355286
2600748017.38058 0.0952074910163879
2626554379.24919 0.0944473469734192
2652768747.09005 0.0933891896247864
2683710462.15941 0.0924991017341614
2705761200.11402 0.0921939351081848
2725283946.03778 0.0918075520515442
};
\addplot [thick, black, opacity=0.4]
table {%
638825850.402111 0.0614134815692902
674049073.02727 0.0617694751739502
713364238.577661 0.0602754064559936
748529850.6863 0.0594373668193817
792661100.052786 0.0611458664894104
841684837.232335 0.0620032850265503
907612872.451938 0.0630423970222473
979744392.417071 0.0679913192749023
1053295162.00452 0.0717373963356018
1128385031.41935 0.0703071329116821
1208144577.6515 0.074190613079071
1338535430.23125 0.0783651012621516
};
\addplot [semithick, red, dotted, mark=*, mark size=1, mark options={solid}]
table {%
923732065 0.122606126915241
1383475241 0.106072069146673
1653911497 0.116012760530035
2529598017 0.10462279221073
2851527169 0.112869462045565
};
\addplot [semithick, blue, dashed, mark=square*, mark size=1, mark options={solid}]
table {%
892236033 0.131208449524189
1817832705 0.120629054233296
2743429377 0.138969349420266
3668775169 0.132975073156932
4099178241 0.123174675402561
};
\end{axis}

\end{tikzpicture}
  \end{subfigure}\\
  \vspace*{-.9em}
  \caption{Accuracy (Top-1 \& Top-5) and uncertainty metrics (NLPD and ECE) on a budgeted batch classification task as a function of average computational budget per image (FLOPs) on different data sets with a small/medium/large model, and ResNet/DenseNet baselines.\looseness-1} 
\label{fig:result}
\vspace*{-3pt}
\end{figure*}

\smallskip
\noindent
\textbf{CIFAR-100}
experiment results are visualized in \cref{fig:result} and numerical results are presented in \cref{tbl:results}. The three model sizes, small, medium, and large, are plotted as separate curves in \cref{fig:result}. From the results in \cref{fig:result} we see that our model improves Top-1 and Top-5 accuracies over all tested computational budget levels, compared to the vanilla MSDNet model (in-line with results in \cite{huang2018multi}), and improves uncertainty quantification and calibration properties, which is seen in the decrease of negative log-predictive density (NLPD) and expected calibration error (ECE). From the curves, we can pinpoint at $10^8$ FLOPs an improvement of 1.2~\%-points in Top-1 accuracy and 1.1~\%-points in Top-5 accuracy. We also note that although the vanilla MSDNet has clearly superior Top-1 accuracy compared to baseline ResNet and DenseNet models, it has poor performance in terms of NLPD and ECE in comparison to the baselines. CIFAR-100 experiment details are in \cref{sec:app_details_cifar100}.\looseness-1

To investigate the contribution of better decision-making on the improved predictive performance, we performed an experiment separating the improvement due to better decision-making from the improvement due to better predictions at individual intermediate exits. In this experiment, we replaced the vanilla model decision-making with the decision-making of our model, while using the vanilla model predictions for calculating the results. Results on CIFAR-100 are in \cref{fig:cifar100_result_decision_maintext} showing that our approach improves both decision-making (orange vs. light blue) and prediction quality (light blue vs. black). Interestingly, apart from improving accuracy, better decision-making also improves calibration, as seen from the improved ECE (details in \cref{sec:app_results}).\looseness-1

\smallskip
\noindent
\textbf{ImageNet}
experiment results are visualized in \cref{fig:result}, and numerical results are presented in \cref{tbl:results}. \cref{fig:result} shows that on larger computational budgets, our model achieves improvements over the vanilla MSDNet model on all metrics. We note that the ECE numbers achieved by the vanilla MSDNet on ImageNet data are much better than those achieved by the vanilla MSDNet on CIFAR-100 data, suggesting relatively good calibration especially on lower budgets, and resulting in limited usefulness of uncertainty quantification methods. As the computational budget increases, vanilla MSDNet becomes less calibrated, negatively affecting the decision-making at the intermediate classifiers, which is seen in worse accuracy compared to our model. The good calibration of the vanilla MSDNet on ImageNet can be explained by the relatively small model size preventing overfitting the data. The largest MSDNet model used for ImageNet here has $62$ million parameters, while the current state-of-the-art model \cite{chen2023symbolic} has $2440$ million parameters. From the accuracy curves in \cref{fig:result} we can pinpoint at $2.5 \cdot 10^9$ FLOPs an improvement of 0.63~\%-points in Top-1 accuracy and 0.34~\%-points in Top-5 accuracy. ImageNet experiment details are in \cref{sec:app_details_imagenet}.\looseness-1

\smallskip
\noindent
\textbf{Caltech-256}
is an image classification data set with similar resolution images as ImageNet, but with a small number of training samples. For Caltech-256 the same backbone DNN models were used as for ImageNet. Results are visualized in \cref{fig:result} where we see similar trends as for CIFAR-100, with uncertainty quantification techniques improving performance on all metrics. Numerical results are presented in \cref{tbl:results_caltech}, where different uncertainty quantification methods excel at different metrics, but improve consistently over the vanilla MSDNet. We note that the same backbone DNNs that were used for ImageNet, benefit more from our uncertainty quantification methods when used on Caltech-256, possibly due to the smaller training set size that allows overfitting with these model sizes. We can deduce that uncertainty quantification in DNNs is beneficial especially when the standard model is overfitting the data, which is usually the case. From the accuracy curves in \cref{fig:result} we can pinpoint at $2.5 \cdot 10^9$ FLOPs an improvement of 1.4~\%-points in Top-1 accuracy and 1.8~\%-points in Top-5 accuracy. Caltech-256 experiment details are in \cref{sec:app_details_caltech}.\looseness-1

\section{Conclusion and Discussion}
We have demonstrated the importance of uncertainty quantification in dynamic neural networks (DNNs).
For this purpose, we employed a probabilistic treatment of DNNs and leveraged a computationally efficient post-hoc posterior approximation through multiple last-layer Laplace approximations together with model-internal ensembling (MIE).
Our approach substantially improves the internal decision-making process fundamental to DNNs, as evidenced by improved Top-1 and Top-5 accuracy, NLPD, and ECE on CIFAR-100, ImageNet, and Caltech-256.
We found that uncertainty quantification and calibration are especially crucial for large-scale models that overfit training data---stressing their importance in DNNs applied to real-world scenarios.\looseness-1

\smallskip
\noindent
\textbf{Why Laplace and MIE?}
Uncertainty quantification comes in many facets and, if done in DNNs, has to be computationally efficient to be viable for budget-constrained applications.
In this work, we proposed to employ last-layer Laplace approximations at each exit of the DNN.
Although the Laplace approximation is a more crude approximation to the posterior in comparison to techniques such as deep ensembles, it provides a good trade-off between accuracy of the approximation and computational costs, which is essential to DNNs.
In the experiments, we showed that our efficient Laplace approach adds little computational overhead and provides overall NLPD and ECE improvements in most experiments.
In addition, we showed that MIE can further boost performance with little additional costs by informing consecutive blocks in the DNN about the predictive uncertainties of previous exits. As a conclusion, the Laplace approximations account for epistemic uncertainties of each exit \emph{independently} while MIE allows us to incorporate dependence between the exits---hence, complementary to each other. \looseness-1

\smallskip
\noindent
\textbf{Broader Impact.}
This work contributes towards improving the resource and energy efficiency of often prohibitively expensive deep learning models. For example, Microsoft reported \cite{Bing2019} that answering Bing queries using BERT requires 2000 Azure GPU virtual machines to run in parallel. Uncertainty quantification is typically seen as a means of improving robustness and even safety of deep learning models, generally adding to the compute. This work takes the opposite direction leveraging uncertainty to reduce the overall number of floating point operations required for making predictions---while also providing more reliable uncertainty estimates for downstream applications. Improvements in DNNs allow using more powerful models in edge computing and on mobile hardware, and decreases the total energy usage of a heavy deep-learning task on non-constrained hardware.\looseness-1

The codes to replicate the results are available at \url{https://github.com/AaltoML/calibrated-dnn}.

{\small
\bibliographystyle{ieee_fullname}
\bibliography{bibliography}
}

\clearpage
\appendix

\section{Experiment Details}
\label{sec:app_details}
In each experiment, the MSDNet backbone models are trained on the training data to minimise the L2 regularised sum of cross-entropy losses computed for all exits of the model. The L2 regularisation is implicit through weight decay in the SGD optimizer, and the loss for a single input sample is $\mathcal{L}_i = \sum_{k=1}^{n_\text{block}} -\ln p_k(\hat{\vy}_i = \vy_i \mid \vx_i)$, where $p_k(\hat{\vy}_i = \vy_i \mid \vx_i)$ is the predicted softmax confidence on the correct label $\vy_i$ at classifier $k$. Moreover, validation data is used after every epoch to assess the performance of the model. Subsequently, we select the model that achieved the highest Top-1 accuracy on the validation set at the last classifier as the final model.

After training, we evaluate each model using the test set on a budgeted batch classification setup. For this, the computational budget is fixed, and each model must aim to classify the test set samples within the given budget. To achieve this, we use the validation set to calculate thresholds $t_{k}, k = 1, 2, \ldots, n_\text{block}$, one for each exit, such that the overall cost of classifying all test samples does not exceed the predefined budget $B$ in expectation. To utilise the thresholds $t_{k}$, we need to employ a metric of uncertainty assigned to each prediction. If the uncertainty does not exceed the pre-calculated threshold at stage $k$ we exit the model at the current stage and otherwise pass the test sample to the next block and continue with the computation. In our experiments, this uncertainty metric is the prediction confidence: the maximum softmax output value for each sample in the case of a vanilla MSDNet. For our model leveraging Laplace approximations we compute the maximum of the predictive posterior probabilities. As increasing prediction confidence corresponds to decreasing uncertainty, a sample is exited at exit $k$ if the predicted confidence exceeds the confidence threshold $t_{k}$ that has been calculated on the validation set.

A grid search is used to select values for the temperature scaling hyperparameter $T_k$ and Laplace approximation prior variance $\sigma_k$. The values of $T_k$ used in the grid search for all models are the following: [0.3, 0.5, 0.7, 1.0, 1.3, 1.5, 1.7, 2.0, 2.5, 3.0]. The values of Laplace prior variance $\sigma_k$ used in the grid search are [0.5, 0.7, 1.0, 1.3, 1.5, 1.7, 2.0, 2.5, 3.0, 4.0]. If the temperature scale or Laplace prior variance are not optimised for in a grid search, as in many ablation study model options, they are set to their default values of 1.0 and 2.0 respectively. When MIE is not used, the grid search for $T_k$ and $\sigma_k$ is performed independently for each exit $k$ minimising the NLPD on the validation set. When using MIE, $T_k$ and $\sigma_k$ are optimised one exit at a time sequentially, starting from the first exit, minimizing the ensemble prediction NLPD. When performing the grid search for exit $j$, the already optimised values of $T_k$ and $\sigma_k$ for exits $k=1 \ldots j-1$ remain fixed, and the values $T_j$ and $\sigma_j$ are optimised by selecting the pair of values that minimises the MIE prediction NLPD for exit $j$.

To obtain the numbers in \cref{tbl:results,tbl:results_caltech,tbl:results_appendix,tbl:results_appendix_caltech}, results for each model over a range of budgets are averaged. The budget range is different for ImageNet/Caltech-256 and CIFAR-100 models, and for each of the small, medium, and large size models. The computational budget ranges for averaging results for each model are listed in \cref{tbl:budget_ranges}. All experiments are implemented with PyTorch \cite{PyTorch}.

\begin{table}[b!]
  \caption{Ranges of computational budgets (FLOPs) over which results are averaged for different models to obtain the results shown in \cref{tbl:results,tbl:results_caltech,tbl:results_appendix,tbl:results_appendix_caltech}.}
  \vspace{-1em}
  \scriptsize
   \renewcommand{\arraystretch}{.9}
  \setlength{\tabcolsep}{0pt}
  \setlength{\tblw}{0.1\textwidth}  
  \begin{tabularx}{0.49\textwidth}{l @{\extracolsep{\fill}} C{\tblw} C{\tblw} }
  \multicolumn{3}{c}{CIFAR-100} \\
  \toprule
  Model size & Lower budget limit & Upper budget limit \\
  \midrule
  Small & $7.0 \cdot 10^6$ & $2.6 \cdot 10^7$\\
  Medium & $2.5 \cdot 10^7$ & $0.6 \cdot 10^8$\\
  Large & $0.5 \cdot 10^8$ & $1.4 \cdot 10^8$\\
  \bottomrule \\
   
  \multicolumn{3}{c}{ImageNet and Caltech-256} \\
  \toprule  
  Model size & Lower budget limit & Upper budget limit \\
  \midrule
  Small & $3.5 \cdot 10^8$ & $1.1 \cdot 10^9$\\
  Medium & $7.5 \cdot 10^8$ & $2.2 \cdot 10^9$\\
  Large & $1.5 \cdot 10^9$ & $2.6 \cdot 10^9$\\
  \bottomrule
  \end{tabularx}
  \label{tbl:budget_ranges}
\end{table}

\begin{table*}[t!]
  \caption{Additional details on the different MSDNet backbone architectures used for different model sizes on CIFAR-100, ImageNet, and Caltech-256. Each row in the table shows details of a specific block in the architecture. $L$ is the number of layers, FLOPs is the computational cost of processing one input sample, and $n_\text{params}$ is the number of model parameters. FLOPs and $n_\text{params}$ are cumulative numbers \ie they include the numbers from the previous blocks. This means that the cost for the entire architecture is the cost shown for the last block.}
  \vspace{-1em}
  \scriptsize
   \renewcommand{\arraystretch}{.9}
  \setlength{\tabcolsep}{0pt}
  \setlength{\tblw}{0.1\textwidth}  
  \begin{tabularx}{0.99\textwidth}{l @{\extracolsep{\fill}} C{\tblw} C{\tblw}  C{\tblw} | C{\tblw} C{\tblw}  C{\tblw} | C{\tblw}  C{\tblw} C{\tblw}}
  \toprule
  \bf CIFAR-100 & \multicolumn{3}{c}{Small} &  \multicolumn{3}{c}{Medium} &  \multicolumn{3}{c}{Large}  \\
  \midrule
  Block number &$L$ & FLOPs $(10^6)$& $n_\text{params} (10^6)$ &$L$ & FLOPs $(10^6)$& $n_\text{params} (10^6)$ &$L$ & FLOPs $(10^6)$& $n_\text{params} (10^6)$  \\
  \midrule
  1 		& $1$ & $6.86$ & $0.30$ 		& $1$ & $6.86$ & $0.30$		& $1$ & $6.86$ & $0.30$		\\
  2 		& $2$ & $14.35$ & $0.65$ 		& $2$ & $14.35$ & $0.65$		& $2$ & $14.35$ & $0.65$		\\
  3 		& $3$ & $26.13$ & $1.02$ 		& $3$ & $27.29$ & $1.11$		& $3$ & $27.29$ & $1.11$		\\
  4 		& $4$ & $38.04$ & $1.42$ 		& $4$ & $46.56$ & $1.61$		& $4$ & $48.45$ & $1.73$		\\
  5 		& - &- & - 					& $5$ & $67.43$ & $2.11$		& $5$ & $81.57$ & $2.39$		\\
  6 		& - & - & - 					& $6$ & $89.09$ & $2.85$		& $6$ & $112.64$ & $3.18$	\\
  7 		& - & - & - 					& - & - & -					& $7$ & $152.92$ & $4.10$	\\
  8 		& - & - & - 					& - & - & -					& $8$ & $192.69$ & $5.31$	\\
  \\
    \toprule
  \bf ImageNet & \multicolumn{3}{c}{Small} &  \multicolumn{3}{c}{Medium} &  \multicolumn{3}{c}{Large} \\
  \midrule
   Block number &$L$ & FLOPs $(10^6)$& $n_\text{params} (10^6)$ &$L$ & FLOPs $(10^6)$&  $n_\text{params} (10^6)$  &$L$ & FLOPs $(10^6)$& $n_\text{params} (10^6)$ \\
   \midrule
  1 				& $4$ & $339.90$ & $4.24$		& $6$ & $514.66$ & $7.08$		& $7$ & $615.6$ & $8.76$\\
  2 				& $4$ & $685.46$ & $8.77$		& $6$ & $1171.18$ & $15.69$		& $7$ & $1436.39$ & $20.15$\\
  3 				& $4$ & $1008.16$ & $13.07$		& $6$ & $1844.52$ & $24.01$		& $7$ & $2283.21$ & $31.73$\\
  4 				& $4$ & $1254.47$ & $16.75$		& $6$ & $2501.40$ & $42.19$		& $7$ & $2967.42$ & $41.86$\\
  5 				& $4$ & $1360.53$ & $23.96$		& $6$ & $2742.06$ & $56.53$		& $7$ & $3253.79$ & $62.31$\\
  \\
    \toprule
  \bf Caltech-256 & \multicolumn{3}{c}{Small} &  \multicolumn{3}{c}{Medium} &  \multicolumn{3}{c}{Large} \\
  \midrule
   Block number &$L$ & FLOPs $(10^6)$& $n_\text{params} (10^6)$ &$L$ & FLOPs $(10^6)$&  $n_\text{params} (10^6)$  &$L$ & FLOPs $(10^6)$& $n_\text{params} (10^6)$ \\
   \midrule
  1 				& $4$ & $339.62$ & $3.95$		& $6$ & $514.28$ & $6.70$		& $7$ & $615.26$ & $8.33$\\
  2 				& $4$ & $684.88$ & $8.20$		& $6$ & $1170.40$ & $14.90$		& $7$ & $1435.49$ & $19.25$\\
  3 				& $4$ & $1007.32$ & $12.24$		& $6$ & $1843.36$ & $22.86$		& $7$ & $2281.85$ & $30.37$\\
  4 				& $4$ & $1253.41$ & $15.69$		& $6$ & $2499.59$ & $40.38$		& $7$ & $2965.67$ & $40.11$\\
  5 				& $4$ & $1359.05$ & $22.48$		& $6$ & $2739.66$ & $54.13$		& $7$ & $3251.32$ & $59.84$\\
  \bottomrule
  \end{tabularx}
  \label{tbl:arch_details}
\end{table*}

\begin{table}[t!]
  \caption{Details of the baseline ResNet and DenseNet models that are shown for comparison in \cref{fig:result}.  FLOPs is the computational cost of processing one input sample, and $n_\text{params}$ is the number of model parameters. $L_{\text{B}i}$ is the number of layers in block $i$ of the model (a block here refers to all layers between transition layers that change the feature map size). For CIFAR-100 models $i=1 \ldots 3$ and for ImageNet models $i=1 \ldots 4$.}
  \vspace{-1em}
  \scriptsize
   \renewcommand{\arraystretch}{.9}
  \setlength{\tabcolsep}{0pt}
  \setlength{\tblw}{0.1\textwidth}  
  \begin{tabularx}{0.49\textwidth}{l @{\extracolsep{\fill}} C{\tblw} C{\tblw} C{0.5\tblw} C{0.5\tblw} C{0.5\tblw} C{0.5\tblw} }
  \multicolumn{7}{c}{CIFAR-100 baseline models} \\
  \toprule
  Model name  & FLOPs & $n_\text{params}$ & $L_\text{B1}$ & $L_\text{B2}$ & $L_\text{B3}$ &  \\
  \midrule
  ResNet8 & $12.6 \cdot 10^6$ & $8.39 \cdot 10^4$  &1  &1  &1  & \\
  ResNet14 & $26.8 \cdot 10^6$ & $18.1 \cdot 10^4$ &2  &2  &2  & \\
  ResNet20 & $41.0 \cdot 10^6$ & $27.8 \cdot 10^4$ &3  &3  &3  & \\
  ResNet26 & $55.2 \cdot 10^6$ & $37.6 \cdot 10^4$  &4  &4  &4  & \\
  ResNet38 & $83.7 \cdot 10^6$ & $57.0 \cdot 10^4$ &6  &6  &6  & \\
  ResNet50 & $112 \cdot 10^6$ & $76.4 \cdot 10^4$ &8  &8  & 8 & \\
  ResNet62 & $141 \cdot 10^6$ & $95.9 \cdot 10^4$  &10  &10  &10  & \\
  ResNet86 & $197 \cdot 10^6$ & $135 \cdot 10^4$ &14  &14  &14  & \\
  ResNet110 & $254 \cdot 10^6$ & $174 \cdot 10^4$ &18  &18  &18  & \\
  DenseNet10 & $10.0 \cdot 10^6$ & $2.34 \cdot 10^4$ &1  &1  &1  & \\
  DenseNet16 & $20.3 \cdot 10^6$ & $4.88 \cdot 10^4$ &2  &2  &2  & \\
  DenseNet22 & $31.7 \cdot 10^6$ & $7.80 \cdot 10^4$ &3  &3  &3  & \\
  DenseNet28 & $44.4 \cdot 10^6$ & $11.1 \cdot 10^4$ &4  &4  &4  & \\
  DenseNet40 & $73.4 \cdot 10^6$ & $18.8 \cdot 10^4$ &6  &6  &6  & \\
  DenseNet52 & $107 \cdot 10^6$ & $28.0 \cdot 10^4$ &8  &8  &8  & \\
  DenseNet64 & $146 \cdot 10^6$ & $38.8 \cdot 10^4$ &10  &10  &10  & \\
  DenseNet76 & $190 \cdot 10^6$ & $51.0 \cdot 10^4$ &12  &12  &12  & \\
  DenseNet88 & $239 \cdot 10^6$ & $64.8 \cdot 10^4$ &14  &14  & 14 & \\
  \bottomrule \\
   
  \multicolumn{7}{c}{ImageNet baseline models} \\
  \toprule  
  Model name  & FLOPs & $n_\text{params}$  & $L_\text{B1}$ & $L_\text{B2}$ & $L_\text{B3}$ & $L_\text{B4}$\\
  \midrule
  ResNet10 & $8.93 \cdot 10^8$ & $5.42 \cdot 10^6$  &1  &1  &1  &1 \\
  ResNet18 & $18.2 \cdot 10^8$ & $11.7 \cdot 10^6$ &2  &2  &2  &2 \\
  ResNet26 & $27.4 \cdot 10^8$ & $18.0 \cdot 10^6$ &3  &3  &3  &3 \\
  ResNet34 & $36.7 \cdot 10^8$ & $21.8 \cdot 10^6$ &3  &4  &6  &3 \\
  ResNet50 & $41.0 \cdot 10^8$ & $25.6 \cdot 10^6$ &3  &4  &6  &3 \\
  DenseNet57 & $9.24 \cdot 10^8$ & $2.44 \cdot 10^6$ &2  &6  &10  &8 \\
  DenseNet65 & $13.8 \cdot 10^8$ & $2.93 \cdot 10^6$ &4  &6  &12  &8 \\
  DenseNet81 & $16.5 \cdot 10^8$ & $4.18 \cdot 10^6$ &4  &8  &16  & 10 \\
  DenseNet97 & $25.3 \cdot 10^8$ & $5.44 \cdot 10^6$ &6  &12  &16  &12 \\
  DenseNet121 & $28.5 \cdot 10^8$ & $7.98 \cdot 10^6$ &6  &12  &24  &16 \\
  \bottomrule \\
  
    \multicolumn{7}{c}{Caltech-256 baseline models} \\
  \toprule  
  Model name  & FLOPs & $n_\text{params}$  & $L_\text{B1}$ & $L_\text{B2}$ & $L_\text{B3}$ & $L_\text{B4}$\\
  \midrule
  ResNet10 & $8.92 \cdot 10^8$ & $5.04 \cdot 10^6$  &1  &1  &1  &1 \\
  ResNet18 & $18.2 \cdot 10^8$ & $11.3 \cdot 10^6$ &2  &2  &2  &2 \\
  ResNet26 & $27.4 \cdot 10^8$ & $17.6 \cdot 10^6$ &3  &3  &3  &3 \\
  ResNet34 & $36.7 \cdot 10^8$ & $21.4 \cdot 10^6$ &3  &4  &6  &3 \\
  ResNet50 & $41.0 \cdot 10^8$ & $24.0 \cdot 10^6$ &3  &4  &6  &3 \\
  DenseNet57 & $9.24 \cdot 10^8$ & $2.08 \cdot 10^6$ &2  &6  &10  &8 \\
  DenseNet65 & $13.8 \cdot 10^8$ & $2.55 \cdot 10^6$ &4  &6  &12  &8 \\
  DenseNet81 & $16.5 \cdot 10^8$ & $3.69 \cdot 10^6$ &4  &8  &16  & 10 \\
  DenseNet97 & $25.3 \cdot 10^8$ & $4.87 \cdot 10^6$ &6  &12  &16  &12 \\
  DenseNet121 & $28.5 \cdot 10^8$ & $7.22 \cdot 10^6$ &6  &12  &24  &16 \\
  \bottomrule
  \end{tabularx}
  \label{tbl:baseline_models}
\end{table}

\subsection{CIFAR-100 Model and Training Details}
\label{sec:app_details_cifar100}
The CIFAR-100 training set of 50,000 images is split into 45,000 training images and 5,000 validation images. The test set has 10,000 images. On CIFAR-100, models are trained for 300 epochs using a batch size of 64 images, and the learning rate is initially set to 0.1 and is decayed to one-tenth at epochs 150 and 225. The optimiser is SGD with a momentum of 0.9 and weight decay of $10^{-4}$. The MSDNet backbone of all CIFAR-100 models has three scales of features. The number of channels after the first layer is 16, and the number of channels added by each layer is 6.

On CIFAR-100 the `small', `medium', and `large' DNN backbones have 4, 6, and 8 blocks and classifiers each, respectively. The small architecture has a total of 10 layers, and transition layers that reduce the number of scales by one are performed at layers 5 and 9. The medium architecture has a total of 21 layers, and transition layers are performed at layers 8 and 15. The large architecture has a total of 36 layers, with transition layers performed at layers 13 and 25. \cref{tbl:arch_details} shows more detailed information separately for each block of the small, medium, and large models.

\subsection{ImageNet Model and Training Details}
\label{sec:app_details_imagenet}
 The ImageNet training set of 1,281,167 images is split into 1,231,167 training images and 50,000 validation images. The test set is the standard test set of 50,000 images. In preprocessing, the sample images that are of varying resolutions are resized to a resolution of 224 by 224 pixels.
On ImageNet, models are trained for 90 epochs using a batch size of 256, and the learning rate is initially set to 0.1 and it is decayed to one-tenth at epochs 30 and 60. The optimiser is SGD with a momentum of 0.9 and weight decay of $10^{-4}$. The MSDNet backbone of all ImageNet models has four scales of features and five blocks. The number of channels after the first layer is 32, and the number of channels added by each layer is 16.

The small, medium, and large models for ImageNet have five blocks and five classifiers each, but vary in the number of layers in each block (model architecture design follows \cite{huang2018multi}). The small architecture has a total of 20 layers, and transition layers that reduce the number of scales by one are performed at layers 6, 11, and 16. The medium architecture has a total of 30 layers, and transition layers are performed at layers 9, 17, and 25. The large architecture has a total of 35 layers, with transition layers at layers 10, 19, and 28. \cref{tbl:arch_details} shows more detailed information separately for each block of the small, medium, and large models.

\subsection{Caltech-256 Model and Training Details}
\label{sec:app_details_caltech}
The Caltech-256 data set of 30,607 images is split to a training set of 23,107, a validation set of 2,500, and a test set of 5,000 samples. In preprocessing, the sample images that are of varying resolutions are resized to a resolution of 224 by 224 pixels, the same as for ImageNet data. The Caltech-256 data set has images from 257 categories.

On Caltech-256 the MSDNet backbone models are the same as for ImageNet, with the difference of the output dimensionality being 257 instead of 1,000, which affects the number of parameters and FLOPs slightly. Caltech-256 models are trained for 180 epochs with a batch size of 128. This difference in training compared to ImageNet models is due to the smaller number of training samples and the smaller number of classes. The learning rate is initially set to 0.1 and is decayed to one-tenth at epochs 90 and 135. Apart from the mentioned differences, the Caltech-256 models are trained the same as the ImageNet models.

\subsection{Details on Baseline Models and Their Training}
\label{sec:baselines}
We use ResNet and DenseNet models as baseline architectures for CIFAR-100, ImageNet, and Caltech-256 experiments. \cref{tbl:baseline_models} shows the model architecture details for all the used baseline models. Both ResNet and DenseNet models are implemented using the implementations from Torchvision \cite{torchvision}. All ResNet models are built using the basic residual layer with two consecutive 3 by 3 convolutions, except for the ImageNet/Caltech-256 ResNet50, which uses the bottleneck residual layer. For ImageNet and Caltech-256 DenseNet models, the growth rate is 32 and the number of initial features is 64. For CIFAR-100 DenseNet models, the growth rate is 12 and the number of initial features is 24. For ImageNet and Caltech-256, the Torchvision implementations are used as they are. On Caltech-256 the baseline models are the same as for ImageNet, with the difference of the output dimensionality being $257$, which affects the number of parameters and FLOPs slightly.

For CIFAR-100, the architectures need some modifications due to the smaller input dimensionality. For CIFAR-100 ResNets, the first convolutional layer is replaced by a 3 by 3 convolution with stride 1 and 16 output channels, and the max pooling layers are removed. For CIFAR-100 DenseNets, the first convolutional layer is replaced by a 3 by 3 convolution with stride 1, and the first batch normalization, max pooling, and ReLU operations are removed.

On CIFAR-100, ResNet and DenseNet models are trained for 300 epochs, and the learning rate is initially set to 0.1 and is decayed to one-tenth at epochs 150 and 225. On ImageNet, the ResNet and DenseNet models are trained for 90 epochs, and the learning rate is initially set to 0.1 and it is decayed to one-tenth at epochs 30 and 60. On Caltech-256 the baseline models are trained for 180 epochs with a batch size of 128. The learning rate is initially set to 0.1 and is decayed to one-tenth at epochs 90 and 135. For all data sets all ResNet and DenseNet models use the SGD optimiser with a momentum of 0.9 and weight decay of $10^{-4}$.

\subsection{Details on the Used Performance Metrics}
\label{sec:app_metrics}
For reporting model performances, we use the Top-1 and Top-5 accuracies, the negative log-predictive density (NLPD), and the expected calibration error (ECE). Top-1 accuracy is the standard accuracy metric, and is the percentage of test predictions for which the highest model predicted probability was on the correct class. Top-5 accuracy is the percentage of test predictions, for which the correct class is among the five classes that the model assigned the highest probability. Negative log-predictive density (NLPD) is defined as:
\begin{equation}\textstyle
  \nlpd = - \sum^{n_{\text{test}}}_{i=1} \log p(\hat\vy_{i} = \vy_{i}\mid \vx_{i}),
\end{equation}
where $p(\hat\vy_{i} = \vy_{i}\mid \vx_{i})$ is the model predicted probability on the correct label $\vy_{i}$. NLPD is a metric that captures both the quality of uncertainty estimates as well as the correctness of the predictions, most heavily penalizing overconfident incorrect predictions.

Expected calibration error (ECE) is defined as:
\begin{equation}\textstyle
  \ece = \sum^{m}_{j=1} b_{j} \| (p_{j} - \mu_{j}) \|,
\end{equation}
where $b_{j}$ is the fraction of test samples in bin $j=1,\ldots,m$, $p_{j}$ is the Top-1 accuracy of the $j$\textsuperscript{th} bin, and $\mu_{j}$ is the average confidence of the predictions in the bin. We use $m=10$ in our experiments. ECE assesses the calibration of each model, \ie, how consistent the confidence scores are with the posterior probabilities.

\section{Additional Results}
\label{sec:app_results}

\begin{figure*}
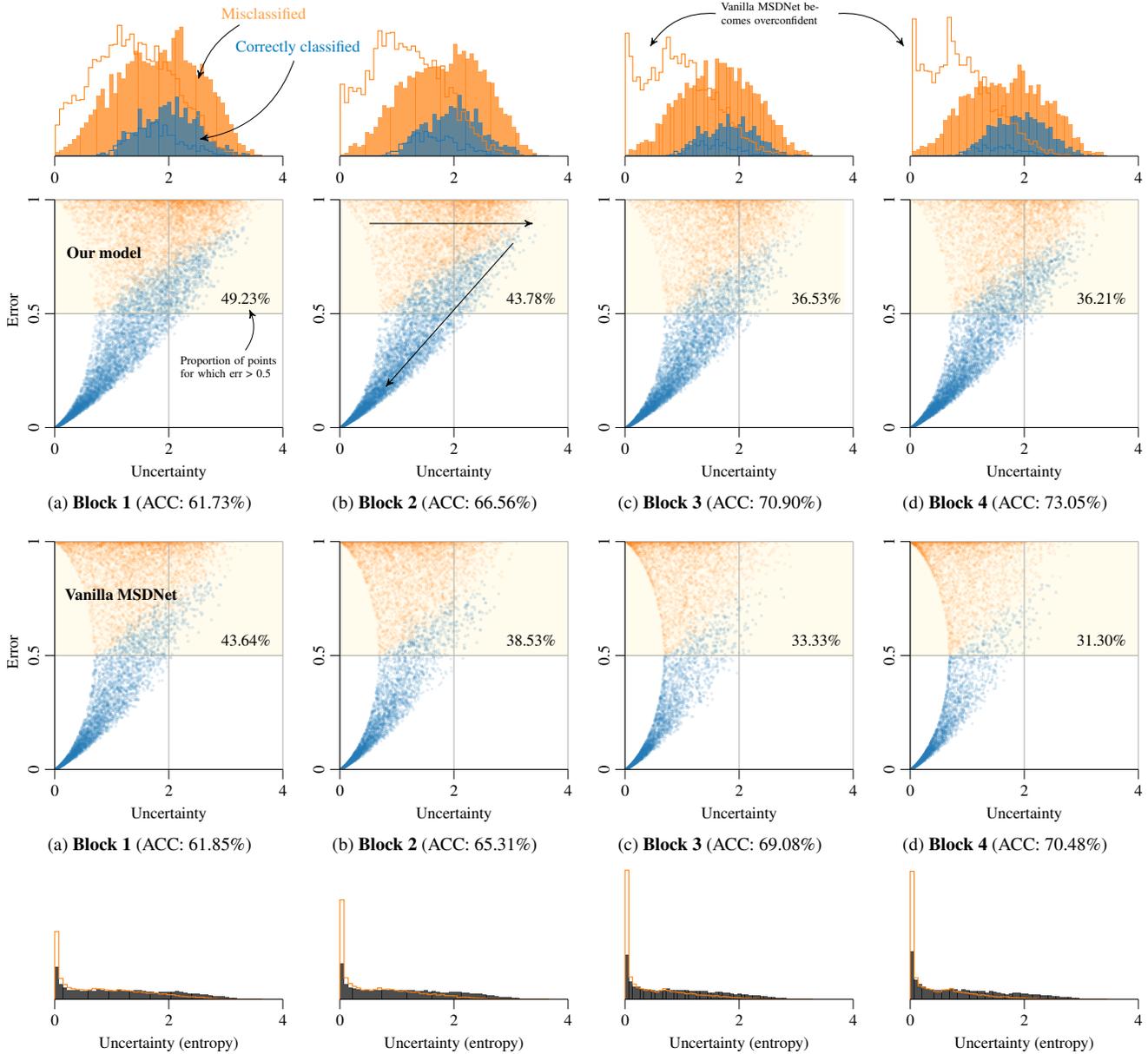

  \centering\scriptsize
  \pgfplotsset{axis on top,scale only axis,width=\figurewidth,height=\figureheight,
    xtick align=outside,
    ytick align=outside,  
    axis x line*= bottom, axis y line*= left,
    y tick label style={rotate=90}}
  \begin{tikzpicture}

    \setlength{\figurewidth}{0.20\textwidth}
    \setlength{\figureheight}{0.7\figurewidth}  

    \begingroup
    
    \pgfplotsset{y axis line style={draw=none,draw opacity=0}}

    \foreach \x [count=\i] in {0,1,2,3} 
      \node[anchor=north] at (\i*0.25*\textwidth,0) {\begin{minipage}{.25\textwidth}\raggedleft{\input{fig/upper_hist_block\x.tex}}\end{minipage}};

    \endgroup

    \setlength{\figurewidth}{0.20\textwidth}
    \setlength{\figureheight}{\figurewidth}  
    \pgfplotsset{xlabel={Uncertainty}}

    \foreach \x [count=\i] in {0,1,2,3}  
      \node[anchor=north] (scatter-\i) at (\i*0.25*\textwidth,-0.17*\textwidth) {\begin{minipage}{.25\textwidth}\raggedleft{\input{fig/errunc_scatter_block\x.tex}}\end{minipage}};

    \foreach \a/\x/\y [count=\i] in {a/49.23/61.73,b/43.78/66.56,c/36.53/70.90,d/36.21/73.05} { 
      \node at (\i*0.25*\textwidth,-0.45*\textwidth) {\footnotesize (\a) \textbf{Block~\i}\ (ACC: \y\%)};     
      \node (perr-\i) at (\i*0.25*\textwidth+6em,-0.27*\textwidth) {\x\%};            
    }
    
    \node[rotate=90] at (0.13*\textwidth,-0.28*\textwidth) {Error};
    \node[rotate=0] at (0.21*\textwidth,-0.23*\textwidth) {\textbf{Our model}};
    
    \foreach \x [count=\i] in {0,1,2,3}  
      \node[anchor=north] (scatterv-\i) at (\i*0.25*\textwidth,-0.47*\textwidth) {\begin{minipage}{.25\textwidth}\raggedleft{\input{fig/vanilla_errunc_scatter_block\x.tex}}\end{minipage}};
      
    \foreach \a/\x/\y [count=\i] in {a/43.64/61.85,b/38.53/65.31,c/33.33/69.08,d/31.30/70.48} { 
      \node at (\i*0.25*\textwidth,-0.75*\textwidth) {\footnotesize (\a) \textbf{Block~\i}\ (ACC: \y\%)};     
      \node (perrv-\i) at (\i*0.25*\textwidth+6em,-0.57*\textwidth) {\x\%};            
    }
    
    \node[rotate=90] at (0.13*\textwidth,-0.58*\textwidth) {Error};
    \node[rotate=0] at (0.225*\textwidth,-0.53*\textwidth) {\textbf{Vanilla MSDNet}};

    \node[color=C0] (a) at (0.35*\textwidth,-0.02*\textwidth) {Misclassified};
    \node[color=C1] (b) at (0.38*\textwidth,-0.05*\textwidth) {Correctly classified};
    \draw[->] (a) to[bend right=20] ++(-1cm,-1cm);
    \draw[->] (b) to[bend left=30] ++(-1.5cm,-1.4cm);
    \node[text width=1.5cm,font=\tiny] (c) at (0.32*\textwidth,-0.33*\textwidth) {Proportion of points for which err > 0.5};
    \draw[->] (c) to[bend right=30] (perr-1);
    \node[text width=1.7cm,font=\tiny] (d) at (0.8*\textwidth,-0.02*\textwidth) {Vanilla MSDNet {becomes} overconfident};
    \draw[->] (d.east) to[bend left=30] ++(1cm,-.5cm);
    \draw[->] (d.west) to[bend right=30] ++(-1cm,-.5cm);
        
    \draw[->,draw=black] ($(scatter-2) + (-1cm,1.75cm)$) -- ($(scatter-2) + (1.5cm,1.75cm)$);
    \draw[->,draw=black] ($(scatter-2) + (1.2cm,1.45cm)$) -- ($(scatter-2) + (-.75cm,-.75cm)$);
  
    \setlength{\figurewidth}{0.20\textwidth}
    \setlength{\figureheight}{0.6\figurewidth}  

    \begingroup
    
    \pgfplotsset{y axis line style={draw=none,draw opacity=0}}

    \foreach \x [count=\i] in {0,1,2,3} 
      \node[anchor=north] at (\i*0.25*\textwidth,-0.76*\textwidth) {\begin{minipage}{.25\textwidth}\raggedleft{\input{fig/full_hist_block\x.tex}}\end{minipage}};
    \endgroup
  \end{tikzpicture}
\vspace*{-2.8em}
     \caption{The second row scatter plots show \includegraphics[width=.7em]{fig/correct}~correctly classified and \includegraphics[width=.7em]{fig/incorrect}~misclassified test points for our model on an uncertainty vs.\ error axis (repeated from \cref{fig:uncertainty_error_app}) and the third row shows the corresponding scatter plots for the vanilla MSDNet model. The uncertainty of the points in the top half of each scatter plot is summarized as a histogram in the top row (repeated from \cref{fig:uncertainty_error_app}). Our model result is shown in the solid histograms, while the vanilla MSDNet results are shown as a histogram outline. Uncertainty histograms of all points in the uncertainty--error scatter plots are shown in the bottom row, comparing our model (black histogram) to the vanilla MSDNet model (orange outline). We can see that the vanilla MSDNet has overall much less uncertainty in its predictions}
\label{fig:uncertainty_error_app}
\end{figure*}
\begin{table*}[t]
  \caption{Table of Top-1/Top-5 accuracy, negative log-predictive density (NLPD), and expected calibration error (ECE) for different models on CIFAR-100 and ImageNet data. All numbers are averages over a range of computational budgets in the budgeted batch classification setup. `MIE Laplace $T_\textrm{opt}$ $\sigma_\textrm{opt}$'-model corresponds to `Our model' that is referred to in other figures. $T_\textrm{opt}$ and $\sigma_\textrm{opt}$ refer to grid search optimisation of the temperature scale and Laplace prior variance, respectively. $n_\text{train}$ is the number of training samples, $d$ is the input dimensionality, $c$ is the number of classes, and $n_\text{batch}$ is the batch size. The red and green numbers show the decrease or increase in performance compared to MSDNet (vanilla). The best performing model for each metric and each model size, on each dataset, is shown in bold.\looseness-1}
  \vspace*{-1em}
  \scriptsize
  \renewcommand{\arraystretch}{.9}
  \newcommand{\fooo}[1]{\textcolor{mycolor2}{\tiny #1}}
   \newcommand{\foog}[1]{\textcolor{mycolor1}{\tiny #1}}
   \newcommand{\foe}[0]{\textcolor{mycolor1}{\tiny $\phantom{+0.0}$}}
      \newcommand{\fooe}[0]{\textcolor{mycolor1}{\tiny $\phantom{+0.00}$}}
      \newcommand{\fooee}[0]{\textcolor{mycolor1}{\tiny $\phantom{+0.000}$}}
  \setlength{\tabcolsep}{0pt}
  \setlength{\tblw}{0.1\textwidth}  
  \begin{tabularx}{\textwidth}{l l @{\extracolsep{\fill}} C{\tblw}  C{\tblw} C{\tblw} C{\tblw} | C{\tblw} C{\tblw} C{\tblw}  C{\tblw} }
  \toprule

& & \multicolumn{4}{c}{\sc CIFAR-100} & \multicolumn{4}{c}{\sc ImageNet} \\
& ($n_\textrm{train}$, $d$, $c$, $n_\textrm{batch}$) & \multicolumn{4}{c}{(50000, 3072, 100, 64)} & \multicolumn{4}{c}{(1281167, 150528, 1000, 256)} \\
\midrule
& & Top-1 ACC $\uparrow$ & Top-5 ACC $\uparrow$ & NLPD $\downarrow$ & ECE $\downarrow$ & Top-1 ACC $\uparrow$ & Top-5 ACC $\uparrow$ & NLPD $\downarrow$ & ECE $\downarrow$ \\
\midrule
\parbox[t]{7mm}{\multirow{12}{*}{\rotatebox[origin=c]{90}{\bf Small}}}
& MSDNet (vanilla) & $ 69.25$~\fooe{} & $ 90.48$~\fooe{} & $ 1.498$~\fooee{} & $ 0.182$~\fooee{} & $ 68.15$~\fooe{} & $ 88.22$~\fooe{} & $ 1.338$~\fooee{} & $ 0.019$~\fooee{} \\
& ~~~+ $T_\textrm{opt}$ & $ 69.06$~\fooo{$-0.19$} & $ 90.62$~\foog{$+0.14$} & $ 1.207$~\foog{$-0.291$} & $ 0.079$~\foog{$-0.103$} & $ 68.15$~\foog{$+0.00$} & $\bf 88.22$~\foog{$+0.00$} & $ 1.338$~\foog{$-0.000$} & $ 0.019$~\foog{$-0.000$} \\
& ~~~+ Laplace & $ 69.21$~\fooo{$-0.04$} & $ 90.46$~\fooo{$-0.02$} & $ 1.419$~\foog{$-0.079$} & $ 0.155$~\foog{$-0.027$} & $ 68.14$~\fooo{$-0.01$} & $ 88.21$~\fooo{$-0.01$} & $\bf 1.335$~\foog{$-0.003$} & $ 0.016$~\foog{$-0.003$} \\
& ~~~+ Laplace $T_\textrm{opt}$ & $ 69.02$~\fooo{$-0.22$} & $ 90.65$~\foog{$+0.17$} & $ 1.196$~\foog{$-0.302$} & $ 0.060$~\foog{$-0.121$} & $ 68.13$~\fooo{$-0.01$} & $ 88.18$~\fooo{$-0.04$} & $ 1.337$~\foog{$-0.001$} & $ 0.016$~\foog{$-0.004$} \\
& ~~~+ Laplace $\sigma_\textrm{opt}$ & $ 69.21$~\fooo{$-0.04$} & $ 90.42$~\fooo{$-0.06$} & $ 1.415$~\foog{$-0.082$} & $ 0.154$~\foog{$-0.028$} & $ 68.13$~\fooo{$-0.02$} & $ 88.17$~\fooo{$-0.05$} & $ 1.337$~\foog{$-0.001$} & $ 0.016$~\foog{$-0.003$} \\
& ~~~+ Laplace $T_\textrm{opt}$ $\sigma_\textrm{opt}$ & $ 69.06$~\fooo{$-0.19$} & $ 90.58$~\foog{$+0.10$} & $ 1.208$~\foog{$-0.289$} & $ 0.073$~\foog{$-0.109$} & $ 68.10$~\fooo{$-0.05$} & $ 88.18$~\fooo{$-0.04$} & $ 1.337$~\foog{$-0.001$} & $\bf 0.015$~\foog{$-0.005$} \\
& ~~~+ MIE & $ 69.97$~\foog{$+0.72$} & $ 90.88$~\foog{$+0.40$} & $ 1.218$~\foog{$-0.280$} & $ 0.080$~\foog{$-0.102$} & $ 68.27$~\foog{$+0.12$} & $ 88.13$~\fooo{$-0.10$} & $ 1.355$~\fooo{$+0.017$} & $ 0.055$~\fooo{$+0.036$} \\
& ~~~+ MIE $T_\textrm{opt}$ & $ 69.74$~\foog{$+0.50$} & $\bf 91.11$~\foog{$+0.63$} & $\bf 1.133$~\foog{$-0.365$} & $ 0.028$~\foog{$-0.154$} & $ 68.25$~\foog{$+0.10$} & $ 88.04$~\fooo{$-0.18$} & $ 1.353$~\fooo{$+0.015$} & $ 0.038$~\fooo{$+0.019$} \\
& ~~~+ MIE Laplace & $\bf 69.99$~\foog{$+0.74$} & $ 90.88$~\foog{$+0.40$} & $ 1.189$~\foog{$-0.308$} & $ 0.056$~\foog{$-0.125$} & $ 68.26$~\foog{$+0.11$} & $ 88.11$~\fooo{$-0.11$} & $ 1.361$~\fooo{$+0.023$} & $ 0.070$~\fooo{$+0.051$} \\
& ~~~+ MIE Laplace $T_\textrm{opt}$ & $ 69.83$~\foog{$+0.58$} & $ 91.10$~\foog{$+0.62$} & $ 1.135$~\foog{$-0.363$} & $ 0.021$~\foog{$-0.161$} & $ 68.22$~\foog{$+0.07$} & $ 88.06$~\fooo{$-0.16$} & $ 1.357$~\fooo{$+0.019$} & $ 0.055$~\fooo{$+0.036$} \\
& ~~~+ MIE Laplace $\sigma_\textrm{opt}$ & $ 69.89$~\foog{$+0.64$} & $ 90.94$~\foog{$+0.46$} & $ 1.192$~\foog{$-0.306$} & $ 0.059$~\foog{$-0.122$} & $ 68.25$~\foog{$+0.10$} & $ 88.14$~\fooo{$-0.08$} & $ 1.360$~\fooo{$+0.022$} & $ 0.070$~\fooo{$+0.051$} \\
& ~~~+ MIE Laplace $T_\textrm{opt}$ $\sigma_\textrm{opt}$ & $ 69.84$~\foog{$+0.59$} & $ 91.09$~\foog{$+0.61$} & $ 1.133$~\foog{$-0.364$} & $\bf 0.017$~\foog{$-0.165$} & $\bf 68.31$~\foog{$+0.16$} & $ 88.11$~\fooo{$-0.11$} & $ 1.356$~\fooo{$+0.018$} & $ 0.052$~\fooo{$+0.032$} \\
\midrule
\parbox[t]{7mm}{\multirow{12}{*}{\rotatebox[origin=c]{90}{\bf Medium}}}
& MSDNet (vanilla) & $ 74.12$~\fooe{} & $ 91.94$~\fooe{} & $ 1.549$~\fooee{} & $ 0.190$~\fooee{} & $ 72.78$~\fooe{} & $ 91.01$~\fooe{} & $ 1.123$~\fooee{} & $ 0.033$~\fooee{} \\
& ~~~+ $T_\textrm{opt}$ & $ 73.96$~\fooo{$-0.17$} & $ 92.05$~\foog{$+0.10$} & $ 1.063$~\foog{$-0.486$} & $ 0.078$~\foog{$-0.112$} & $ 72.78$~\fooo{$-0.00$} & $ 91.01$~\foog{$+0.00$} & $ 1.123$~\fooo{$+0.000$} & $ 0.033$~\fooo{$+0.000$} \\
& ~~~+ Laplace & $ 73.96$~\fooo{$-0.16$} & $ 91.94$~\fooo{$-0.00$} & $ 1.436$~\foog{$-0.113$} & $ 0.172$~\foog{$-0.018$} & $ 72.69$~\fooo{$-0.09$} & $ 91.04$~\foog{$+0.03$} & $\bf 1.117$~\foog{$-0.006$} & $\bf 0.012$~\foog{$-0.021$} \\
& ~~~+ Laplace $T_\textrm{opt}$ & $ 73.98$~\fooo{$-0.15$} & $ 92.01$~\foog{$+0.07$} & $ 1.056$~\foog{$-0.493$} & $ 0.070$~\foog{$-0.120$} & $ 72.68$~\fooo{$-0.10$} & $ 90.98$~\fooo{$-0.03$} & $ 1.117$~\foog{$-0.005$} & $ 0.013$~\foog{$-0.020$} \\
& ~~~+ Laplace $\sigma_\textrm{opt}$ & $ 74.18$~\foog{$+0.05$} & $ 91.85$~\fooo{$-0.09$} & $ 1.405$~\foog{$-0.144$} & $ 0.164$~\foog{$-0.026$} & $ 72.70$~\fooo{$-0.08$} & $ 91.00$~\fooo{$-0.01$} & $ 1.117$~\foog{$-0.005$} & $ 0.013$~\foog{$-0.020$} \\
& ~~~+ Laplace $T_\textrm{opt}$ $\sigma_\textrm{opt}$ & $ 73.92$~\fooo{$-0.20$} & $ 92.01$~\foog{$+0.06$} & $ 1.070$~\foog{$-0.479$} & $ 0.083$~\foog{$-0.107$} & $ 72.72$~\fooo{$-0.07$} & $ 91.03$~\foog{$+0.03$} & $ 1.118$~\foog{$-0.005$} & $ 0.018$~\foog{$-0.015$} \\
& ~~~+ MIE & $ 75.03$~\foog{$+0.91$} & $ 92.97$~\foog{$+1.03$} & $ 1.011$~\foog{$-0.538$} & $ 0.050$~\foog{$-0.140$} & $ 72.98$~\foog{$+0.20$} & $ 91.12$~\foog{$+0.11$} & $ 1.119$~\foog{$-0.004$} & $ 0.042$~\fooo{$+0.009$} \\
& ~~~+ MIE $T_\textrm{opt}$ & $ 74.94$~\foog{$+0.82$} & $ 93.23$~\foog{$+1.29$} & $\bf 0.941$~\foog{$-0.608$} & $ 0.028$~\foog{$-0.162$} & $ 72.99$~\foog{$+0.21$} & $ 91.09$~\foog{$+0.08$} & $ 1.119$~\foog{$-0.004$} & $ 0.038$~\fooo{$+0.005$} \\
& ~~~+ MIE Laplace & $ 74.99$~\foog{$+0.86$} & $ 93.01$~\foog{$+1.07$} & $ 0.990$~\foog{$-0.559$} & $ 0.032$~\foog{$-0.158$} & $ 72.95$~\foog{$+0.17$} & $\bf 91.15$~\foog{$+0.14$} & $ 1.128$~\fooo{$+0.005$} & $ 0.065$~\fooo{$+0.032$} \\
& ~~~+ MIE Laplace $T_\textrm{opt}$ & $ 74.96$~\foog{$+0.84$} & $ 93.19$~\foog{$+1.24$} & $ 0.947$~\foog{$-0.602$} & $\bf 0.015$~\foog{$-0.175$} & $ 72.88$~\foog{$+0.10$} & $ 91.06$~\foog{$+0.06$} & $ 1.124$~\fooo{$+0.001$} & $ 0.045$~\fooo{$+0.012$} \\
& ~~~+ MIE Laplace $\sigma_\textrm{opt}$ & $\bf 75.04$~\foog{$+0.92$} & $ 92.95$~\foog{$+1.01$} & $ 0.989$~\foog{$-0.560$} & $ 0.031$~\foog{$-0.159$} & $ 72.97$~\foog{$+0.19$} & $ 91.12$~\foog{$+0.11$} & $ 1.126$~\fooo{$+0.003$} & $ 0.064$~\fooo{$+0.030$} \\
& ~~~+ MIE Laplace $T_\textrm{opt}$ $\sigma_\textrm{opt}$ & $ 74.99$~\foog{$+0.86$} & $\bf 93.23$~\foog{$+1.29$} & $ 0.944$~\foog{$-0.605$} & $ 0.026$~\foog{$-0.164$} & $\bf 73.04$~\foog{$+0.26$} & $ 90.96$~\fooo{$-0.05$} & $ 1.121$~\foog{$-0.002$} & $ 0.031$~\foog{$-0.003$} \\
\midrule
\parbox[t]{7mm}{\multirow{12}{*}{\rotatebox[origin=c]{90}{\bf Large}}}
& MSDNet (vanilla) & $ 75.36$~\fooe{} & $ 92.78$~\fooe{} & $ 1.475$~\fooee{} & $ 0.178$~\fooee{} & $ 74.33$~\fooe{} & $ 91.57$~\fooe{} & $ 1.066$~\fooee{} & $ 0.050$~\fooee{} \\
& ~~~+ $T_\textrm{opt}$ & $ 75.27$~\fooo{$-0.10$} & $ 92.76$~\fooo{$-0.02$} & $ 0.984$~\foog{$-0.491$} & $ 0.059$~\foog{$-0.119$} & $ 74.33$~\foog{$+0.00$} & $ 91.57$~\foog{$+0.00$} & $ 1.066$~\fooo{$+0.000$} & $ 0.050$~\fooo{$+0.000$} \\
& ~~~+ Laplace & $ 75.41$~\foog{$+0.05$} & $ 92.76$~\fooo{$-0.02$} & $ 1.347$~\foog{$-0.128$} & $ 0.157$~\foog{$-0.021$} & $ 74.25$~\fooo{$-0.08$} & $ 91.55$~\fooo{$-0.02$} & $ 1.052$~\foog{$-0.014$} & $ 0.019$~\foog{$-0.031$} \\
& ~~~+ Laplace $T_\textrm{opt}$ & $ 75.28$~\fooo{$-0.08$} & $ 92.79$~\foog{$+0.01$} & $ 0.999$~\foog{$-0.476$} & $ 0.077$~\foog{$-0.101$} & $ 74.25$~\fooo{$-0.08$} & $ 91.55$~\fooo{$-0.02$} & $ 1.053$~\foog{$-0.012$} & $ 0.019$~\foog{$-0.031$} \\
& ~~~+ Laplace $\sigma_\textrm{opt}$ & $ 75.36$~\fooo{$-0.01$} & $ 92.75$~\fooo{$-0.04$} & $ 1.338$~\foog{$-0.137$} & $ 0.157$~\foog{$-0.020$} & $ 74.25$~\fooo{$-0.08$} & $ 91.55$~\fooo{$-0.03$} & $ 1.053$~\foog{$-0.013$} & $\bf 0.017$~\foog{$-0.033$} \\
& ~~~+ Laplace $T_\textrm{opt}$ $\sigma_\textrm{opt}$ & $ 75.32$~\fooo{$-0.05$} & $ 92.83$~\foog{$+0.05$} & $ 0.996$~\foog{$-0.479$} & $ 0.075$~\foog{$-0.103$} & $ 74.29$~\fooo{$-0.04$} & $ 91.53$~\fooo{$-0.04$} & $ 1.053$~\foog{$-0.013$} & $ 0.020$~\foog{$-0.030$} \\
& ~~~+ MIE & $ 76.32$~\foog{$+0.95$} & $ 93.50$~\foog{$+0.72$} & $ 0.949$~\foog{$-0.525$} & $ 0.061$~\foog{$-0.117$} & $ 74.82$~\foog{$+0.49$} & $ 91.88$~\foog{$+0.30$} & $\bf 1.029$~\foog{$-0.037$} & $ 0.028$~\foog{$-0.022$} \\
& ~~~+ MIE $T_\textrm{opt}$ & $ 76.22$~\foog{$+0.85$} & $ 93.75$~\foog{$+0.97$} & $ 0.886$~\foog{$-0.589$} & $ 0.032$~\foog{$-0.145$} & $\bf 74.90$~\foog{$+0.58$} & $ 91.87$~\foog{$+0.30$} & $ 1.029$~\foog{$-0.037$} & $ 0.022$~\foog{$-0.028$} \\
& ~~~+ MIE Laplace & $\bf 76.43$~\foog{$+1.07$} & $ 93.55$~\foog{$+0.76$} & $ 0.924$~\foog{$-0.551$} & $ 0.040$~\foog{$-0.137$} & $ 74.76$~\foog{$+0.43$} & $\bf 91.90$~\foog{$+0.33$} & $ 1.035$~\foog{$-0.030$} & $ 0.052$~\fooo{$+0.002$} \\
& ~~~+ MIE Laplace $T_\textrm{opt}$ & $ 76.30$~\foog{$+0.93$} & $ 93.74$~\foog{$+0.96$} & $ 0.887$~\foog{$-0.588$} & $ 0.036$~\foog{$-0.142$} & $ 74.86$~\foog{$+0.53$} & $ 91.78$~\foog{$+0.21$} & $ 1.032$~\foog{$-0.034$} & $ 0.026$~\foog{$-0.024$} \\
& ~~~+ MIE Laplace $\sigma_\textrm{opt}$ & $ 76.33$~\foog{$+0.96$} & $ 93.54$~\foog{$+0.75$} & $ 0.925$~\foog{$-0.550$} & $ 0.043$~\foog{$-0.135$} & $ 74.81$~\foog{$+0.49$} & $ 91.87$~\foog{$+0.29$} & $ 1.033$~\foog{$-0.032$} & $ 0.050$~\foog{$-0.000$} \\
& ~~~+ MIE Laplace $T_\textrm{opt}$ $\sigma_\textrm{opt}$ & $ 76.34$~\foog{$+0.98$} & $\bf 93.84$~\foog{$+1.05$} & $\bf 0.885$~\foog{$-0.590$} & $\bf 0.025$~\foog{$-0.152$} & $ 74.80$~\foog{$+0.47$} & $ 91.81$~\foog{$+0.24$} & $ 1.032$~\foog{$-0.034$} & $ 0.032$~\foog{$-0.019$} \\
  \bottomrule
  \end{tabularx}
  \label{tbl:results_appendix}
\end{table*}
\begin{table}[t]
  \caption{Table of Top-1/Top-5 accuracy, NLPD, ECE for different models on Caltech-256. All numbers are averages over a range of computational budgets. `MIE Lap $T_\textrm{opt}$ $\sigma_\textrm{opt}$'-model corresponds to `Our model'. $T_\textrm{opt}$ and $\sigma_\textrm{opt}$ refer to grid search optimisation of the temperature scale and Laplace prior variance, respectively. $n_\text{train}$ is the number of training samples, $d$ is the input dimensionality, $c$ is the number of classes, and $n_\text{batch}$ is the batch size. The red and green numbers show the decrease or increase in performance compared to MSDNet (vanilla). The best performing model for each metric and each model size, on each dataset, is shown in bold.\looseness-1}
  \vspace*{-1em}
  \scriptsize
  \renewcommand{\arraystretch}{.9}
  \newcommand{\fooo}[1]{\textcolor{mycolor2}{\tiny #1}}
   \newcommand{\foog}[1]{\textcolor{mycolor1}{\tiny #1}}
   \newcommand{\foe}[0]{\textcolor{mycolor1}{\tiny $\phantom{+0.0}$}}
      \newcommand{\fooe}[0]{\textcolor{mycolor1}{\tiny $\phantom{+0.00}$}}
      \newcommand{\fooee}[0]{\textcolor{mycolor1}{\tiny $\phantom{+0.000}$}}
  \setlength{\tabcolsep}{0pt}
  \setlength{\tblw}{0.08\textwidth}  
  \begin{tabularx}{.5\textwidth}{l l @{\extracolsep{\fill}} C{\tblw}  C{\tblw} C{\tblw} C{\tblw} }
  \toprule

& & \multicolumn{4}{c}{\sc Caltech-256} \\
& ($n_\textrm{train}$, $d$, $c$, $n_\textrm{batch}$) & \multicolumn{4}{c}{(25607, 150528, 257, 128)} \\
\midrule
& & Top-1 ACC $\uparrow$ & Top-5 ACC $\uparrow$ & NLPD $\downarrow$ & ECE $\downarrow$ \\
\midrule
\parbox[t]{3mm}{\multirow{12}{*}{\rotatebox[origin=c]{90}{\bf Small}}}
& MSDNet (vanilla) & $ 61.0$~\foe{} & $ 78.2$~\foe{} & $ 2.16$~\fooe{} & $ 0.18$~\fooe{} \\
& ~+ $T_\textrm{opt}$ & $ 60.8$~\fooo{$-0.3$} & $ 78.4$~\foog{$+0.2$} & $ 1.84$~\foog{$-0.31$} & $\bf 0.01$~\foog{$-0.16$} \\
& ~+ Lap & $ 60.7$~\fooo{$-0.3$} & $ 78.0$~\fooo{$-0.2$} & $ 1.98$~\foog{$-0.18$} & $ 0.06$~\foog{$-0.12$} \\
& ~+ Lap $T_\textrm{opt}$ & $ 60.5$~\fooo{$-0.6$} & $ 78.3$~\foog{$+0.0$} & $ 1.86$~\foog{$-0.29$} & $ 0.05$~\foog{$-0.13$} \\
& ~+ Lap $\sigma_\textrm{opt}$ & $ 60.6$~\fooo{$-0.4$} & $ 78.0$~\fooo{$-0.2$} & $ 1.99$~\foog{$-0.17$} & $ 0.06$~\foog{$-0.12$} \\
& ~+ Lap $T_\textrm{opt}$ $\sigma_\textrm{opt}$ & $ 60.5$~\fooo{$-0.5$} & $ 78.1$~\fooo{$-0.1$} & $ 1.86$~\foog{$-0.29$} & $ 0.05$~\foog{$-0.13$} \\
& ~+ MIE & $\bf 61.9$~\foog{$+0.9$} & $ 78.8$~\foog{$+0.6$} & $ 1.94$~\foog{$-0.21$} & $ 0.08$~\foog{$-0.10$} \\
& ~+ MIE $T_\textrm{opt}$ & $ 61.5$~\foog{$+0.5$} & $\bf 79.1$~\foog{$+0.9$} & $\bf 1.79$~\foog{$-0.37$} & $ 0.04$~\foog{$-0.13$} \\
& ~+ MIE Lap & $ 61.8$~\foog{$+0.7$} & $ 78.8$~\foog{$+0.5$} & $ 1.86$~\foog{$-0.30$} & $ 0.04$~\foog{$-0.14$} \\
& ~+ MIE Lap $T_\textrm{opt}$ & $ 61.5$~\foog{$+0.5$} & $ 79.1$~\foog{$+0.9$} & $ 1.82$~\foog{$-0.34$} & $ 0.08$~\foog{$-0.10$} \\
& ~+ MIE Lap $\sigma_\textrm{opt}$ & $ 61.8$~\foog{$+0.8$} & $ 79.0$~\foog{$+0.8$} & $ 1.86$~\foog{$-0.30$} & $ 0.03$~\foog{$-0.14$} \\
& ~+ MIE Lap $T_\textrm{opt}$ $\sigma_\textrm{opt}$ & $ 61.7$~\foog{$+0.6$} & $ 79.0$~\foog{$+0.8$} & $ 1.81$~\foog{$-0.34$} & $ 0.09$~\foog{$-0.09$} \\
\midrule
\parbox[t]{3mm}{\multirow{12}{*}{\rotatebox[origin=c]{90}{\bf Medium}}}
& MSDNet (vanilla) & $ 63.8$~\foe{} & $ 80.2$~\foe{} & $ 1.98$~\fooe{} & $ 0.17$~\fooe{} \\
& ~+ $T_\textrm{opt}$ & $ 63.5$~\fooo{$-0.3$} & $ 80.2$~\foog{$+0.0$} & $ 1.70$~\foog{$-0.28$} & $\bf 0.02$~\foog{$-0.15$} \\
& ~+ Lap & $ 63.4$~\fooo{$-0.4$} & $ 79.4$~\fooo{$-0.8$} & $ 1.80$~\foog{$-0.18$} & $ 0.03$~\foog{$-0.14$} \\
& ~+ Lap $T_\textrm{opt}$ & $ 63.3$~\fooo{$-0.5$} & $ 79.8$~\fooo{$-0.3$} & $ 1.73$~\foog{$-0.25$} & $ 0.08$~\foog{$-0.09$} \\
& ~+ Lap $\sigma_\textrm{opt}$ & $ 63.4$~\fooo{$-0.4$} & $ 79.9$~\fooo{$-0.3$} & $ 1.79$~\foog{$-0.19$} & $ 0.04$~\foog{$-0.13$} \\
& ~+ Lap $T_\textrm{opt}$ $\sigma_\textrm{opt}$ & $ 63.4$~\fooo{$-0.4$} & $ 79.9$~\fooo{$-0.3$} & $ 1.74$~\foog{$-0.24$} & $ 0.07$~\foog{$-0.10$} \\
& ~+ MIE & $\bf 65.1$~\foog{$+1.3$} & $ 81.4$~\foog{$+1.2$} & $ 1.72$~\foog{$-0.26$} & $ 0.08$~\foog{$-0.09$} \\
& ~+ MIE $T_\textrm{opt}$ & $ 64.6$~\foog{$+0.9$} & $\bf 81.7$~\foog{$+1.5$} & $\bf 1.61$~\foog{$-0.37$} & $ 0.03$~\foog{$-0.14$} \\
& ~+ MIE Lap & $ 64.9$~\foog{$+1.2$} & $ 81.2$~\foog{$+1.1$} & $ 1.67$~\foog{$-0.31$} & $ 0.06$~\foog{$-0.11$} \\
& ~+ MIE Lap $T_\textrm{opt}$ & $ 64.5$~\foog{$+0.7$} & $ 81.3$~\foog{$+1.1$} & $ 1.67$~\foog{$-0.31$} & $ 0.09$~\foog{$-0.08$} \\
& ~+ MIE Lap $\sigma_\textrm{opt}$ & $ 64.7$~\foog{$+0.9$} & $ 81.2$~\foog{$+1.0$} & $ 1.67$~\foog{$-0.31$} & $ 0.04$~\foog{$-0.13$} \\
& ~+ MIE Lap $T_\textrm{opt}$ $\sigma_\textrm{opt}$ & $ 64.3$~\foog{$+0.5$} & $ 81.3$~\foog{$+1.1$} & $ 1.65$~\foog{$-0.33$} & $ 0.08$~\foog{$-0.09$} \\
\midrule
\parbox[t]{3mm}{\multirow{12}{*}{\rotatebox[origin=c]{90}{\bf Large}}}
& MSDNet (vanilla) & $ 64.9$~\foe{} & $ 80.7$~\foe{} & $ 1.90$~\fooe{} & $ 0.16$~\fooe{} \\
& ~+ $T_\textrm{opt}$ & $ 64.8$~\fooo{$-0.1$} & $ 80.7$~\fooo{$-0.0$} & $ 1.64$~\foog{$-0.26$} & $\bf 0.03$~\foog{$-0.13$} \\
& ~+ Lap & $ 64.3$~\fooo{$-0.6$} & $ 80.1$~\fooo{$-0.6$} & $ 1.72$~\foog{$-0.18$} & $ 0.04$~\foog{$-0.12$} \\
& ~+ Lap $T_\textrm{opt}$ & $ 64.4$~\fooo{$-0.5$} & $ 80.0$~\fooo{$-0.7$} & $ 1.68$~\foog{$-0.22$} & $ 0.07$~\foog{$-0.09$} \\
& ~+ Lap $\sigma_\textrm{opt}$ & $ 64.8$~\fooo{$-0.1$} & $ 80.1$~\fooo{$-0.6$} & $ 1.73$~\foog{$-0.17$} & $ 0.03$~\foog{$-0.13$} \\
& ~+ Lap $T_\textrm{opt}$ $\sigma_\textrm{opt}$ & $ 64.7$~\fooo{$-0.2$} & $ 80.7$~\foog{$+0.0$} & $ 1.65$~\foog{$-0.25$} & $ 0.04$~\foog{$-0.12$} \\
& ~+ MIE & $ 65.9$~\foog{$+0.9$} & $ 82.4$~\foog{$+1.8$} & $ 1.62$~\foog{$-0.28$} & $ 0.06$~\foog{$-0.10$} \\
& ~+ MIE $T_\textrm{opt}$ & $\bf 65.9$~\foog{$+1.0$} & $\bf 82.5$~\foog{$+1.9$} & $\bf 1.54$~\foog{$-0.36$} & $ 0.04$~\foog{$-0.12$} \\
& ~+ MIE Lap & $ 65.9$~\foog{$+1.0$} & $ 82.1$~\foog{$+1.5$} & $ 1.59$~\foog{$-0.31$} & $ 0.08$~\foog{$-0.08$} \\
& ~+ MIE Lap $T_\textrm{opt}$ & $ 65.9$~\foog{$+0.9$} & $ 82.4$~\foog{$+1.8$} & $ 1.59$~\foog{$-0.31$} & $ 0.11$~\foog{$-0.05$} \\
& ~+ MIE Lap $\sigma_\textrm{opt}$ & $ 65.9$~\foog{$+0.9$} & $ 82.4$~\foog{$+1.7$} & $ 1.58$~\foog{$-0.31$} & $ 0.07$~\foog{$-0.09$} \\
& ~+ MIE Lap $T_\textrm{opt}$ $\sigma_\textrm{opt}$ & $ 65.6$~\foog{$+0.7$} & $ 82.5$~\foog{$+1.8$} & $ 1.58$~\foog{$-0.32$} & $ 0.09$~\foog{$-0.07$} \\
  \bottomrule
  \end{tabularx}
  \label{tbl:results_appendix_caltech}
\end{table}
\cref{fig:uncertainty_error_app} repeats the results from \cref{fig:uncertainty_error} and additionally shows corresponding scatter plots for the vanilla MSDNet for comparison. The bottom row in \cref{fig:uncertainty_error_app} shows predictive uncertainty histograms for all samples in the CIFAR-100 test set, comparing the vanilla MSDNet model to our model. The results are obtained using the small CIFAR-100 model. Comparing the scatter plots of our model with those of the vanilla model, we see that the vanilla model has more points in the top left corner of the plots, representing overconfident incorrect predictions. Looking at the bottom row histograms, we observe that the predictions from the vanilla MSDNet are overall more confident than those of our model.

\cref{tbl:results_appendix} shows an extended version of the ablation study seen in \cref{tbl:results}, adding model versions that use Laplace approximation, but optimise only temperature scaling or Laplace prior variance in a grid search, or optimise neither using fixed default values. We also include an ablation result where only temperature scaling is used on the vanilla model predictions. \cref{tbl:results_appendix} also shows results for optimising the temperature scaling parameter when using MIE but without Laplace approximation. Similarly for Caltech-256, \cref{tbl:results_appendix_caltech} shows a more extensive version of \cref{tbl:results_caltech}.

Looking at the results in these tables, we notice that although Laplace approximation alone often slightly decreases top-1 accuracy, when used together with MIE it increases top-1 accuracy above what MIE alone would achieve, suggesting that these two methods are suitable to be used together. In \cref{tbl:results_appendix} the result using only temperature scaling for ImageNet has identical results with the vanilla model, as the best temperature after optimization ended up being the default temperature, suggesting that the vanilla MSDNet on ImageNet is already quite well calibrated. This is reflected in the results in \cref{tbl:results_appendix} also through the fact that our methods that attempt to improve decision-making through improved calibration, do not achieve major improvement in top-1 accuracy on ImageNet, as there is not much room to improve calibration over the vanilla MSDNet. This is likely explained by the vanilla MSDNet underfitting the ImageNet data, as even the largest MSDNet architecture we used for ImageNet is several magnitudes smaller than the state-of-the art models. On CIFAR-100 and Caltech-256 the MSDNet models are large enough to overfit, as is usually the case for most models on most datasets, and we see considerable improvements in also top-1 accuracy.

\begin{figure*}[t!]
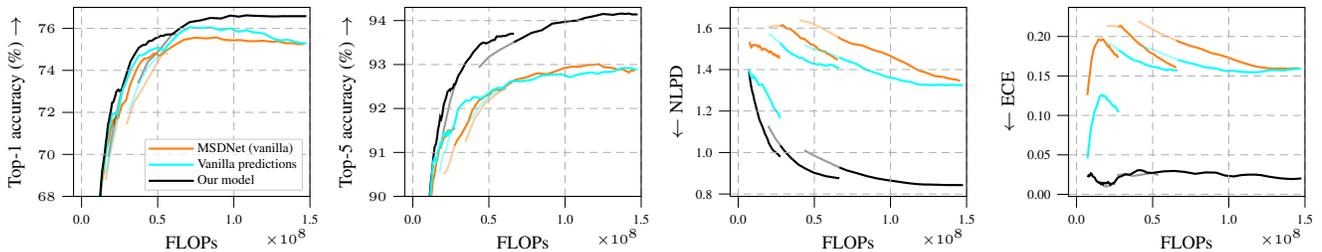

  \centering\scriptsize
  \pgfplotsset{axis on top,scale only axis,width=\figurewidth,height=\figureheight,
    tick label style={font=\tiny},
    grid style={line width=.1pt, draw=gray!05,densely dashed},
    x tick scale label style={yshift=.75em},
    tick scale binop=\times,
    xtick={}
  }
  \setlength{\figurewidth}{.18\textwidth}
  \setlength{\figureheight}{.8\figurewidth}  
  \begin{subfigure}[t]{.24\textwidth}
    \raggedleft
\begin{tikzpicture}

\definecolor{color0}{rgb}{1,0.498039215686275,0.0549019607843137}
\definecolor{color1}{rgb}{0,1,1}

\begin{axis}[
legend cell align={left},
legend style={
  font=\tiny,
  fill opacity=0.8,
  draw opacity=1,
  text opacity=1,
  at={(0.97,0.03)},
  anchor=south east,
  draw=white!80!black
},
tick align=outside,
tick pos=left,
x grid style={white!69.0196078431373!black},
xlabel={FLOPs},
xmajorgrids,
xmin=-5280348.53027194, xmax=150000000,
xtick style={color=black},
xtick={-50000000,0,50000000,100000000,150000000,200000000,250000000,300000000},
xticklabels={−0.5,0.0,0.5,1.0,1.5,2.0,2.5,3.0},
y grid style={white!69.0196078431373!black},
ylabel={Top-1 accuracy (\%) \(\displaystyle \rightarrow\)},
ymajorgrids,
ymin=68, ymax=77,
ytick style={color=black}
]
\addplot [thick, color0]
table {%
7107824 62.33
7433575.5 62.97
7824412 63.61
8216395.5 64.14
8683373 64.85
9287369 65.25
9930998 65.82
10515920 66.43
11173550 66.92
11935525 67.58
12679738 68.14
13384698 68.71
14116088 68.98
14772887 69.35
15558605 70
16324249 70.34
17068580 70.42
17845372 70.68
18466548 70.94
19134888 71.18
19820094 71.35
20431254 71.58
20972820 71.63
21512296 71.48
22070524 71.67
22691414 71.78
23179666 71.81
23633144 71.74
24105429.4119488 71.7649643578724
};
\addlegendentry{MSDNet (vanilla)}

\addplot [thick, black, opacity=0.4, forget plot]
table {%
26383677.4924007 73.0662871572738
26502622.3776224 73.07
26844975.9272237 73.11
27156992.7428124 73.1
};
\addplot [thick, black, forget plot]
table {%
26383677.4924007 73.0662871572738
27114957.2485038 73.28
29833869.7630945 73.9
32193888.3221878 74.43
34902012.545505 74.77
37505023 75.07
39965614.9145572 75.2
42091339.3743385 75.22
44270834.2128857 75.4
46139542.7845058 75.41
48170764.7514962 75.52
50171590.9730974 75.57
52056350.5428953 75.65
53776141.9914138 75.64
55140028.6506293 75.7
56692586.1654135 75.71
58068878.1313194 75.71
59659736.3985777 75.7274168604498
};
\addplot [thick, black, opacity=0.4, forget plot]
table {%
17670232.3676155 69.89
19641217.800252 70.74
21816748.9328533 71.58
26383677.4924007 73.0662871572738
};
\addplot [thick, black, opacity=0.4, forget plot]
table {%
59659736.3985777 75.7274168604498
60628822.0878741 75.72
61616735.0994646 75.71
62521620.4327495 75.71
};
\addplot [thick, black, forget plot]
table {%
59659736.3985777 75.7274168604498
64233524.5772711 75.98
71038681 76.3
77993048.4721301 76.38
84715634.1080382 76.5
91594311.8345746 76.47
97614458.5695152 76.61
103256333.376398 76.56
108160176.534447 76.62
113064533.113946 76.6
117581385.339844 76.59
121380453.806219 76.58
124818427.473434 76.57
128518258.319017 76.57
131570776.498282 76.58
134615441.88127 76.58
137322823.854245 76.58
139457809.29821 76.58
141946674.679242 76.58
144026238.203987 76.58
146046644.008189 76.58
147637730.35694 76.58
};
\addplot [thick, black, opacity=0.4, forget plot]
table {%
37026894.5865197 73.38
43367200.6236015 74.39
49827344.4937234 74.86
59659736.3985777 75.7274168604498
};
\addplot [thick, color1]
table {%
7125618.77811424 62.3
7400580.23537354 62.89
7783853.88093973 63.44
8203447.8974359 63.96
8728716.41176471 64.4
9366414.42625265 64.86
9946913.26569331 65.55
10601025.3990148 66.46
11273676.0225106 66.96
11991577.3333333 67.52
12722417.4660393 68.39
13492461.1029412 69.06
14335418.7962401 69.35
15221740.0552704 69.72
15892448.2571429 70.04
16592819.4363144 70.43
17382477.0011376 71
18061638.4158186 71.16
18710095.8763435 71.43
19329389 71.61
20053034.5407036 71.94
20642700.1366085 71.9
21233120.0923223 71.91
21839705.195231 71.92
22333076.5636856 71.97
22803949.5238403 72.01
23531488.5266145 71.9104464086334
};
\addlegendentry{Vanilla predictions}

\addplot [thick, color0, opacity=0.4, forget plot]
table {%
24105429.4119488 71.7649643578724
24411224 71.8
24769926 71.8
25141960 71.79
};
\addplot [thick, color0, forget plot]
table {%
24105429.4119488 71.7649643578724
24304066 71.83
26532344 72.35
29241978 72.58
31584578 73.24
34147504 73.71
36494240 74.1
38787372 74.38
41154776 74.56
43361168 74.58
45359856 74.66
47402496 74.8
49196656 74.8
50801440 74.69
53011407.411766 74.7800000000001
};
\addplot [thick, color0, opacity=0.4, forget plot]
table {%
15969418 68.78
17779276 69.65
19822496 70.42
24105429.4119488 71.7649643578724
};
\addplot [thick, color0, opacity=0.4, forget plot]
table {%
53011407.411766 74.7800000000001
53927804 74.78
55503776 74.81
57045520 74.78
};
\addplot [thick, color0, forget plot]
table {%
53011407.411766 74.7800000000001
53689056 74.85
60636272 75.18
67259920 75.46
74724928 75.56
81708528 75.52
88334648 75.57
94084560 75.45
99491560 75.43
104465712 75.44
109811712 75.39
114441208 75.39
118463360 75.41
121874664 75.39
125652376 75.36
129252896 75.39
132409536 75.32
135074560 75.29
137454576 75.3
140058576 75.28
141967392 75.25
143816160 75.25
145699632 75.24
};
\addplot [thick, color0, opacity=0.4, forget plot]
table {%
29415024 71.42
34514704 72.47
40481004 73.2
53011407.411766 74.7800000000001
};
\addplot [thick, black]
table {%
7125618.77811424 62.26
7400580.23537354 62.83
7783853.88093973 63.38
8203447.8974359 64.07
8728716.41176471 64.57
9366414.42625265 65.02
9946913.26569331 65.68
10601025.3990148 66.55
11273676.0225106 67.15
11991577.3333333 67.77
12722417.4660393 68.51
13492461.1029412 69.18
14335418.7962401 69.61
15221740.0552704 70.05
15892448.2571429 70.52
16592819.4363144 70.9
17382477.0011376 71.42
18061638.4158186 71.63
18710095.8763435 71.91
19329389 72.1
20053034.5407036 72.46
20642700.1366085 72.47
21233120.0923223 72.57
21839705.195231 72.71
22333076.5636856 72.83
22803949.5238403 72.98
23289995.511524 73.02
23752341.9954955 73.07
24210268.4186222 73.08
24630222.1538462 73.06
24963878.472173 73.02
25412812.5203112 73.04
25821995.3059544 73.07
26383677.4924007 73.0662871572738
};
\addlegendentry{Our model}

\addplot [thick, color1, opacity=0.4, forget plot]
table {%
23531488.5266145 71.9104464086334
23752341.9954955 71.92
24210268.4186222 71.79
24630222.1538462 71.71
};
\addplot [thick, color1, forget plot]
table {%
23531488.5266145 71.9104464086334
24445963.6471636 72.22
27114957.2485038 73.1
29833869.7630945 73.56
32193888.3221878 74.02
34902012.545505 74.36
37505023 74.7
39965614.9145572 74.73
42091339.3743385 74.8
44270834.2128857 74.94
46139542.7845058 74.98
48170764.7514962 75.03
50171590.9730974 75.09
52056350.5428953 75
54314890.7913802 74.9797504974602
};
\addplot [thick, color1, opacity=0.4, forget plot]
table {%
15866300.2508412 69.06
17670232.3676155 69.76
19641217.800252 70.55
23531488.5266145 71.9104464086334
};
\addplot [thick, color1, opacity=0.4, forget plot]
table {%
54314890.7913802 74.9797504974602
55140028.6506293 75.01
56692586.1654135 74.98
58068878.1313194 75
};
\addplot [thick, color1, forget plot]
table {%
54314890.7913802 74.9797504974602
56990304.4845342 75.23
64233524.5772711 75.77
71038681 76.06
77993048.4721301 76.01
84715634.1080382 76.05
91594311.8345746 75.91
97614458.5695152 76
103256333.376398 75.84
108160176.534447 75.91
113064533.113946 75.81
117581385.339844 75.79
121380453.806219 75.7
124818427.473434 75.55
128518258.319017 75.45
131570776.498282 75.43
134615441.88127 75.46
137322823.854245 75.43
139457809.29821 75.37
141946674.679242 75.34
144026238.203987 75.31
146046644.008189 75.29
147637730.35694 75.29
};
\addplot [thick, color1, opacity=0.4, forget plot]
table {%
31276498.1800991 72.1
37026894.5865197 73.26
43367200.6236015 74.14
54314890.7913802 74.9797504974602
};
\addlegendentry{ResNets}
\end{axis}

\end{tikzpicture}
  \end{subfigure}
  \hfill
  \begin{subfigure}[t]{.24\textwidth}
    \raggedleft
    \input{fig/CIFAR100_top5acc_dec.tex}
  \end{subfigure}
  \hfill
  \begin{subfigure}[t]{.24\textwidth}
    \raggedleft
    \input{fig/CIFAR100_NLPD_dec.tex}
  \end{subfigure}
  \hfill
  \begin{subfigure}[t]{.24\textwidth}
    \raggedleft
    \input{fig/CIFAR100_ECE_dec.tex}
  \end{subfigure}
  \vspace*{-1.5em}
   \caption{Accuracy (Top-1 \& Top-5) and uncertainty metrics (NLPD and ECE) on a budgeted batch classification task as a function of average computational budget per image (FLOPs) on the CIFAR-100 data set with a small/medium/large model. These results show a decision-making experiment, where vanilla MSDNet and `Our model' results are compared to results obtained by using a setup where our model is used for decision-making, and predictions are from vanilla MSDNet (labelled `Vanilla predictions').}
\label{fig:cifar100_result_decision}
\end{figure*}

In order to investigate the contribution of better decision-making on the improvements in the predictive performance, we performed an experiment trying to separate the improvement due to better decision-making from the improvement due to better predictions at each individual intermediate exit of the dynamic neural network. In this experiment, we replaced the vanilla model decision-making with the decision-making of our model, while using the vanilla model predictions for calculating the results. Full results for CIFAR-100 are in \cref{fig:cifar100_result_decision} showing that our approach improves both decision-making (orange curve vs. light blue curve) and prediction quality (light blue curve vs. black curve). Interestingly, apart from improving accuracy, better decision-making also improves calibration and uncertainty estimates, as seen from the improved ECE and NLPD. However, this experiment is  problematic in providing information on decision-making quality, as one model is making decisions using predictions from another model, potentially resulting in false interpretation of bad decision-making, if different models predict different samples correctly. 

In addition to the experiments shown, we experimented using predictive variance or entropy as the uncertainty metric to make decisions on when to exit the MSDNet pipeline. In our experiments these metrics performed worse than the model predicted confidence, and hence we use model predictive confidence for decision-making in all experiments shown in this paper.

\begin{table*}[t!]
  \caption{Table of Top-1/Top-5 accuracy, negative log-predictive density (NLPD), and expected calibration error (ECE) for different models on CIFAR-100. `Our model' corresponds to `MIE Laplace $T_\textrm{opt}$ $\sigma_\textrm{opt}$'-model in other result tables. These results show a decision-making experiment, where vanilla MSDNet and `Our model' results are compared to results obtained by using a setup where our model is used for decision-making, and predictions are from vanilla MSDNet. The best-performing model for each metric and each model size is shown in bold.\looseness-1}
  \vspace*{-1em}
  \newcommand{\fooo}[1]{\textcolor{mycolor2}{\tiny #1}}
   \newcommand{\foog}[1]{\textcolor{mycolor1}{\tiny #1}}
      \newcommand{\fooe}[0]{\textcolor{mycolor1}{\tiny $\phantom{+0.00}$}}
      \newcommand{\fooee}[0]{\textcolor{mycolor1}{\tiny $\phantom{+0.000}$}}
  \footnotesize
  \setlength{\tabcolsep}{0pt}
  \setlength{\tblw}{0.1\textwidth}  
  \begin{tabularx}{\textwidth}{l l @{\extracolsep{\fill}} C{\tblw}  C{\tblw} C{\tblw} C{\tblw} }
  \toprule

& & \multicolumn{4}{c}{\sc CIFAR-100} \\
& ($n_\textrm{train}$, $d$, $c$, $n_\textrm{batch}$) & \multicolumn{4}{c}{(50000, 3072,100, 64)}  \\
\midrule
& & Top-1 ACC $\uparrow$ & Top-5 ACC $\uparrow$ & NLPD $\downarrow$ & ECE $\downarrow$ \\
\midrule
\parbox[t]{7mm}{\multirow{3}{*}{\rotatebox[origin=c]{90}{\bf Small}}}
& MSDNet (vanilla) & $ 69.25$~\fooe{} & $ 90.48$~\fooe{} & $ 1.498$~\fooee{} & $ 0.182$~\fooee{}  \\
& Vanilla predictions, our model decisions & $ 69.33$~\foog{$+0.09$} & $ 90.60$~\foog{$+0.12$} & $ 1.300$~\foog{$-0.197$} & $ 0.108$~\foog{$-0.074$}  \\
& Our model & $\bf  69.84$~\foog{$+0.59$} & $\bf  91.09$~\foog{$+0.61$} & $\bf  1.133$~\foog{$-0.364$} & $\bf 0.017$~\foog{$-0.165$} \\
\midrule
\parbox[t]{7mm}{\multirow{3}{*}{\rotatebox[origin=c]{90}{\bf Medium}}}
& MSDNet (vanilla) & $ 74.12$~\fooe{} & $ 91.94$~\fooe{} & $ 1.549$~\fooee{} & $ 0.190$~\fooee{} \\
& Vanilla predictions, our model decisions & $ 74.51$~\foog{$+0.39$} & $ 92.20$~\foog{$+0.25$} & $ 1.460$~\foog{$-0.089$} & $ 0.168$~\foog{$-0.022$} \\
& Our model & $\bf  74.99$~\foog{$+0.86$} & $\bf  93.23$~\foog{$+1.29$} & $\bf  0.944$~\foog{$-0.605$} & $\bf 0.026$~\foog{$-0.164$} \\
\midrule
\parbox[t]{7mm}{\multirow{3}{*}{\rotatebox[origin=c]{90}{\bf Large}}}
& MSDNet (vanilla) & $ 75.36$~\fooe{} & $ 92.78$~\fooe{} & $ 1.475$~\fooee{} & $ 0.178$~\fooee{} \\
& Vanilla predictions, our model decisions & $ 75.72$~\foog{$+0.36$} & $ 92.71$~\fooo{$-0.08$} & $ 1.388$~\foog{$-0.086$} & $ 0.162$~\foog{$-0.015$} \\
& Our model & $\bf  76.34$~\foog{$+0.98$} & $\bf  93.84$~\foog{$+1.05$} & $\bf 0.885$~\foog{$-0.590$} & $\bf  0.025$~\foog{$-0.152$}\\
  \bottomrule
  \end{tabularx}
  \label{tbl:results_decision}
\end{table*}

\section{Analysis on Laplace Approximation and MIE Computational Cost}
\label{app:laplace_cost}
Using the Laplace approximation requires calculating the approximate inverse Hessians $\MH^{-1}$ and $\MSigma_\vb$ for the last layer weights and biases respectively, for each intermediate classifier. As this can be precomputed before observing the test data, the only additional test time cost of the Laplace approximation comes from transforming the Laplace approximated distribution for the last layer weights $\mathrm{N}(\vtheta\mid\vtheta_{\text{MAP}}, \MSigma)$ to an output predictive distribution $p(\hat{\vz}_i \mid \mbf{x}_i)$ and sampling from this distribution for $n_\text{MC}$ times. Recall that $\hat{\vy}_i = \softmax(\hat{\vz}_{i} )$, where $\hat{\vz}_{i} = \MW\vphi_{i} + \vb$.

\paragraph{Na\"ive approach} We use the KFAC approximation to the inverse Hessian $\MH^{-1} = \MV^{-1} \otimes \MU^{-1}$.
Let $\MW \in \R^{p \times c}$, $\vb \in \R^{1 \times c}$ denote the weight matrix and bias terms of the $k$\textsuperscript{th} exit, and $\vphi_i \in \R^p$ denote the features of input sample $\vx_i$ before the last linear layer of the $k$\textsuperscript{th} exit.
Then the additional cost associated to a \naive implementation of the Laplace approximation at the $k$\textsuperscript{th} exit is based on
\begin{equation}
  \hat{\vz}_i \sim \mathrm{N}(\underbrace{\MW_\text{MAP}^\top \vphi_i + \vb_\text{MAP}}_{\text{\scriptsize also needed for vanilla }}, \underbrace{(\vphi_i^\top \MV \vphi_i)\MU + \MSigma_\vb}_{(p+1)(2p-1) + 2c^{2}})
  \label{eq:naive}
\end{equation}
with additional costs to compute the Cholesky factorisation ($\frac{1}{3}c^{3}$) and rescaling and shifting the samples drawn from a standard normal, resulting in a total of
\begin{equation}
  \text{FLOPs}_{\text{\naive}} = 2c^{2}(n_{\text{MC}}+1) + \frac{1}{3}c^{3} + 2p^2 + p -1.
\end{equation}
additional FLOPs. The calculation of the mean in \cref{eq:naive} does not add computation as this operation is also performed to obtain the vanilla MSDNet prediction. The cubic scaling of FLOPs with the number of classes is what makes the \naive approach prohibitively expensive to be viable in the budget restricted regime if the number of classes is large. The Cholesky factorization of the covariance matrix that is required for sampling can not be pre-computed before test time as the covariance matrix $(\vphi_i^\top \MV \vphi_i)\MU + \MSigma_\vb$ depends on the test samples.

\paragraph{Efficient approach}
Sampling from the Laplace predictive distribution can be made more efficient by absorbing the bias terms into the weight matrix, \ie, $\hat\MW \in \R^{p+1 \times c}$.
This appends an additional dimension to $\vphi$ and $\MV$ which we now denote by $\hat\vphi = (\vphi^{\top}, 1)^{\top}$ and $\hat\MV$, respectively. Now, the output predictive distribution takes the form $\hat{\vz}_i \sim \mathrm{N}(\hat\MW_\text{MAP}^\top \hat\vphi_i, (\hat\vphi_i^\top \hat\MV \hat\vphi_i)\MU)$, which means that the costly operations can all be pre-computed offline. Most notably, although the covariance matrix $(\hat\vphi_i^\top \hat\MV \hat\vphi_i)\MU$ still depends on the test samples, the test sample dependent term $\hat\vphi_i^\top \hat\MV \hat\vphi_i$ is a scalar and can be taken out of the costly Cholesky factorization, which allows performing this on the matrix $\MU$ defore test time. This means that for one MC sample $l$, the pre-softmax output can be evaluated as
\begin{equation}
 \hat{\vz}_i^{(l)} =  \underbrace{\hat\MW_\text{MAP}^\top \hat\vphi_i}_{\text{\scriptsize also needed for vanilla }} + \underbrace{\sqrt{\hat\vphi_i^\top \hat\MV \hat\vphi_i}}_{(p+2)(2p+1)} \underbrace{\big(\ML\vg^{(l)}\big)}_{2c^{2}-c},
\end{equation}
where $\vg^{(l)} \sim \mathrm{N}(\bm{0},\MI)$ and $\ML$ is the pre-calculated Cholesky factor of $\MU$. Even the samples $\vg^{(l)}$ can be pre-drawn and pre-multiplied with $\ML$, which further reduces the computational cost to
\begin{equation}
  \text{FLOPs}_{\text{efficient}} = 2cn_{\text{MC}} + 2p^2 + 5p + 2.
\end{equation}

\begin{figure*}
  \centering\scriptsize
  \pgfplotsset{axis on top,scale only axis,width=\figurewidth,height=\figureheight,
    tick label style={font=\tiny},
    x tick scale label style={yshift=.75em},
    tick scale binop=\times,
    xtick={}
  }
  \setlength{\figurewidth}{.26\textwidth}
  \setlength{\figureheight}{0.5\figurewidth} 
  \begin{subfigure}[b]{\textwidth}
    \raggedright
    \tikz\node[font=\bf,fill=C0,rounded corners=1pt]{CIFAR-100};
  \end{subfigure}\\
  \begin{subfigure}[b]{.32\textwidth}
    \raggedleft
\begin{tikzpicture}

\definecolor{color0}{rgb}{1,0.498039215686275,0.0549019607843137}

\begin{axis}[
height=\figureheight,
tick align=outside,
tick pos=left,
width=\figurewidth,
x grid style={white!69.0196078431373!black},
xmin=0.52, xmax=4.48,
xtick style={color=black},
y grid style={white!69.0196078431373!black},
ylabel={FLOPs},
ymin=0, ymax=43616350.4,
ytick style={color=black}
]
\draw[draw=none,fill=color0] (axis cs:1,0) rectangle (axis cs:0.7,6860000);
\draw[draw=none,fill=color0] (axis cs:2,0) rectangle (axis cs:1.7,14350000);
\draw[draw=none,fill=color0] (axis cs:3,0) rectangle (axis cs:2.7,26130000);
\draw[draw=none,fill=color0] (axis cs:4,0) rectangle (axis cs:3.7,38040000);
\draw[draw=none,fill=black] (axis cs:1,0) rectangle (axis cs:1.3,6886770);
\draw[draw=none,fill=black] (axis cs:2,0) rectangle (axis cs:2.3,14403540);
\draw[draw=none,fill=black] (axis cs:3,0) rectangle (axis cs:3.3,26210310);
\draw[draw=none,fill=black] (axis cs:4,0) rectangle (axis cs:4.3,38147080);
\draw[draw=none,fill=black,fill opacity=0.5] (axis cs:1,0) rectangle (axis cs:1.3,7734845.33333333);
\draw[draw=none,fill=black,fill opacity=0.5] (axis cs:2,0) rectangle (axis cs:2.3,16099690.6666667);
\draw[draw=none,fill=black,fill opacity=0.5] (axis cs:3,0) rectangle (axis cs:3.3,28754536);
\draw[draw=none,fill=black,fill opacity=0.5] (axis cs:4,0) rectangle (axis cs:4.3,41539381.3333333);
\end{axis}

\end{tikzpicture}
  \end{subfigure}
  \hfill
  \begin{subfigure}[b]{.32\textwidth}
    \raggedleft
\begin{tikzpicture}

\definecolor{color0}{rgb}{1,0.498039215686275,0.0549019607843137}

\begin{axis}[
height=\figureheight,
tick align=outside,
tick pos=left,
width=\figurewidth,
x grid style={white!69.0196078431373!black},
xmin=0.42, xmax=6.58,
xtick style={color=black},
y grid style={white!69.0196078431373!black},
ymin=0, ymax=99056025.6,
ytick style={color=black},
xtick = {1,2,3,4,5,6},
xticklabels = {1,2,3,4,5,6}
]
\draw[draw=none,fill=color0] (axis cs:1,0) rectangle (axis cs:0.7,6860000);
\draw[draw=none,fill=color0] (axis cs:2,0) rectangle (axis cs:1.7,14350000);
\draw[draw=none,fill=color0] (axis cs:3,0) rectangle (axis cs:2.7,27290000);
\draw[draw=none,fill=color0] (axis cs:4,0) rectangle (axis cs:3.7,46560000);
\draw[draw=none,fill=color0] (axis cs:5,0) rectangle (axis cs:4.7,67430000);
\draw[draw=none,fill=color0] (axis cs:6,0) rectangle (axis cs:5.7,89090000);
\draw[draw=none,fill=black] (axis cs:1,0) rectangle (axis cs:1.3,6886770);
\draw[draw=none,fill=black] (axis cs:2,0) rectangle (axis cs:2.3,14403540);
\draw[draw=none,fill=black] (axis cs:3,0) rectangle (axis cs:3.3,27370310);
\draw[draw=none,fill=black] (axis cs:4,0) rectangle (axis cs:4.3,46667080);
\draw[draw=none,fill=black] (axis cs:5,0) rectangle (axis cs:5.3,67563850);
\draw[draw=none,fill=black] (axis cs:6,0) rectangle (axis cs:6.3,89250620);
\draw[draw=none,fill=black,fill opacity=0.5] (axis cs:1,0) rectangle (axis cs:1.3,7734845.33333333);
\draw[draw=none,fill=black,fill opacity=0.5] (axis cs:2,0) rectangle (axis cs:2.3,16099690.6666667);
\draw[draw=none,fill=black,fill opacity=0.5] (axis cs:3,0) rectangle (axis cs:3.3,29914536);
\draw[draw=none,fill=black,fill opacity=0.5] (axis cs:4,0) rectangle (axis cs:4.3,50059381.3333333);
\draw[draw=none,fill=black,fill opacity=0.5] (axis cs:5,0) rectangle (axis cs:5.3,71804226.6666667);
\draw[draw=none,fill=black,fill opacity=0.5] (axis cs:6,0) rectangle (axis cs:6.3,94339072);
\end{axis}

\end{tikzpicture}
  \end{subfigure}
  \hfill
  \begin{subfigure}[b]{.32\textwidth}
    \raggedleft
    \pgfplotsset{ybar,legend image code/.code={\draw [#1,draw=none] (0cm,-0.08cm) rectangle (0.15cm,0.15cm);}}
\begin{tikzpicture}

\definecolor{color0}{rgb}{1,0.498039215686275,0.0549019607843137}

\begin{axis}[
height=\figureheight,
legend cell align={left},
legend style={
  font=\tiny,
  fill opacity=0.8,
  draw opacity=1,
  text opacity=1,
  at={(0.03,0.97)},
  anchor=north west,
  draw=white!80!black
},
tick align=outside,
tick pos=left,
width=\figurewidth,
x grid style={white!69.0196078431373!black},
xmin=0.32, xmax=8.68,
xtick style={color=black},
y grid style={white!69.0196078431373!black},
ymin=0, ymax=209673200.8,
ytick style={color=black},
xtick = {1,2,3,4,5,6,7,8},
xticklabels = {1,2,3,4,5,6,7,8}
]
\draw[draw=none,fill=color0] (axis cs:1,0) rectangle (axis cs:0.7,6860000);
\addlegendimage{ybar,draw=none,fill=color0};
\addlegendentry{Vanilla MSDNet}

\draw[draw=none,fill=color0] (axis cs:2,0) rectangle (axis cs:1.7,14350000);
\draw[draw=none,fill=color0] (axis cs:3,0) rectangle (axis cs:2.7,27290000);
\draw[draw=none,fill=color0] (axis cs:4,0) rectangle (axis cs:3.7,48450000);
\draw[draw=none,fill=color0] (axis cs:5,0) rectangle (axis cs:4.7,81570000);
\draw[draw=none,fill=color0] (axis cs:6,0) rectangle (axis cs:5.7,112640000);
\draw[draw=none,fill=color0] (axis cs:7,0) rectangle (axis cs:6.7,152920000);
\draw[draw=none,fill=color0] (axis cs:8,0) rectangle (axis cs:7.7,192690000);
\draw[draw=none,fill=black] (axis cs:1,0) rectangle (axis cs:1.3,6886770);
\addlegendimage{ybar,draw=none,fill=black};
\addlegendentry{Efficient Laplace}

\draw[draw=none,fill=black] (axis cs:2,0) rectangle (axis cs:2.3,14403540);
\draw[draw=none,fill=black] (axis cs:3,0) rectangle (axis cs:3.3,27370310);
\draw[draw=none,fill=black] (axis cs:4,0) rectangle (axis cs:4.3,48557080);
\draw[draw=none,fill=black] (axis cs:5,0) rectangle (axis cs:5.3,81703850);
\draw[draw=none,fill=black] (axis cs:6,0) rectangle (axis cs:6.3,112800620);
\draw[draw=none,fill=black] (axis cs:7,0) rectangle (axis cs:7.3,153107390);
\draw[draw=none,fill=black] (axis cs:8,0) rectangle (axis cs:8.3,192904160);
\draw[draw=none,fill=black,fill opacity=0.5] (axis cs:1,0) rectangle (axis cs:1.3,7734845.33333333);
\addlegendimage{ybar,draw=none,fill=black,fill opacity=0.5};
\addlegendentry{Naive Laplace}

\draw[draw=none,fill=black,fill opacity=0.5] (axis cs:2,0) rectangle (axis cs:2.3,16099690.6666667);
\draw[draw=none,fill=black,fill opacity=0.5] (axis cs:3,0) rectangle (axis cs:3.3,29914536);
\draw[draw=none,fill=black,fill opacity=0.5] (axis cs:4,0) rectangle (axis cs:4.3,51949381.3333333);
\draw[draw=none,fill=black,fill opacity=0.5] (axis cs:5,0) rectangle (axis cs:5.3,85944226.6666667);
\draw[draw=none,fill=black,fill opacity=0.5] (axis cs:6,0) rectangle (axis cs:6.3,117889072);
\draw[draw=none,fill=black,fill opacity=0.5] (axis cs:7,0) rectangle (axis cs:7.3,159043917.333333);
\draw[draw=none,fill=black,fill opacity=0.5] (axis cs:8,0) rectangle (axis cs:8.3,199688762.666667);
\end{axis}

\end{tikzpicture}
  \end{subfigure}\\
  \begin{subfigure}[t]{.32\textwidth}
    \raggedright
\begin{tikzpicture}

\begin{axis}[
height=\figureheight,
tick align=outside,
tick pos=left,
width=\figurewidth,
x grid style={white!69.0196078431373!black},
xlabel={Exit number},
xmin=0.52, xmax=4.48,
xtick style={color=black},
y grid style={white!69.0196078431373!black},
ylabel={FLOPs relative to vanilla},
ymin=1, ymax=1.15,
ytick style={color=black}
]
\addplot [semithick, black, mark=asterisk, mark size=3, mark options={solid}, only marks]
table {%
1 1.12752847424684
2 1.12192966318235
3 1.10044148488328
4 1.09199214861549
};
\addplot [semithick, black, mark=*, mark size=3, mark options={solid}, only marks]
table {%
1 1.00390233236152
2 1.00373101045296
3 1.00307347876005
4 1.00281493165089
};
\end{axis}

\end{tikzpicture}
  \end{subfigure}
  \hfill
  \begin{subfigure}[t]{.32\textwidth}
    \raggedleft
\begin{tikzpicture}

\begin{axis}[
height=\figureheight,
tick align=outside,
tick pos=left,
width=\figurewidth,
x grid style={white!69.0196078431373!black},
xlabel={Exit number},
xmin=0.42, xmax=6.58,
xtick style={color=black},
y grid style={white!69.0196078431373!black},
ymin=1, ymax=1.15,
ytick style={color=black},
xtick = {1,2,3,4,5,6},
xticklabels = {1,2,3,4,5,6}
]
\addplot [semithick, black, mark=asterisk, mark size=3, mark options={solid}, only marks]
table {%
1 1.12752847424684
2 1.12192966318235
3 1.09617207768413
4 1.07515853379152
5 1.06487063127194
6 1.05891875631384
};
\addplot [semithick, black, mark=*, mark size=3, mark options={solid}, only marks]
table {%
1 1.00390233236152
2 1.00373101045296
3 1.00294283620374
4 1.00229982817869
5 1.00198502150378
6 1.00180289594792
};
\end{axis}

\end{tikzpicture}
  \end{subfigure}
  \hfill
  \begin{subfigure}[t]{.32\textwidth}
    \raggedleft
\begin{tikzpicture}

\begin{axis}[
height=\figureheight,
legend cell align={left},
legend style={  font=\tiny,fill opacity=0.8, draw opacity=1, text opacity=1, draw=white!80!black},
tick align=outside,
tick pos=left,
width=\figurewidth,
x grid style={white!69.0196078431373!black},
xlabel={Exit number},
xmin=0.32, xmax=8.68,
xtick style={color=black},
y grid style={white!69.0196078431373!black},
ymin=1, ymax=1.15,
ytick style={color=black},
xtick = {1,2,3,4,5,6,7,8},
xticklabels = {1,2,3,4,5,6,7,8}
]
\addplot [semithick, black, mark=asterisk, mark size=3, mark options={solid}, only marks]
table {%
1 1.12752847424684
2 1.12192966318235
3 1.09617207768413
4 1.07222665290678
5 1.05362543418741
6 1.04660042613636
7 1.04004654285465
8 1.03632135900496
};
\addlegendentry{Naive Laplace}
\addplot [semithick, black, mark=*, mark size=3, mark options={solid}, only marks]
table {%
1 1.00390233236152
2 1.00373101045296
3 1.00294283620374
4 1.00221011351909
5 1.00164092190756
6 1.00142595880682
7 1.00122541198012
8 1.00111142249209
};
\addlegendentry{Efficient Laplace}
\end{axis}

\end{tikzpicture}
  \end{subfigure}\\
  
  \begin{subfigure}[b]{\textwidth}
    \raggedright
    \tikz\node[font=\bf,fill=C0,rounded corners=1pt]{ImageNet};
  \end{subfigure}\\
  \begin{subfigure}[b]{.32\textwidth}
    \raggedleft
\begin{tikzpicture}

\definecolor{color0}{rgb}{1,0.498039215686275,0.0549019607843137}

\begin{axis}[
height=\figureheight,
tick align=outside,
tick pos=left,
width=\figurewidth,
x grid style={white!69.0196078431373!black},
xmin=0.47, xmax=5.53,
xtick style={color=black},
y grid style={white!69.0196078431373!black},
ylabel={FLOPs},
ymin=0, ymax=3453468340,
ytick style={color=black},
xtick = {1,2,3,4,5},
xticklabels = {1,2,3,4,5}
]
\draw[draw=none,fill=color0] (axis cs:1,0) rectangle (axis cs:0.7,339900000);
\draw[draw=none,fill=color0] (axis cs:2,0) rectangle (axis cs:1.7,685460000);
\draw[draw=none,fill=color0] (axis cs:3,0) rectangle (axis cs:2.7,1008160000);
\draw[draw=none,fill=color0] (axis cs:4,0) rectangle (axis cs:3.7,1254470000);
\draw[draw=none,fill=color0] (axis cs:5,0) rectangle (axis cs:4.7,1360530000);
\draw[draw=none,fill=black] (axis cs:1,0) rectangle (axis cs:1.3,340148610);
\draw[draw=none,fill=black] (axis cs:2,0) rectangle (axis cs:2.3,685957220);
\draw[draw=none,fill=black] (axis cs:3,0) rectangle (axis cs:3.3,1008834886);
\draw[draw=none,fill=black] (axis cs:4,0) rectangle (axis cs:4.3,1255243320);
\draw[draw=none,fill=black] (axis cs:5,0) rectangle (axis cs:5.3,1362606410);
\draw[draw=none,fill=black,fill opacity=0.5] (axis cs:1,0) rectangle (axis cs:1.3,725431173.333333);
\draw[draw=none,fill=black,fill opacity=0.5] (axis cs:2,0) rectangle (axis cs:2.3,1456522346.66667);
\draw[draw=none,fill=black,fill opacity=0.5] (axis cs:3,0) rectangle (axis cs:3.3,2164682768);
\draw[draw=none,fill=black,fill opacity=0.5] (axis cs:4,0) rectangle (axis cs:4.3,2796374213.33333);
\draw[draw=none,fill=black,fill opacity=0.5] (axis cs:5,0) rectangle (axis cs:5.3,3289017466.66667);
\end{axis}

\end{tikzpicture}
  \end{subfigure}
  \hfill
  \begin{subfigure}[b]{.32\textwidth}
    \raggedleft
\begin{tikzpicture}

\definecolor{color0}{rgb}{1,0.498039215686275,0.0549019607843137}

\begin{axis}[
height=\figureheight,
tick align=outside,
tick pos=left,
width=\figurewidth,
x grid style={white!69.0196078431373!black},
xmin=0.47, xmax=5.53,
xtick style={color=black},
y grid style={white!69.0196078431373!black},
ymin=0, ymax=4905722794,
ytick style={color=black},
xtick = {1,2,3,4,5},
xticklabels = {1,2,3,4,5}
]
\draw[draw=none,fill=color0] (axis cs:1,0) rectangle (axis cs:0.7,514660000);
\draw[draw=none,fill=color0] (axis cs:2,0) rectangle (axis cs:1.7,1171180000);
\draw[draw=none,fill=color0] (axis cs:3,0) rectangle (axis cs:2.7,1844520000);
\draw[draw=none,fill=color0] (axis cs:4,0) rectangle (axis cs:3.7,2501400000);
\draw[draw=none,fill=color0] (axis cs:5,0) rectangle (axis cs:4.7,2742060000);
\draw[draw=none,fill=black] (axis cs:1,0) rectangle (axis cs:1.3,515023682);
\draw[draw=none,fill=black] (axis cs:2,0) rectangle (axis cs:2.3,1171975140);
\draw[draw=none,fill=black] (axis cs:3,0) rectangle (axis cs:3.3,1845562518);
\draw[draw=none,fill=black] (axis cs:4,0) rectangle (axis cs:4.3,2504908168);
\draw[draw=none,fill=black] (axis cs:5,0) rectangle (axis cs:5.3,2745708210);
\draw[draw=none,fill=black,fill opacity=0.5] (axis cs:1,0) rectangle (axis cs:1.3,900305989.333333);
\draw[draw=none,fill=black,fill opacity=0.5] (axis cs:2,0) rectangle (axis cs:2.3,1942539626.66667);
\draw[draw=none,fill=black,fill opacity=0.5] (axis cs:3,0) rectangle (axis cs:3.3,3001409536);
\draw[draw=none,fill=black,fill opacity=0.5] (axis cs:4,0) rectangle (axis cs:4.3,4046034453.33333);
\draw[draw=none,fill=black,fill opacity=0.5] (axis cs:5,0) rectangle (axis cs:5.3,4672116946.66667);
\end{axis}

\end{tikzpicture}
  \end{subfigure}
  \hfill
  \begin{subfigure}[b]{.32\textwidth}
    \raggedleft
    \pgfplotsset{ybar,legend image code/.code={\draw [#1,draw=none] (0cm,-0.08cm) rectangle (0.15cm,0.15cm);}}
\begin{tikzpicture}

\definecolor{color0}{rgb}{1,0.498039215686275,0.0549019607843137}

\begin{axis}[
height=\figureheight,
legend cell align={left},
legend style={
  font=\tiny,
  fill opacity=0.8,
  draw opacity=1,
  text opacity=1,
  at={(0.03,0.97)},
  anchor=north west,
  draw=white!80!black
},
tick align=outside,
tick pos=left,
width=\figurewidth,
x grid style={white!69.0196078431373!black},
xmin=0.47, xmax=5.53,
xtick style={color=black},
y grid style={white!69.0196078431373!black},
ymin=0, ymax=5444748148,
ytick style={color=black},
xtick = {1,2,3,4,5},
xticklabels = {1,2,3,4,5}
]
\draw[draw=none,fill=color0] (axis cs:1,0) rectangle (axis cs:0.7,615600000);
\addlegendimage{ybar,draw=none,fill=color0};
\addlegendentry{Vanilla MSDNet}

\draw[draw=none,fill=color0] (axis cs:2,0) rectangle (axis cs:1.7,1436390000);
\draw[draw=none,fill=color0] (axis cs:3,0) rectangle (axis cs:2.7,2283210000);
\draw[draw=none,fill=color0] (axis cs:4,0) rectangle (axis cs:3.7,2967420000);
\draw[draw=none,fill=color0] (axis cs:5,0) rectangle (axis cs:4.7,3253790000);
\draw[draw=none,fill=black] (axis cs:1,0) rectangle (axis cs:1.3,616033506);
\addlegendimage{ybar,draw=none,fill=black};
\addlegendentry{Efficient Laplace}

\draw[draw=none,fill=black] (axis cs:2,0) rectangle (axis cs:2.3,1437413044);
\draw[draw=none,fill=black] (axis cs:3,0) rectangle (axis cs:3.3,2284624470);
\draw[draw=none,fill=black] (axis cs:4,0) rectangle (axis cs:4.3,2968941480);
\draw[draw=none,fill=black] (axis cs:5,0) rectangle (axis cs:5.3,3259067530);
\draw[draw=none,fill=black,fill opacity=0.5] (axis cs:1,0) rectangle (axis cs:1.3,1001315685.33333);
\addlegendimage{ybar,draw=none,fill=black,fill opacity=0.5};
\addlegendentry{Na\"ive Laplace}

\draw[draw=none,fill=black,fill opacity=0.5] (axis cs:2,0) rectangle (axis cs:2.3,2207977146.66667);
\draw[draw=none,fill=black,fill opacity=0.5] (axis cs:3,0) rectangle (axis cs:3.3,3440470816);
\draw[draw=none,fill=black,fill opacity=0.5] (axis cs:4,0) rectangle (axis cs:4.3,4510070581.33333);
\draw[draw=none,fill=black,fill opacity=0.5] (axis cs:5,0) rectangle (axis cs:5.3,5185474426.66667);
\end{axis}

\end{tikzpicture}
  \end{subfigure}\\
  
  \begin{subfigure}[t]{.32\textwidth}
    \raggedleft
\begin{tikzpicture}

\begin{axis}[
height=\figureheight,
tick align=outside,
tick pos=left,
width=\figurewidth,
x grid style={white!69.0196078431373!black},
xlabel={Exit number},
xmin=0.47, xmax=5.53,
xtick style={color=black},
y grid style={white!69.0196078431373!black},
ylabel={FLOPs relative to vanilla},
ymin=1, ymax=2.75,
ytick style={color=black},
xtick = {1,2,3,4,5},
xticklabels = {1,2,3,4,5}
]
\addplot [semithick, black, mark=asterisk, mark size=3, mark options={solid}, only marks]
table {%
1 2.13424881827989
2 2.12488306635933
3 2.1471619266783
4 2.22912800890682
5 2.4174531003849
};
\addplot [semithick, black, mark=*, mark size=3, mark options={solid}, only marks]
table {%
1 1.00073142100618
2 1.00072538149564
3 1.00066942350421
4 1.00061645156919
5 1.00152617729855
};
\end{axis}

\end{tikzpicture}
  \end{subfigure}
  \hfill
  \begin{subfigure}[t]{.32\textwidth}
     \raggedleft
\begin{tikzpicture}

\begin{axis}[
height=\figureheight,
tick align=outside,
tick pos=left,
width=\figurewidth,
x grid style={white!69.0196078431373!black},
xlabel={Exit number},
xmin=0.47, xmax=5.53,
xtick style={color=black},
y grid style={white!69.0196078431373!black},
ymin=1, ymax=2.75,
ytick style={color=black},
xtick = {1,2,3,4,5},
xticklabels = {1,2,3,4,5}
]
\addplot [semithick, black, mark=asterisk, mark size=3, mark options={solid}, only marks]
table {%
1 1.74932186168215
2 1.65861748549896
3 1.62720357382951
4 1.61750797686629
5 1.70387115769409
};
\addplot [semithick, black, mark=*, mark size=3, mark options={solid}, only marks]
table {%
1 1.0007066451638
2 1.00067892211274
3 1.00056519744974
4 1.00140248181019
5 1.00133046322838
};
\end{axis}

\end{tikzpicture}
  \end{subfigure}
  \hfill
  \begin{subfigure}[t]{.32\textwidth}
     \raggedleft
\begin{tikzpicture}

\begin{axis}[
height=\figureheight,
legend cell align={left},
legend style={  font=\tiny,fill opacity=0.8, draw opacity=1, text opacity=1, draw=white!80!black},
tick align=outside,
tick pos=left,
width=\figurewidth,
x grid style={white!69.0196078431373!black},
xlabel={Exit number},
xmin=0.47, xmax=5.53,
xtick style={color=black},
y grid style={white!69.0196078431373!black},
ymin=1, ymax=2.75,
ytick style={color=black},
xtick = {1,2,3,4,5},
xticklabels = {1,2,3,4,5}
]
\addplot [semithick, black, mark=asterisk, mark size=3, mark options={solid}, only marks]
table {%
1 1.6265686896253
2 1.53717106542559
3 1.50685693212626
4 1.51986256793219
5 1.59367212594134
};
\addlegendentry{Naive Laplace}
\addplot [semithick, black, mark=*, mark size=3, mark options={solid}, only marks]
table {%
1 1.00070420077973
2 1.00071223275016
3 1.00061950937496
4 1.00051272822856
5 1.00162196392515
};
\addlegendentry{Efficient Laplace}
\end{axis}

\end{tikzpicture}
  \end{subfigure}\\
  
  \begin{subfigure}[b]{\textwidth}
    \raggedright
    \tikz\node[font=\bf,fill=C0,rounded corners=1pt]{Caltech-256};
  \end{subfigure}\\
  \begin{subfigure}[t]{.32\textwidth}
    \raggedleft
\begin{tikzpicture}

\definecolor{color0}{rgb}{1,0.498039215686275,0.0549019607843137}

\begin{axis}[
height=\figureheight,
tick align=outside,
tick pos=left,
width=\figurewidth,
x grid style={white!69.0196078431373!black},
xmin=0.47, xmax=5.53,
xtick style={color=black},
y grid style={white!69.0196078431373!black},
ylabel={FLOPs},
ymin=0, ymax=1478010217.25,
ytick style={color=black},
xtick = {1,2,3,4,5},
xticklabels = {1,2,3,4,5}
]
\draw[draw=none,fill=color0] (axis cs:1,0) rectangle (axis cs:0.7,339900000);
\draw[draw=none,fill=color0] (axis cs:2,0) rectangle (axis cs:1.7,685460000);
\draw[draw=none,fill=color0] (axis cs:3,0) rectangle (axis cs:2.7,1008160000);
\draw[draw=none,fill=color0] (axis cs:4,0) rectangle (axis cs:3.7,1254470000);
\draw[draw=none,fill=color0] (axis cs:5,0) rectangle (axis cs:4.7,1360530000);
\draw[draw=none,fill=black] (axis cs:1,0) rectangle (axis cs:1.3,340074310);
\draw[draw=none,fill=black] (axis cs:2,0) rectangle (axis cs:2.3,685808620);
\draw[draw=none,fill=black] (axis cs:3,0) rectangle (axis cs:3.3,1008611986);
\draw[draw=none,fill=black] (axis cs:4,0) rectangle (axis cs:4.3,1254946120);
\draw[draw=none,fill=black] (axis cs:5,0) rectangle (axis cs:5.3,1362234910);
\draw[draw=none,fill=black,fill opacity=0.5] (axis cs:1,0) rectangle (axis cs:1.3,349153435.666667);
\draw[draw=none,fill=black,fill opacity=0.5] (axis cs:2,0) rectangle (axis cs:2.3,703966871.333333);
\draw[draw=none,fill=black,fill opacity=0.5] (axis cs:3,0) rectangle (axis cs:3.3,1035849555);
\draw[draw=none,fill=black,fill opacity=0.5] (axis cs:4,0) rectangle (axis cs:4.3,1291263262.66667);
\draw[draw=none,fill=black,fill opacity=0.5] (axis cs:5,0) rectangle (axis cs:5.3,1407628778.33333);
\end{axis}

\end{tikzpicture}
  \end{subfigure}
  \hfill
  \begin{subfigure}[t]{.32\textwidth}
    \raggedleft
\begin{tikzpicture}

\definecolor{color0}{rgb}{1,0.498039215686275,0.0549019607843137}

\begin{axis}[
height=\figureheight,
tick align=outside,
tick pos=left,
width=\figurewidth,
x grid style={white!69.0196078431373!black},
xmin=0.47, xmax=5.53,
xtick style={color=black},
y grid style={white!69.0196078431373!black},
ymin=0, ymax=2930264671.25,
ytick style={color=black},
xtick = {1,2,3,4,5},
xticklabels = {1,2,3,4,5}
]
\draw[draw=none,fill=color0] (axis cs:1,0) rectangle (axis cs:0.7,514660000);
\draw[draw=none,fill=color0] (axis cs:2,0) rectangle (axis cs:1.7,1171180000);
\draw[draw=none,fill=color0] (axis cs:3,0) rectangle (axis cs:2.7,1844520000);
\draw[draw=none,fill=color0] (axis cs:4,0) rectangle (axis cs:3.7,2501400000);
\draw[draw=none,fill=color0] (axis cs:5,0) rectangle (axis cs:4.7,2742060000);
\draw[draw=none,fill=black] (axis cs:1,0) rectangle (axis cs:1.3,514949382);
\draw[draw=none,fill=black] (axis cs:2,0) rectangle (axis cs:2.3,1171826540);
\draw[draw=none,fill=black] (axis cs:3,0) rectangle (axis cs:3.3,1845339618);
\draw[draw=none,fill=black] (axis cs:4,0) rectangle (axis cs:4.3,2504610968);
\draw[draw=none,fill=black] (axis cs:5,0) rectangle (axis cs:5.3,2745336710);
\draw[draw=none,fill=black,fill opacity=0.5] (axis cs:1,0) rectangle (axis cs:1.3,524028251.666667);
\draw[draw=none,fill=black,fill opacity=0.5] (axis cs:2,0) rectangle (axis cs:2.3,1189984151.33333);
\draw[draw=none,fill=black,fill opacity=0.5] (axis cs:3,0) rectangle (axis cs:3.3,1872576323);
\draw[draw=none,fill=black,fill opacity=0.5] (axis cs:4,0) rectangle (axis cs:4.3,2540923502.66667);
\draw[draw=none,fill=black,fill opacity=0.5] (axis cs:5,0) rectangle (axis cs:5.3,2790728258.33333);
\end{axis}

\end{tikzpicture}
  \end{subfigure}
  \hfill
  \begin{subfigure}[t]{.32\textwidth}
    \raggedleft
    \pgfplotsset{ybar,legend image code/.code={\draw [#1,draw=none] (0cm,-0.08cm) rectangle (0.15cm,0.15cm);}}
\begin{tikzpicture}

\definecolor{color0}{rgb}{1,0.498039215686275,0.0549019607843137}

\begin{axis}[
height=\figureheight,
legend cell align={left},
legend style={
  font=\tiny,
  fill opacity=0.8,
  draw opacity=1,
  text opacity=1,
  at={(0.03,0.97)},
  anchor=north west,
  draw=white!80!black
},
tick align=outside,
tick pos=left,
width=\figurewidth,
x grid style={white!69.0196078431373!black},
xmin=0.47, xmax=5.53,
xtick style={color=black},
y grid style={white!69.0196078431373!black},
ymin=0, ymax=3469290025.25,
ytick style={color=black},
xtick = {1,2,3,4,5},
xticklabels = {1,2,3,4,5}
]
\draw[draw=none,fill=color0] (axis cs:1,0) rectangle (axis cs:0.7,615600000);
\addlegendimage{ybar,draw=none,fill=color0};
\addlegendentry{Vanilla MSDNet}

\draw[draw=none,fill=color0] (axis cs:2,0) rectangle (axis cs:1.7,1436390000);
\draw[draw=none,fill=color0] (axis cs:3,0) rectangle (axis cs:2.7,2283210000);
\draw[draw=none,fill=color0] (axis cs:4,0) rectangle (axis cs:3.7,2967420000);
\draw[draw=none,fill=color0] (axis cs:5,0) rectangle (axis cs:4.7,3253790000);
\draw[draw=none,fill=black] (axis cs:1,0) rectangle (axis cs:1.3,615959206);
\addlegendimage{ybar,draw=none,fill=black};
\addlegendentry{Efficient Laplace}

\draw[draw=none,fill=black] (axis cs:2,0) rectangle (axis cs:2.3,1437264444);
\draw[draw=none,fill=black] (axis cs:3,0) rectangle (axis cs:3.3,2284401570);
\draw[draw=none,fill=black] (axis cs:4,0) rectangle (axis cs:4.3,2968644280);
\draw[draw=none,fill=black] (axis cs:5,0) rectangle (axis cs:5.3,3258696030);
\draw[draw=none,fill=black,fill opacity=0.5] (axis cs:1,0) rectangle (axis cs:1.3,625037947.666667);
\addlegendimage{ybar,draw=none,fill=black,fill opacity=0.5};
\addlegendentry{Naive Laplace}

\draw[draw=none,fill=black,fill opacity=0.5] (axis cs:2,0) rectangle (axis cs:2.3,1455421671.33333);
\draw[draw=none,fill=black,fill opacity=0.5] (axis cs:3,0) rectangle (axis cs:3.3,2311637603);
\draw[draw=none,fill=black,fill opacity=0.5] (axis cs:4,0) rectangle (axis cs:4.3,3004959630.66667);
\draw[draw=none,fill=black,fill opacity=0.5] (axis cs:5,0) rectangle (axis cs:5.3,3304085738.33333);
\end{axis}

\end{tikzpicture}
  \end{subfigure}\\
  
  \begin{subfigure}[t]{.32\textwidth}
    \raggedleft
\begin{tikzpicture}

\begin{axis}[
height=\figureheight,
tick align=outside,
tick pos=left,
width=\figurewidth,
x grid style={white!69.0196078431373!black},
xlabel={Exit number},
xmin=0.47, xmax=5.53,
xtick style={color=black},
y grid style={white!69.0196078431373!black},
ylabel={FLOPs relative to vanilla},
ymin=1, ymax=1.05,
ytick style={color=black},
xtick = {1,2,3,4,5},
xticklabels = {1,2,3,4,5}
]
\addplot [semithick, black, mark=asterisk, mark size=3, mark options={solid}, only marks]
table {%
1 1.02722399431205
2 1.02699919956428
3 1.0274654370338
4 1.02932972702948
5 1.03461796383272
};
\addplot [semithick, black, mark=*, mark size=3, mark options={solid}, only marks]
table {%
1 1.00051282730215
2 1.00050859276982
3 1.00044832764641
4 1.00037953876936
5 1.00125312194512
};
\end{axis}

\end{tikzpicture}
    \vspace{-0.8em}
    \caption{\footnotesize small models}
  \end{subfigure}
  \hfill
  \begin{subfigure}[t]{.32\textwidth}
     \raggedleft
\begin{tikzpicture}

\begin{axis}[
height=\figureheight,
tick align=outside,
tick pos=left,
width=\figurewidth,
x grid style={white!69.0196078431373!black},
xlabel={Exit number},
xmin=0.47, xmax=5.53,
xtick style={color=black},
y grid style={white!69.0196078431373!black},
ymin=1, ymax=1.05,
ytick style={color=black},
xtick = {1,2,3,4,5},
xticklabels = {1,2,3,4,5}
]
\addplot [semithick, black, mark=asterisk, mark size=3, mark options={solid}, only marks]
table {%
1 1.01820279731603
2 1.0160557312568
3 1.01521063637152
4 1.01580055275712
5 1.01774879409398
};
\addplot [semithick, black, mark=*, mark size=3, mark options={solid}, only marks]
table {%
1 1.00056227800878
2 1.00055204153076
3 1.00044435300241
4 1.00128366834573
5 1.00119498114556
};
\end{axis}

\end{tikzpicture}
    \vspace{0.3em}
    \caption{\footnotesize medium models}
  \end{subfigure}
  \hfill
  \begin{subfigure}[t]{.32\textwidth}
     \raggedleft
\begin{tikzpicture}

\begin{axis}[
height=\figureheight,
legend cell align={left},
legend style={  font=\tiny,fill opacity=0.8, draw opacity=1, text opacity=1, draw=white!80!black},
tick align=outside,
tick pos=left,
width=\figurewidth,
x grid style={white!69.0196078431373!black},
xlabel={Exit number},
xmin=0.47, xmax=5.53,
xtick style={color=black},
y grid style={white!69.0196078431373!black},
ymin=1, ymax=1.05,
ytick style={color=black},
xtick = {1,2,3,4,5},
xticklabels = {1,2,3,4,5}
]
\addplot [semithick, black, mark=asterisk, mark size=3, mark options={solid}, only marks]
table {%
1 1.01533129900368
2 1.01324965457385
3 1.01245071763
4 1.01265059569143
5 1.01545758587166
};
\addlegendentry{Naive Laplace}
\addplot [semithick, black, mark=*, mark size=3, mark options={solid}, only marks]
table {%
1 1.00058350552307
2 1.00060877895279
3 1.00052188366379
4 1.00041257388573
5 1.00150778937793
};
\addlegendentry{Efficient Laplace}
\end{axis}

\end{tikzpicture}
    \vspace{0.3em}
    \caption{\footnotesize large models}
  \end{subfigure}
  \caption{Analysis on the test time computational cost of the Laplace approximation for different models. First rows show bar graphs comparing the computational costs of vanilla MSDNet (orange) and MSDNet with Laplace approximation (black) at each intermediate exit. The solid black bars represent the computational cost of efficient sampling from the predictive distribution, while the grey bars show the cost of \naive sampling. The second rows show the relative computational cost of using Laplace approximation on top of the vanilla MSDNet, both for the efficient and the \naive sampling approach. The figures assume 50 MC samples are drawn from the predictive distribution.}
\label{fig:lap_cost}
\end{figure*}
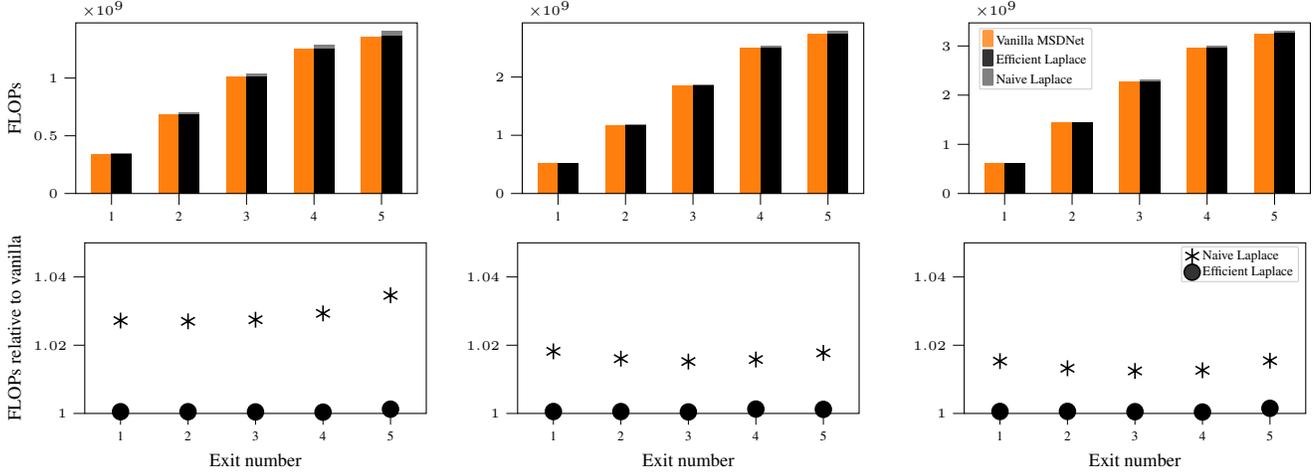

Based on this analysis, \cref{fig:lap_cost} shows the incremental test time computational cost of applying Laplace approximation on top of the vanilla MSDNet model for CIFAR-100, ImageNet, and Caltech-256, showing the increase in computation both for the efficient sampling approach and for the \naive sampling approach. For all data sets the small, medium, and large models are analysed separately, and for each model, the increase in computational cost at each intermediate exit is shown. The calculation of FLOPs in these figures differs from what is shown in the formulas above, as the FLOPs in the figure results are `practical FLOPs' that take into account the ability of most hardware to calculate sequential multiplication and addition in a single operation, resulting in one FLOP instead of two for such a pair of operations. Practical FLOPs are used also for the FLOPs calculation of all other numerical results in this paper. The results of the figure are obtained considering that 50 MC samples are drawn from the Laplace approximated predictive output distribution. The figures show that \naive approach of Laplace approximation adds considerable computation, but this can be mitigated by using the efficient sampling approach. The remaining added computational cost of efficient Laplace that is hard to see in \cref{fig:lap_cost} is 0.1--0.4\% on CIFAR-100, 0.06--0.16\% on ImageNet, and 0.04--0.15\% on Caltech-256 depending on the exit used. As a comparison, using a last layer MC dropout with 50 samples would add 3--9\% computation on CIFAR-100, 4--10\% on ImageNet, and 1--3\% on Caltech-256. 

Using MIE adds even less computation compared to our efficient Laplace. At each exit apart from the first one, the $c$ dimensional output is multiplied by the weight averaging weight $w_k$, added to the cumulative total sum, which is then divided by the total weight $\sum_{j=1}^{k}w_j$, adding up to $3c$ additional FLOPs at each exit after the first one. This results to additional 0.001--0.002\% computation on CIFAR-100, 0.0002--0.0009\% on ImageNet, and 0.0001--0.0002\% on Caltech-256.

\end{document}